\documentclass[10pt,twocolumn,letterpaper]{article}

\usepackage{iccv}
\usepackage{times}
\usepackage{epsfig}
\usepackage{graphicx}
\usepackage{amsmath}
\usepackage{subcaption}
\usepackage{amssymb}
\usepackage{xcolor, colortbl}
\usepackage[numbers,sort]{natbib} 
\usepackage{multibib}
\usepackage{graphicx}
\usepackage{booktabs}
\usepackage{multirow}
\usepackage{enumitem}
\usepackage{float}
\usepackage{dblfloatfix}

\usepackage{enumitem}
\usepackage[T1]{fontenc}

\usepackage{color}
\usepackage{tabularx}
\usepackage{tabulary}
\usepackage{hhline}
\setlist[itemize]{noitemsep, topsep=0pt}

\definecolor{Gray}{gray}{0.9}

\usepackage[pagebackref=true,breaklinks=true,colorlinks,bookmarks=false]{hyperref}

\iccvfinalcopy 

\def\iccvPaperID{6546} 
\newcommand{\COMMENT}[1]{}

\definecolor{myyellow}{RGB}{254, 254, 8}

\begin{document}

\title{From Kinematics To Dynamics: \\ 
Estimating Center of Pressure and Base of Support from Video Frames of Human Motion}
\author{
     Jesse Scott$^{1}$, Christopher Funk$^{1,3}$, Bharadwaj Ravichandran$^{1}$, \\ John H. Challis$^{2}$, Robert T. Collins$^{1}$, Yanxi Liu$^{1}$\\
    $^{1}$School of Electrical Engineering and Computer Science. \quad
    $^{2}$Biomechanics Laboratory.\\
	The Pennsylvania State University, University Park, PA 16802 USA \\ $^{3}$Kitware, Inc\\
	{\small
    	\href{mailto:jus121@psu.edu}{jus121@psu.edu},
    	\href{mailto:christopher.funk@kitware.com}{christopher.funk@kitware.com},
    	\href{mailto:bzr49@psu.edu}{bzr49@psu.edu},
    } \\
    {\small 
    	\href{mailto:jhc10@psu.edu}{jhc10@psu.edu},
    	\href{mailto:rcollins@cse.psu.edu}{rcollins@cse.psu.edu},
    	\href{mailto:yanxi@cse.psu.edu}{yanxi@cse.psu.edu}
	}
}

	\date{}

\maketitle

\begin{abstract}
To gain an understanding of the relation between a given human pose image and the corresponding physical foot pressure of the human subject, we propose and validate two end-to-end deep learning architectures, PressNet and PressNet-Simple, to regress foot pressure heatmaps (dynamics) from  2D human pose (kinematics) derived from a video frame. A unique video and foot pressure data set of 813,050 synchronized pairs, composed of 5-minute long choreographed Taiji movement sequences of 6 subjects, is collected and used for leaving-one-subject-out cross validation. Our initial experimental results demonstrate reliable and repeatable foot pressure prediction from a single image, setting the first baseline for such a complex cross modality mapping problem in computer vision. Furthermore, we compute and quantitatively validate the Center of Pressure (CoP) and Base of Support (BoS) from predicted foot pressure distribution, obtaining key components in pose stability analysis from images with potential applications in kinesiology, medicine, sports and robotics.
\end{abstract}

\section{Introduction}
Current human pose studies in computer vision research focuses on extracting skeletal kinematics from videos, using body pose estimation and tracking to infer pose in each frame as well as the movement of body and limbs over time~\cite{cao2017realtime,bulat2016human,newell2016stacked,chen2016synthesizing,toshev2014deeppose,chen2014articulated,fan2015combining,guler2018densepose}. However, an effective analysis of human movement also must take into account the dynamics of the human body~\cite{Seethapathi2019}. Understanding body dynamics, such as foot pressure, is essential to study the effects of perturbations caused by external forces and torques on the human postural system, which change body equilibrium in static posture and during locomotion~\cite{winter95balancestandandwalk}.  Such analysis of stability has a wide range of applications in the fields of healthcare, kinesiology, and robotics.

\COMMENT{
In the realm of health and sports, precise and quantitative digital recording and analysis of human motion provides rich content for performance characterization and training, health status assessment, and diagnosis or preventive therapy of neurodegenerative syndromes.  Analysis of gait and control of balance/equilibrium has received increasing interest
from the research community~\cite{perry1992gait, winter1991biomechanics, peterka2004dynamic} as a way to study the complex mechanisms of the human postural system for maintaining stable pose. Stability analysis has a wide range of applications in the fields of Healthcare, Kinesiology and Robotics to understand locomotion and replicate human body movements. Understanding body dynamics, such as foot pressure, is essential to study the effects of perturbations caused by external forces and torques on the human postural system, which change body equilibrium in static posture as well as during locomotion~\cite{winter95balancestandandwalk}. 
}
\begin{figure}[t] \centering
    \includegraphics[width=1.0\linewidth]{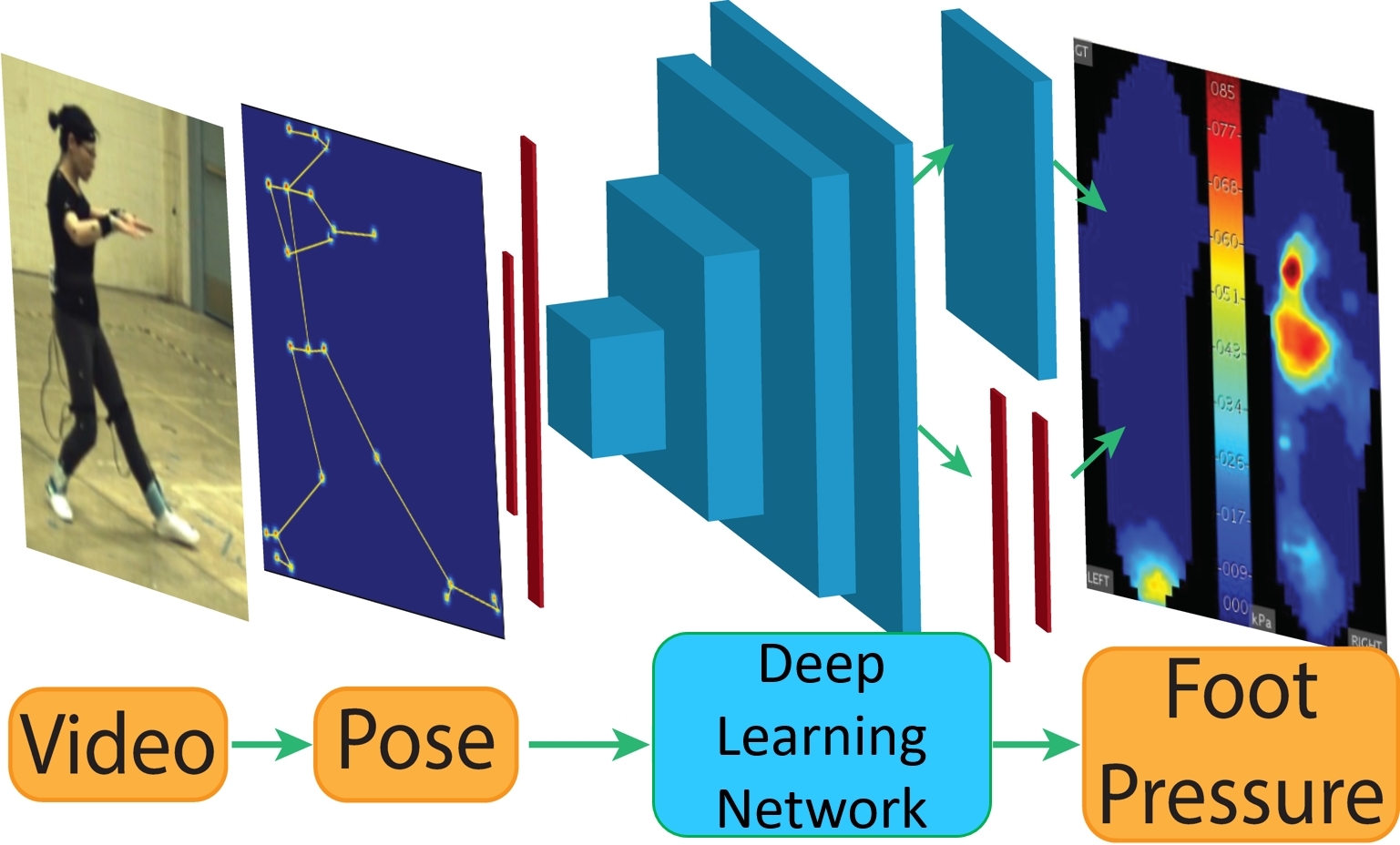}
    \vspace{-10pt}
    \caption{Our PressNet and PressNet-Simple networks learn to predict a foot pressure heatmap from 2D human body joints extracted from a video frame using OpenPose~\cite{cao2017realtime}.}
    \label{fig:1} \vspace{-10pt}
\end{figure}

We have chosen 24-form simplified {\em Taiji Quan}~\cite{wang_BMC_CAM2010} as a testbed for validating our computer vision and machine learning algorithms. Taiji was selected because it is a low-cost, hands-free, and slow-moving exercise sequence containing complex body poses and movements. Taiji is practiced worldwide by millions of people of all genders, races, and ages.  Each routine lasts about 5 minutes and consists of controlled choreographed movements where the subject attempts to remain balanced and stable at all times.

\begin{figure}[!t] \centering
    \resizebox{1.00\linewidth}{!}{
    \vspace{-10pt}
    \setlength{\tabcolsep}{0pt}
    \begin{tabular}{ccc} 
    \includegraphics[width=.62\linewidth]{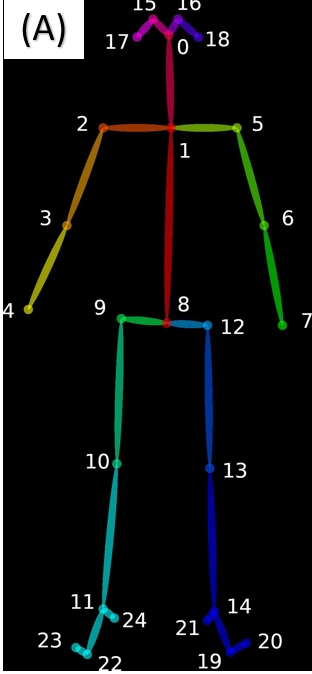} & \includegraphics[width=\linewidth]{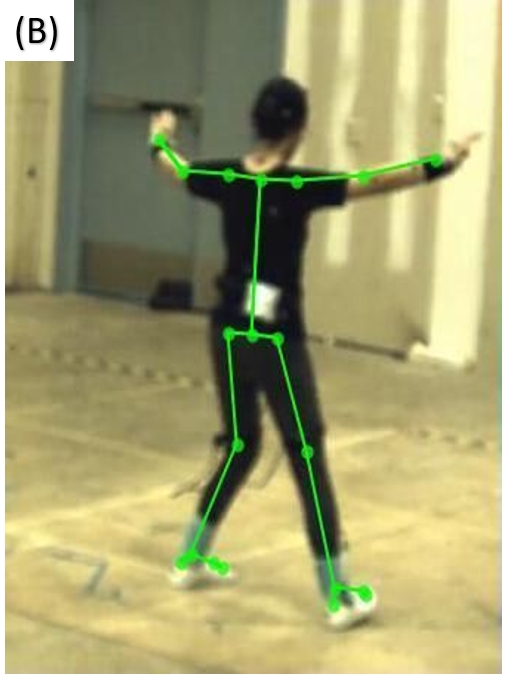} & 
    \includegraphics[width=1.1\linewidth]{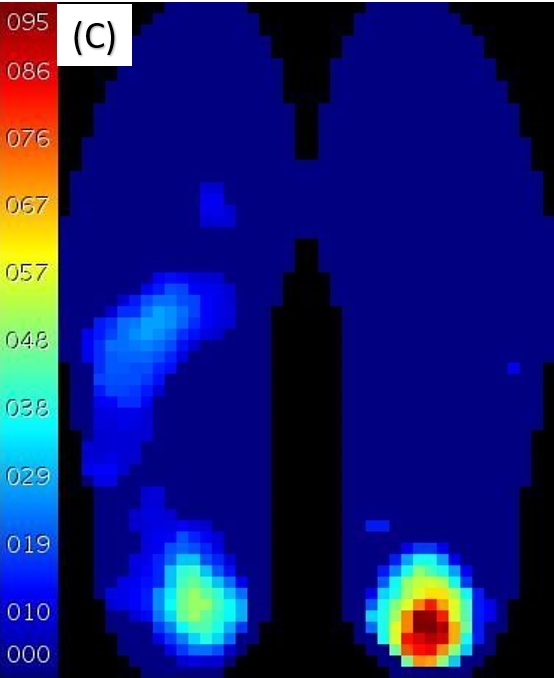}\\ 
    \end{tabular}
    } 
    \vspace{-10pt}
    \caption {\textbf{(A)}: Body25 joint set labeled by Openpose~\cite{cao2017realtime}. \textbf{(B)}: Video data with Body25 joint overlay. \textbf{(C)}: The corresponding measured left and right foot pressure.}
    \label{fig:2}
    \vspace{-10pt}
\end{figure}

To understanding the relation between a body pose and the corresponding foot pressure of the human subject (Figure~\ref{fig:1}), we explore two deep convolutional residual architectures, PressNet and PressNet-Simple (Figure~\ref{fig:res2}), and train them on the largest human motion sequence dataset ever recorded of synchronized video and foot pressure data, containing a total of 813,050 data pairs of 2D pose with corresponding foot pressure measurements (a sample pair is shown in Figure \ref{fig:2}B and \ref{fig:2}C).  Body pose is input to the network as 2D human joint locations extracted using the Openpose~\cite{cao2017realtime} Body25 model. The network predicts as output a foot pressure intensity heatmap that provides the distribution of pressure applied by different points of the foot against the ground, measured in kilopascals (kPa) over discretized foot sole locations. 

The major contributions of this work include:
\textbf{1) Novelty}:
Our PressNet and PressNet-Simple networks  are the first vision-based networks to regress human dynamics (foot pressure) from kinematics (body pose).
Little is known whether quantitative information about dynamics can be inferred from single-view video frame, and we answer this question in the affirmative.
\textbf{2) Dataset}: We have collected the largest  synchronized video and foot pressure dataset ever recorded of a long complex human movement sequence.  This dataset will be made available upon publication.
\textbf{3) Application}: As a sample validation, Center of Pressure (CoP) and Base of Support (BoS) are computed from  regressed foot pressure maps and compared to ground truth. These are two key components in the analysis of bipedal stability, with applications in kinesiology, biomechanics, healthcare, and robotics.

\COMMENT{
Using the steps shown in Figure~\ref{fig:1}, we explore two end-to-end deep learning approaches, PressNet (Figure~\ref{fig:res2}) and PressNet-Simple (Figure~\ref{fig:PressNet-Simple}), to transform kinematics (body pose) to dynamics (foot pressure), and to obtain Center of Pressure (CoP) locations from the regressed foot pressure. In order to achieve this goal, we have created the largest human motion sequence dataset of synchronized video and foot pressure data, with a total of over 810k data pairs of 2D pose with corresponding foot pressure measurements, like the sample shown in Figure \ref{fig:2}B and \ref{fig:2}C.  We represent foot pressure by an intensity heatmap that provides the distribution of pressure applied by different points of the foot against the ground, measured in kilopascals (kPa) over discretized foot sole locations. Body pose is represented by 2D human joint locations extracted using the Openpose~\cite{cao2017realtime} Body25 model on the video frames. We record video and foot pressure maps simultaneously, so  there is a foot pressure map for both feet corresponding to each video frame.
}

\section{Related Work}
Seethapathi et al. \cite{Seethapathi2019} reviewed the limitations of video-based measurement of human motion for use in movement science. One of the enumerated technical hurdles is the need to estimate contact forces. They indicated that more accurate kinematics and the estimation of dynamics information should be the key computer vision research goals in order to use computer vision as a tool in biomechanics.  In this paper we seek to use a body's kinematics to predict its dynamics and hence develop a quantitative method to analyze human stability using foot pressure derived from video. 

Studying human stability during standing and locomotion~\cite{fp1, eckardt2018healthy, arvin2018step} is typically addressed by direct measurement of foot pressure using force plates or insole foot pressure sensors. Previous studies have shown that foot pressure patterns can be used to discriminate between walking subjects~\cite{pataky_etal2012, rodriguez_etal2013}. Instability of the CoP of a standing person is an indication of postural sway and thus a measure of a person's ability to maintain balance~\cite{pai2003movement, hof2008extrapolated, hof2007equations, KoEtal2015}. Grimm~\etal~\cite{fp2} predict the pose of a patient using foot pressure mats. The authors of \cite{McKay2017} and \cite{PUTTI2010} evaluate foot pressure patterns of 1000 subjects over ages 3 to 101 and determine there is a significant difference between gender in the contact area but not in magnitude of foot pressure for adults. As a result the force applied by females is lower but is accounted for by female mass also being significantly lower. In \cite{Cai_2018}, a depth regularization model is trained to estimate dynamics of hand movement from 2D joints obtained from RGB video cameras. Stability analysis of 3D printed models is presented in \cite{Prevost_2013, Bacher_2014, Prevost_2016}. Although these are some insightful ways to analyze stability, there has been no vision-based or deep learning approach to tackle this problem.

Estimation of 2D body pose in images is a well-studied problem in computer vision, with state of the art methods being based on deep networks~\cite{guler2018densepose, gilbert2018fusing, cao2017realtime, huang2017densely, bulat2016human, newell2016stacked, chen2016synthesizing, fan2015combining, toshev2014deeppose, chen2014articulated, tomp}.  We adopt one of the more popular approaches, CMU's OpenPose~\cite{cao2017realtime}, to compute the 2D pose input to our networks. Success in 2D human pose estimation also has encouraged researchers to detect 3D skeletons by extending existing 2D human pose detectors~\cite{bogo2016keep, simo2012single, chen20173d, moreno20173d, nie2017monocular, martinez2017simple, zhou2017towards} or by directly using image features~\cite{agarwal20043d, pavlakos2017coarse, zhou2016sparseness, sun2017compositional, DBLP:journals/corr/abs-1803-00455}. Martinez~\etal~\cite{martinez2017simple} showed that given high-quality 2D joint information, the process of lifting 2D pose to 3D pose can be done efficiently using a relatively simple deep feed-forward network. All these papers concentrate on pose estimation by learning to infer joint angles or joint locations, which can be broadly classified as learning basic kinematics of a body skeleton. These methods do not delve into the external torques/forces exerted by the environment, balance, or physical interaction of the body with the scene.

\begin{figure}[!t]
    \vspace{-1em}
    \begin{center}
        \includegraphics[width=.9\linewidth]{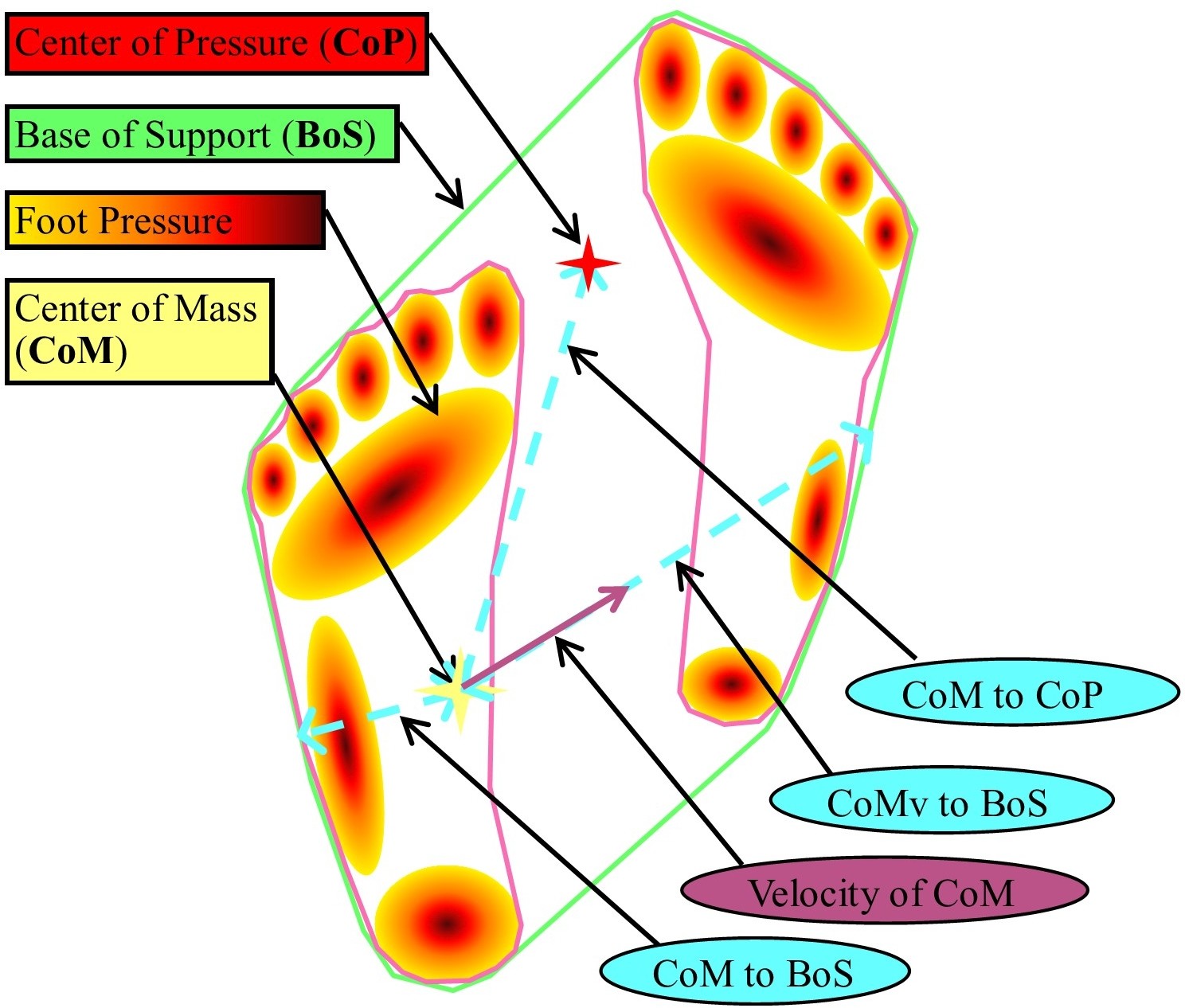}
    \end{center} \vspace{-10pt}
    \caption[Overview of Stability and CoP]{Overview of concepts in stability. Depiction of the components of stability CoP, CoM, and BoS as well as the relation of those components to stability metrics.} \vspace{-10pt}
    \label{fig:sup:6} 
\end{figure}

\COMMENT{
\noindent{\bf Basic Concepts in Stability:}
A major motivation for computing foot pressure maps from video is the application to stability analysis. Fundamental elements used in stability analysis are illustrated in Figure~\ref{fig:sup:6}. These include Center of Mass (CoM), Base of Support (BoS) polygon, and Center of Pressure (CoP). 
The relative locations of  CoP, BoS and CoM  have  been identified as a determinant of stability in a variety of tasks~\cite{hof2007equations,hof2008extrapolated,pai2003movement}.
}

\subsection{Stability}
A major motivation for computing foot pressure maps from video is the application to stability analysis. Fundamental concepts used in stability analysis are illustrated in Figure~\ref{fig:sup:6}. These include Center of Mass (CoM), Base of Support (BoS), and Center of Pressure (CoP). CoM, also known as Center of Gravity, is the ground projection of the body's 3D center of mass~\cite{gos}. Generally speaking, human pose is stable if the CoM is contained withing the convex hull of the BoS, also called the support polygon~\cite{mrozowski2007analysis}.  If the CoM point is outside the support polygon, it is equivalent to the presence of an uncompensated moment acting on the foot, causing rotation around a point on the polygon boundary, resulting in instability and a potential fall. CoP, also known as the Zero Moment Point, is a point where the total moment generated due to gravity and inertia equals zero. Figure~\ref{fig:sup:6} shows a diagram of foot pressure annotated with the CoP (shown as a red star), with pressure from both feet shown as regions color-coded from low pressure (yellow) to moderate pressure (red) to high pressure (brown). Considering CoP as the ground reaction force and CoM as the opposing force, larger distances between the two 2D points could indicate reduced stability. Specifically, the CoP location relative to the whole body center of mass has been identified as a determinant of stability in a variety of tasks  \cite{hof2007equations,hof2008extrapolated,pai2003movement}. Note that the CoP is usually measured directly by force plates or insole foot pressure sensors, whereas in this paper we develop a method that can infer it from video alone. We quantitatively evaluate our results using ground truth data collected by insole foot pressure sensors. While Figure~\ref{fig:sup:6} shows a representation of CoP as a combination of both feet, in our research we will focus on the prediction of foot pressure of each foot separately, to be then spatially registered using motion capture data for foot position and orientation. This allows our analysis to focus on the accuracy of the predicted pressure rather than the position and orientation of each foot. Pressure from each foot is combined using the ground truth feet positions and orientations (based on ankle and toe position from motion capture) to create a full body CoP.

\begin{table*}[!t] \centering
    \resizebox{1\linewidth}{!}{%
    \begin{tabular}{|r||c|c|c|c||c|c|c||c|c|} \hline
    & \multicolumn{4}{c||}{Demographics} & \multicolumn{3}{c||}{Dataset (\# of frames)} & \multicolumn{2}{c|}{Pressure (kPa)} \\ \hline
    Subject & Mass (kg) & Height (m) & Experience (y) & Gender & Training Set & Validation Set & Test Set & Mean & Std \\ \hline
    1 & 52.20 & 1.60 & 9 & Female & 158,875 & 65,417 & 588,758 & 6.44 & 19.31 \\
    2 & 66.67 & 1.72 & 10 & Male & 123,825 & 68,922 & 620,203 & 6.18 & 32.39 \\
    3 & 63.50 & 1.60 & 6 & Female & 101,950 & 71,110 & 639,990 & 6.67 & 28.34 \\
    4 & 77.11 & 1.70 & 9 & Male & 146,700 & 66,635 & 599,715 & 9.46 & 33.46 \\
    5 & 60.00 & 1.56 & 5 & Female & 123,915 & 68,913 & 620,222 & 10.54 & 34.90 \\
    6 & 55.00 & 1.54 & 32 & Female & 157,785 & 65,526 & 589,739 & 9.25 & 35.36 \\  \hline \rowcolor{Gray}
    Mean & 62.41 & 1.62 & 12 &  & 135,508 & 67,754 & 609,771 & 8.09 & 30.63 \\ \rowcolor{Gray}
    Std & 8.17 & 0.07 & 9 &  & 20,674 & 2,068 & 18,597 & 1.71 & 5.55 \\ \hline
    \end{tabular}%
    }
    \vspace{-10pt}
    \caption{Dataset demographic information including subject mass (kilogram), height (meter), experience (year), and gender; dataset information including number of frames for training, validation, and testing sets; and pressure (kiloPascals) statistics including mean and standard deviation. A total of 813,050 frames are available as input for training and testing the KNN, PressNet, and PressNet-Simple methods.}
    \label{tab:subject_stats} 
    \label{tab:data_splits}
    \label{tab:data_stat}
    \vspace{-5pt}
\end{table*}

\section{Data Collection}
To support this research we have collected a large, tri-modal data set containing synchronized video, motion capture, and foot pressure data. We have chosen to record 24-form simplified {\em Taiji Quan}~\cite{wang_BMC_CAM2010} because it is a low-cost, hands-free, and slow-moving exercise sequence containing complex body poses and movements (in all orientations) and is practiced worldwide by people of all genders, races, and ages.  Each routine lasts about 5 minutes and consists of controlled choreographed movements where the subject attempts to remain balanced and stable at all times. 3D mocap data is not used for network training but only used for 2-feet CoP and BoS validations.

\COMMENT{
We present the first tri-modal, choreographed 24-Form Taiji  data set containing synchronized video, motion capture, and foot pressure data (Table~\ref{tab:data_stat}). The subjects wear motion capture markers and insole foot pressure measurement sensors while being recorded. Foot pressure sensor arrays, connected to the Tekscan F-scan measurement system, are inserted as insoles in the shoes of the subject during the performance. Vicon Nexus software is used to spatiotemporally record motion capture and video data in hardware while Tekscan F-scan software is used to simultaneously record foot pressure sensor measurements that are then synchronized to the other data post collection. While used in the analysis of our results, motion capture data is not used in any of the experiments in this paper because:
\begin{enumerate}[noitemsep,topsep=0pt] \setlist{nosep}
    \item We intend to create an end-to-end system to regress foot pressure maps, and hence CoP locations, directly from video;
    \item Video data collection is inexpensive and has very few hardware requirements as compared to the cumbersome process of motion capture data collection and processing; and 
    \item There are multiple existing pose prediction networks that can be used to extract 2D human body keypoints directly from video, to use as input to our network.
\end{enumerate}
}

\begin{figure*}[!t] \centering
    \resizebox{1\linewidth}{!}{
    \setlength{\tabcolsep}{-30pt}
    \begin{tabular}{cccc}
    \includegraphics{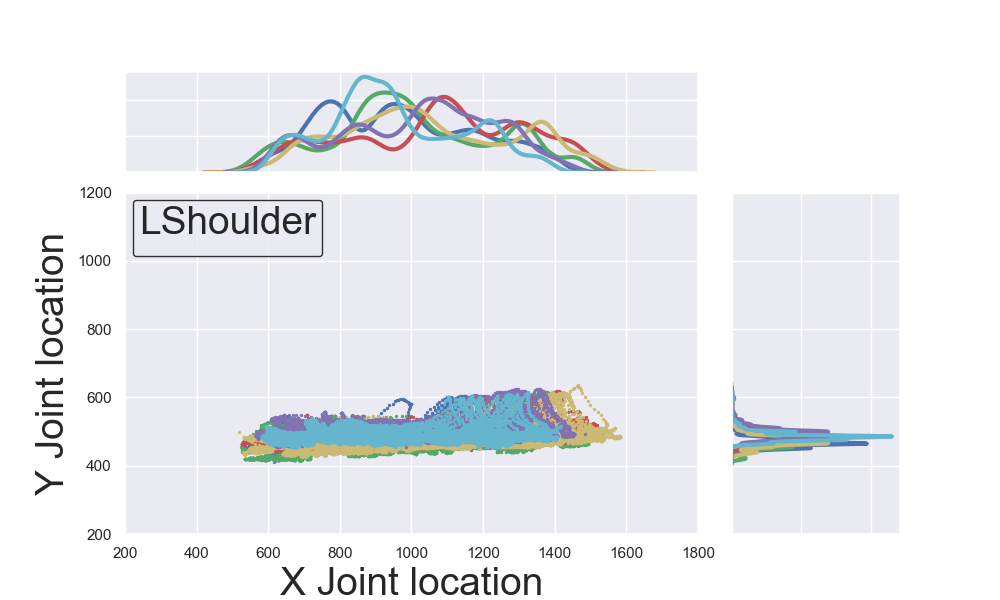} & \includegraphics{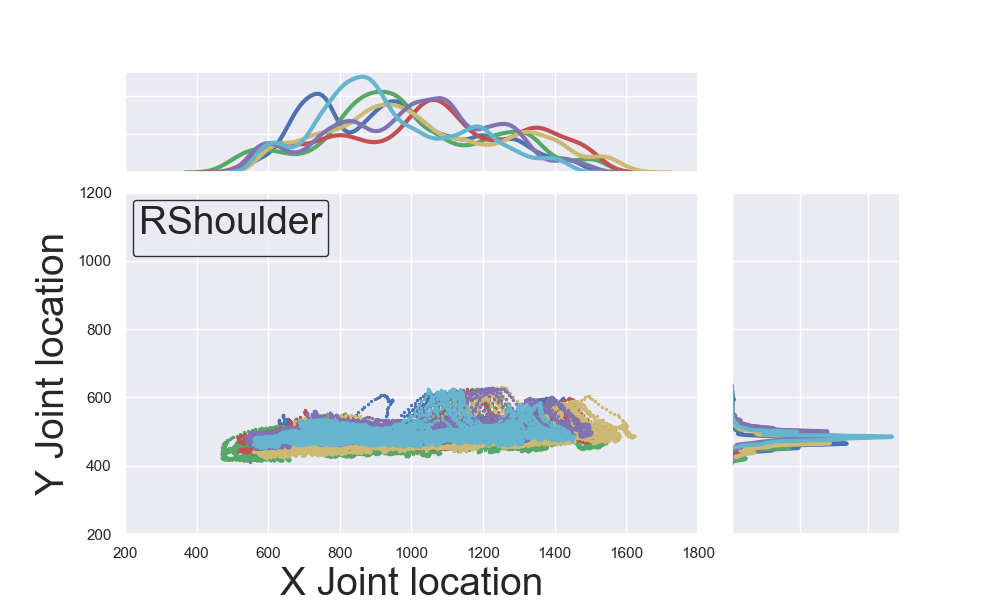} & 
    \includegraphics{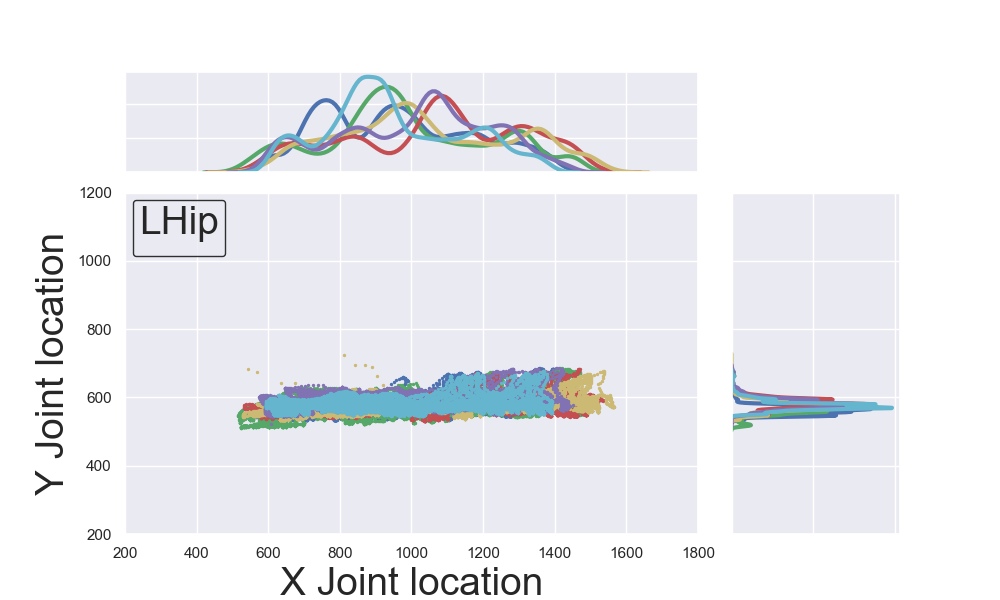}      & \includegraphics{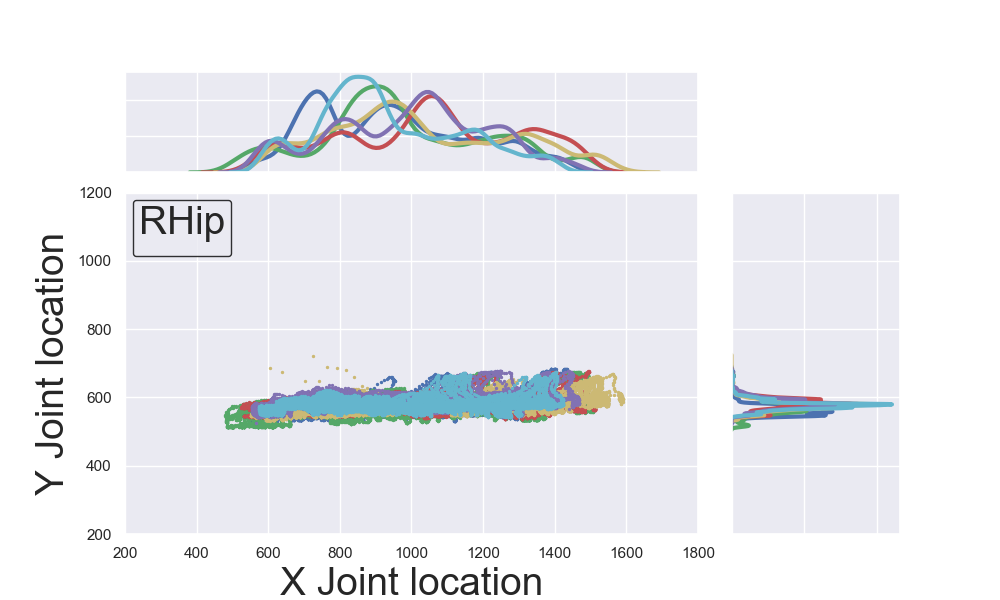}      \\      
    \includegraphics{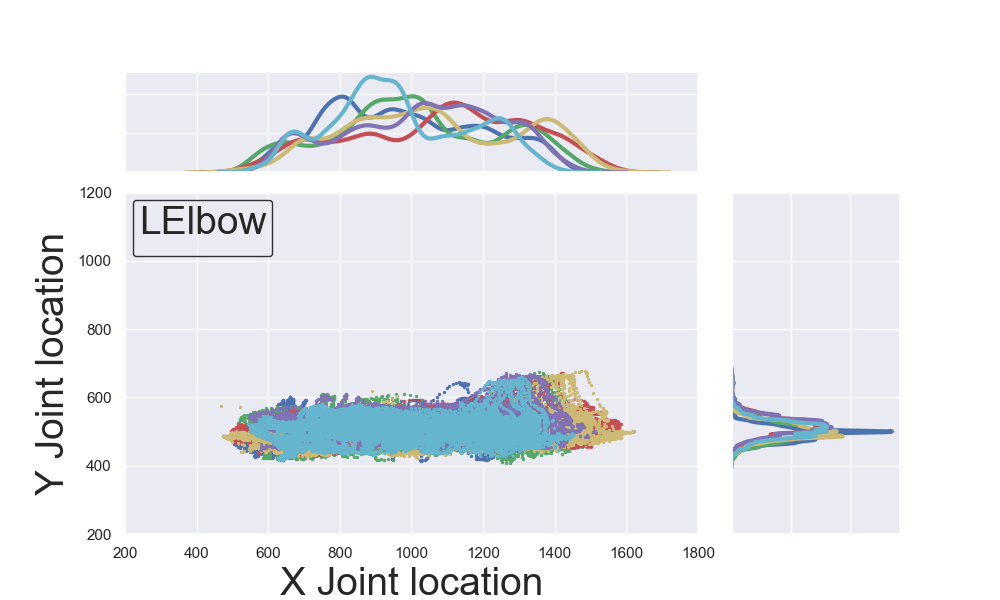}    & \includegraphics{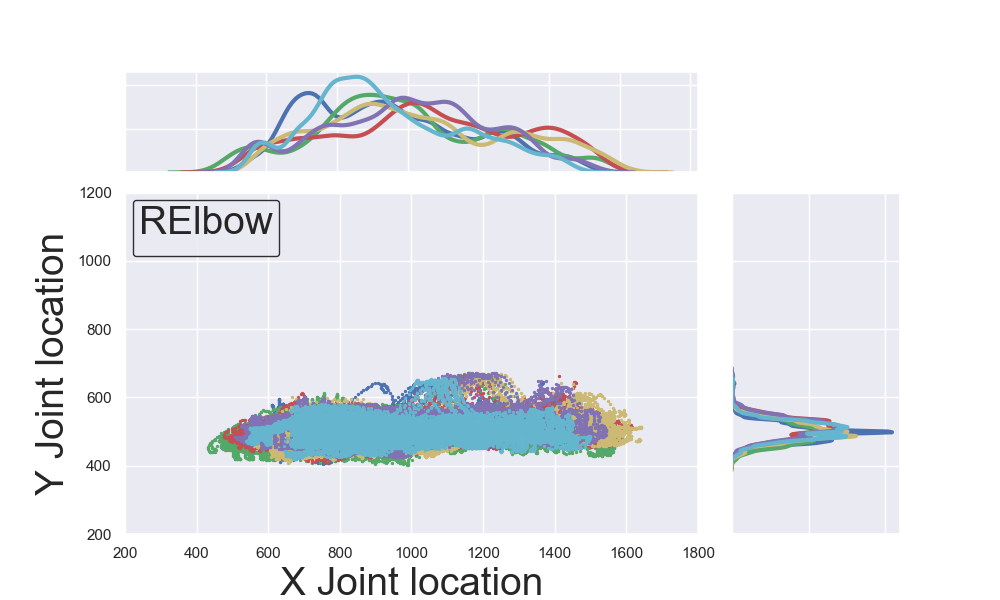}    & 
    \includegraphics{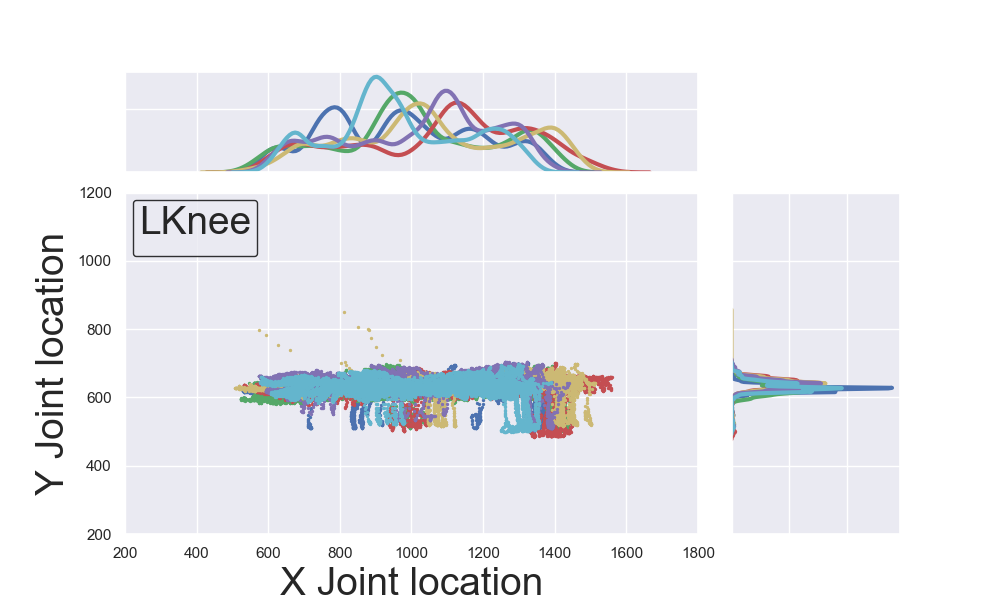}     & \includegraphics{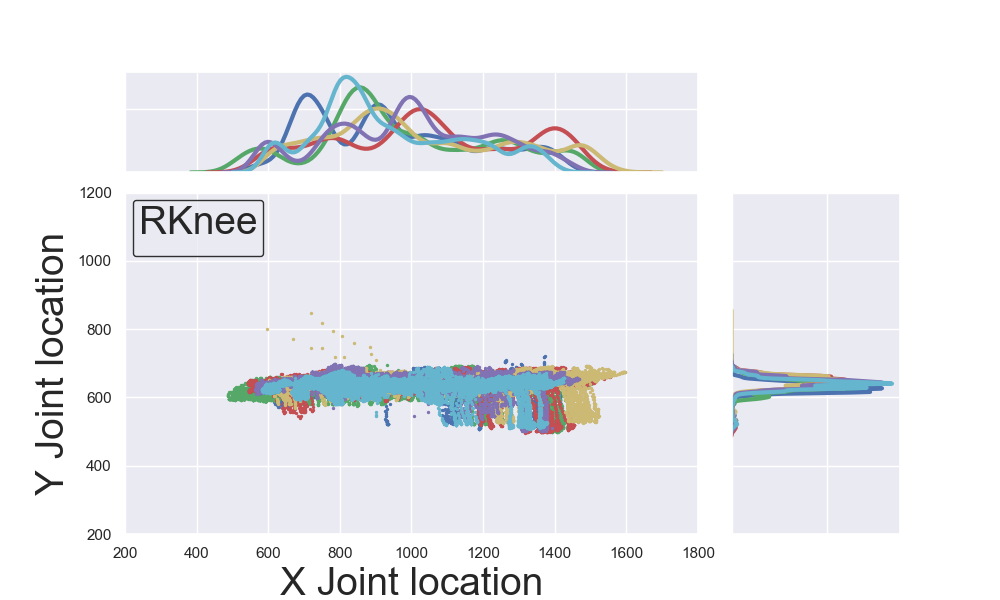}     \\
    \includegraphics{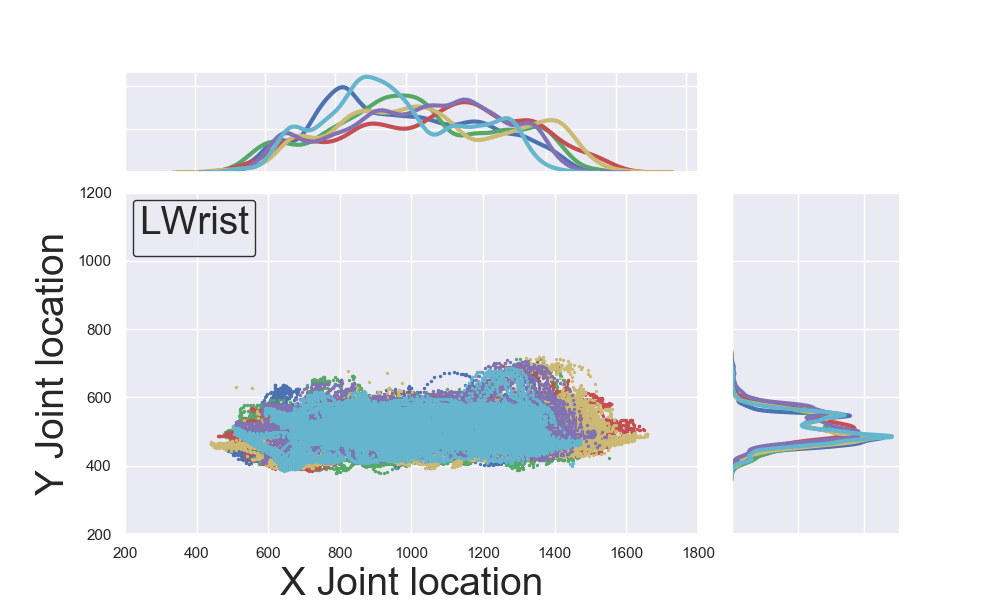}    & \includegraphics{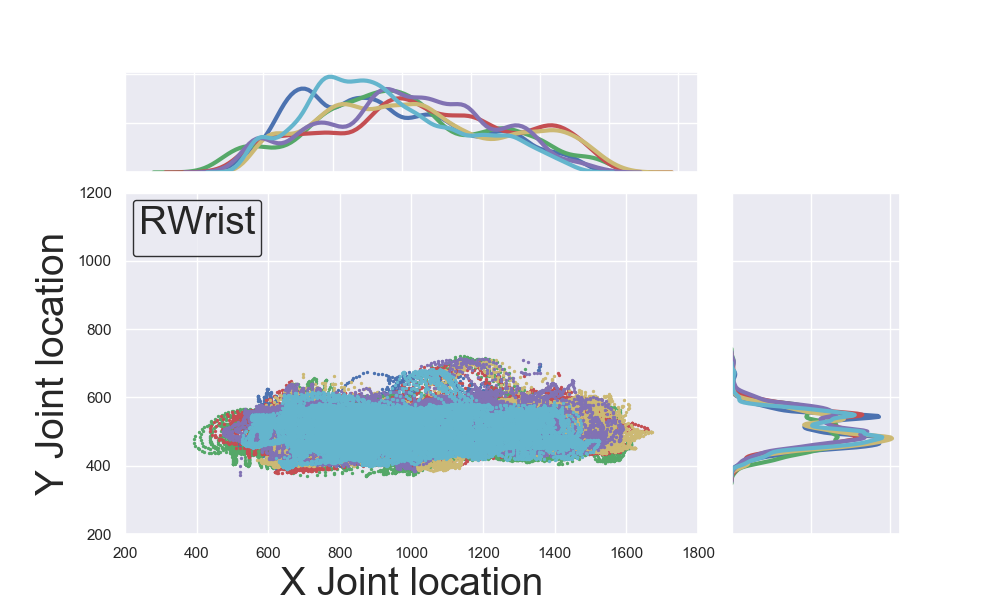}    &
    \includegraphics{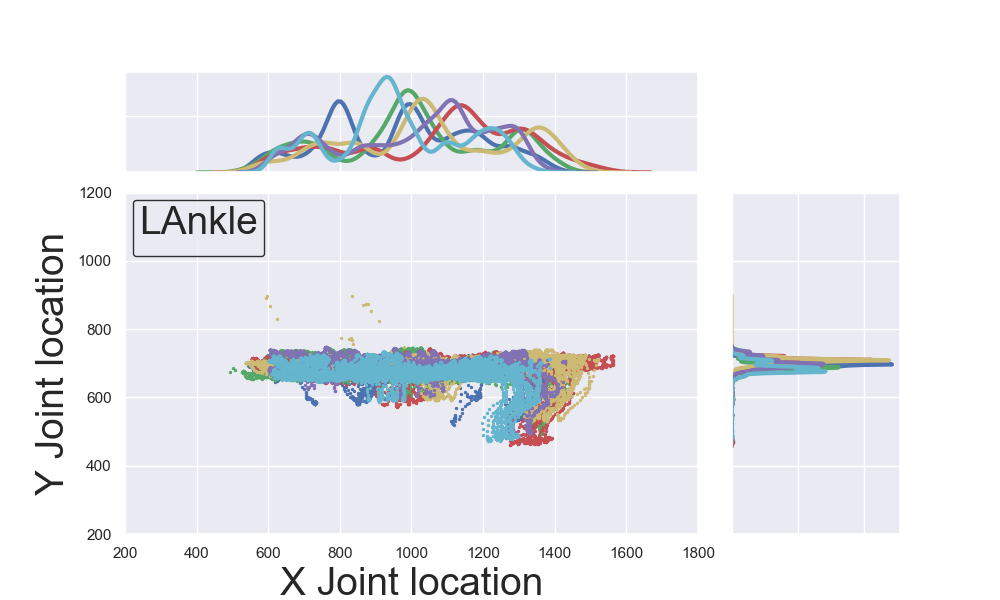}    & \includegraphics{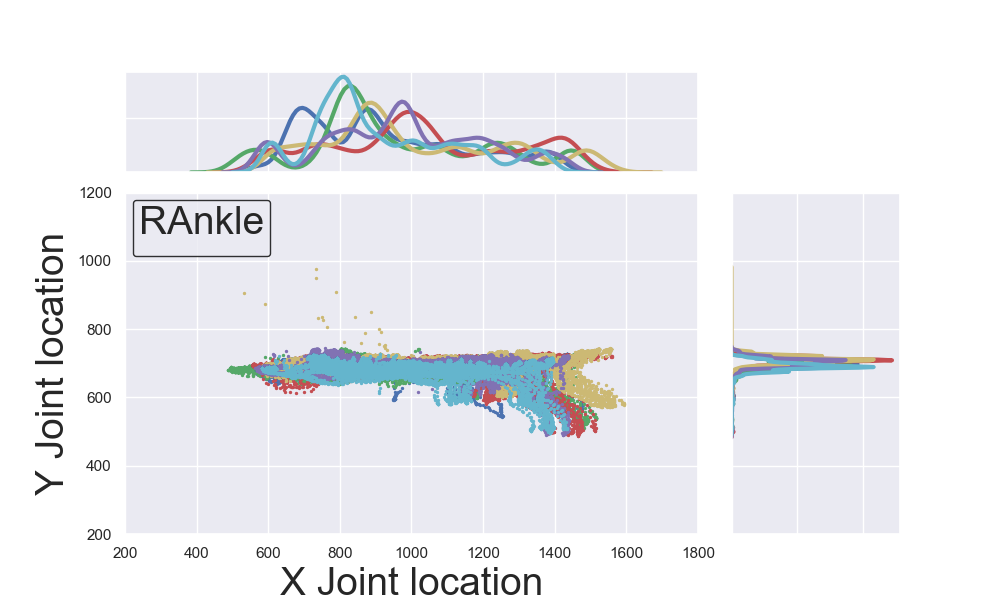}    \\
    \includegraphics{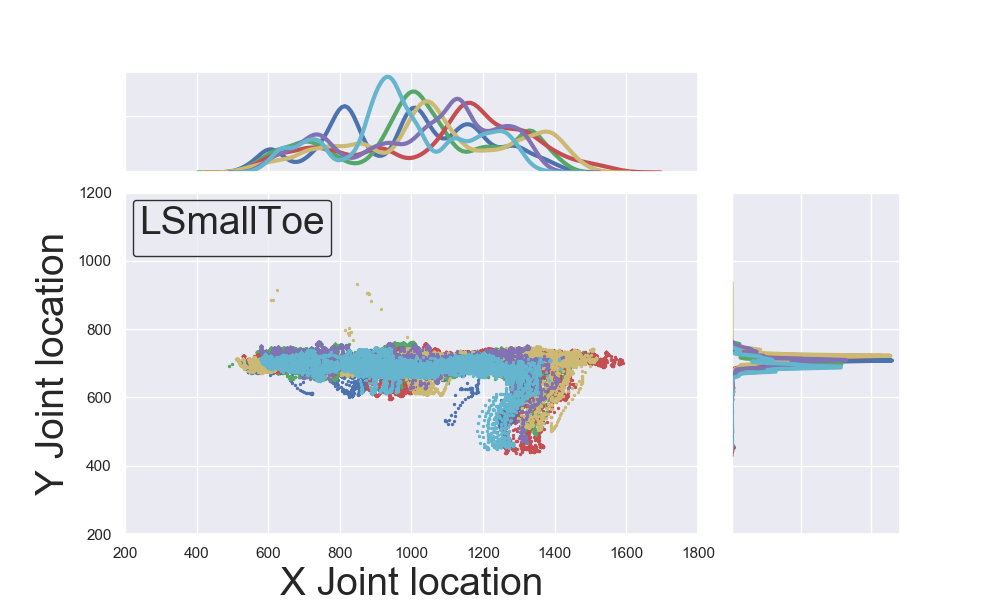} & \includegraphics{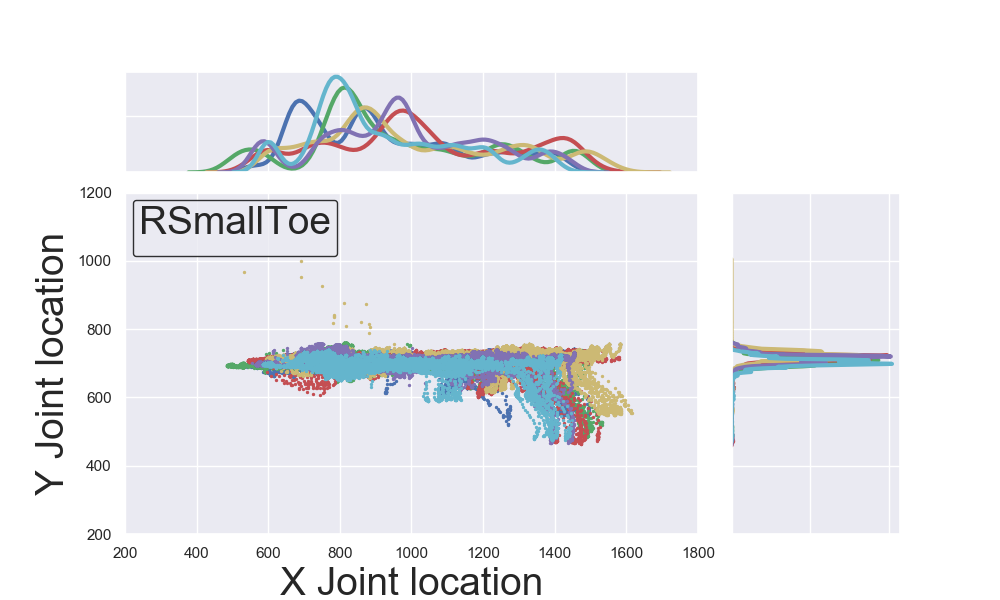} &
    \includegraphics{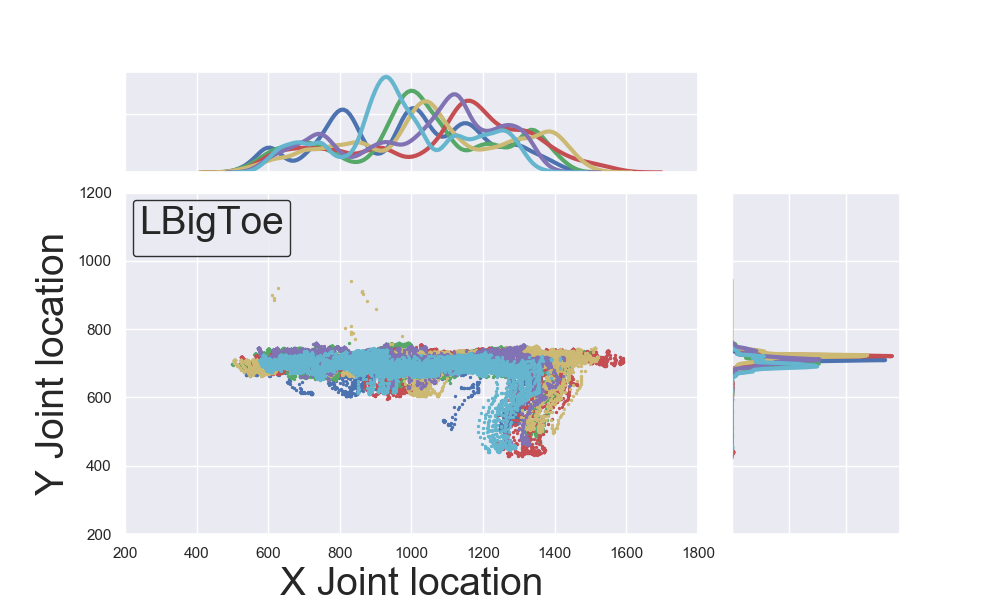}   & \includegraphics{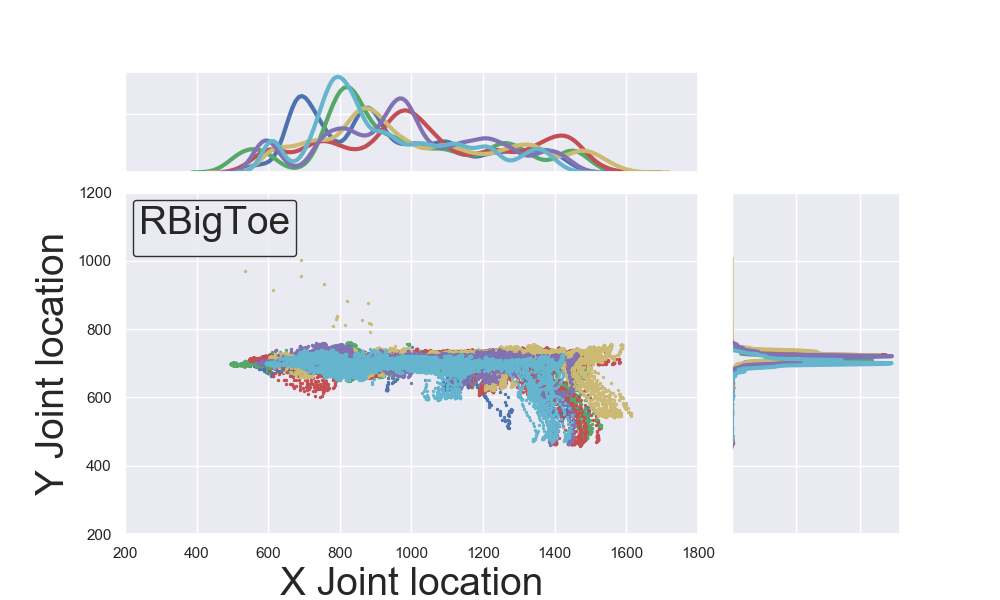}   \\
    \includegraphics{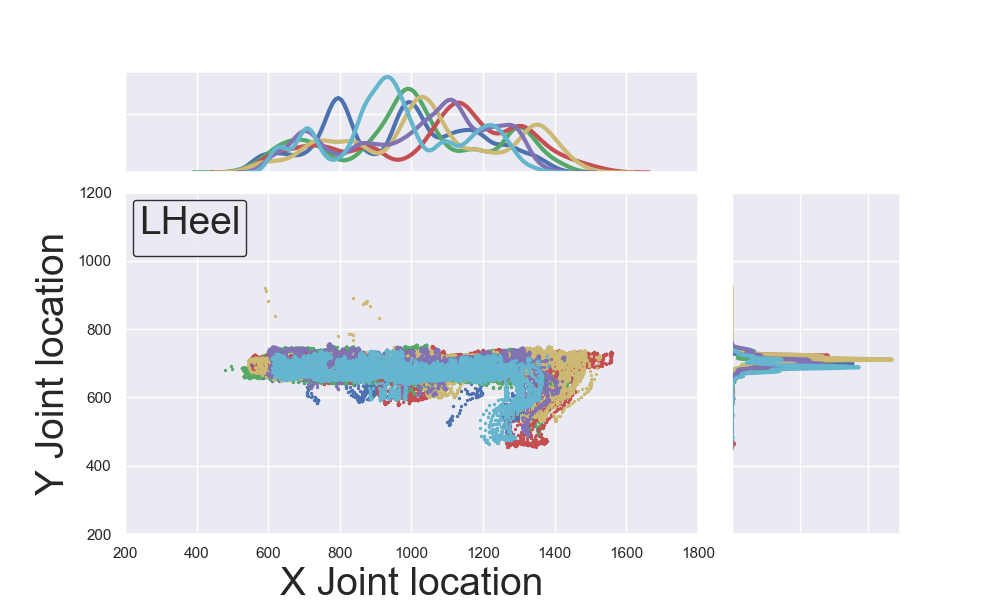}     & \includegraphics{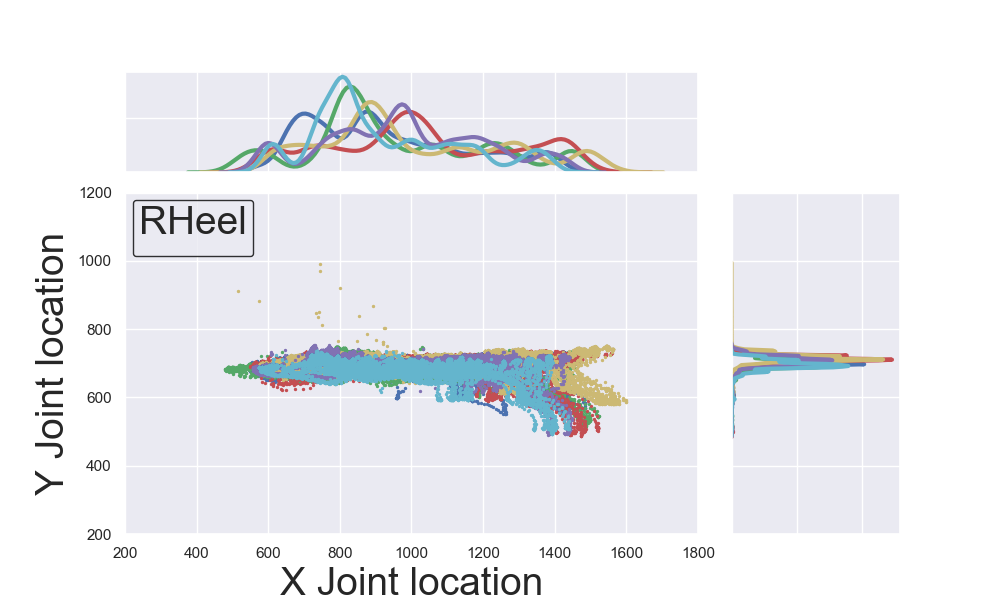}     &
    \includegraphics{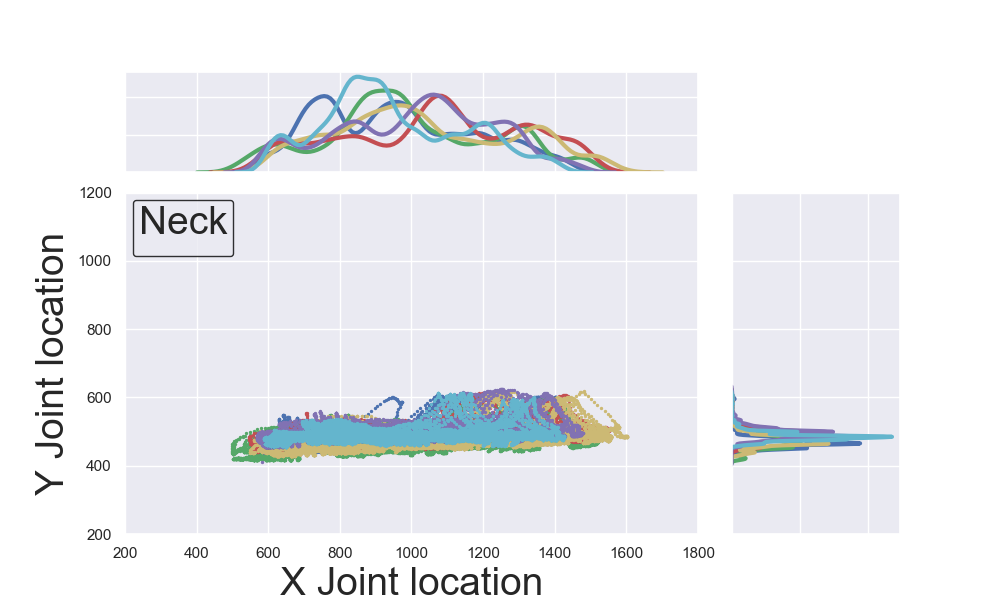}      & \includegraphics{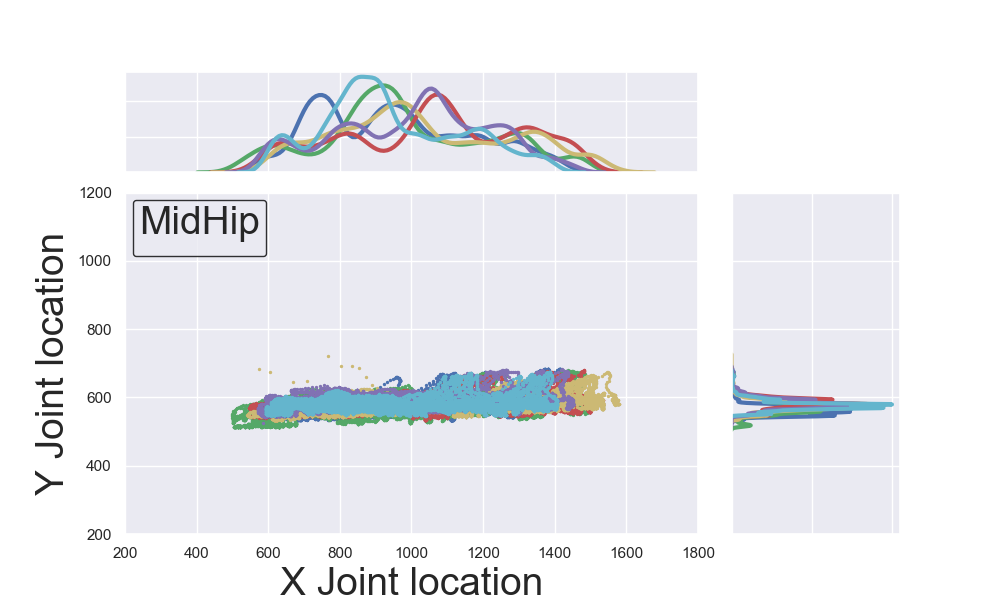}    \\
    \end{tabular}
    }
    \vspace{-10pt}
    \caption{Per body joint Kernel Density plots and 2D scatter plots of OpenPose~\cite{cao2017realtime} data for all subjects showing spatial similarity of joints between subjects in the dataset. The data for each subject is represented with different colors: Subject 1 = Blue, 2 = Orange, 3 = Green, 4 = Red, 5 = Purple, and Subject 6  = Brown. }
    \label{fig:hist}  \vspace{-5pt}
\end{figure*}

\subsection{Video and Pose Extraction}
Raw video data is collected at 50 fps and processed using Vicon Nexus and FFmpeg to transcode to a compressed video, with each video having its own spatiotemporal calibration. Human pose predictions are extracted from the compressed video using OpenPose~\cite{cao2017realtime}. OpenPose Body25 model uses non-parametric representations called Part Affinity Fields to regress joint positions and body segment connections between the joints. The output from OpenPose  has 3 channels, (X, Y, confidence), denoting the X and Y pixel coordinates and confidence of prediction for each of the 25 joints, making it an array of size $(25 \times 3)$.

\par Figure~\ref{fig:2}A shows the Body25 joints labeled by OpenPose. The 25 keypoints are 0:Nose, 1:Neck, 2:RShoulder, 3:RElbow, 4:RWrist, 5:LShoulder, 6:LElbow, 7:LWrist, 8:MidHip, 9:RHip, 10:RKnee, 11:RAnkle, 12:LHip, 13:LKnee, 14:LAnkle, 15:REye, 16:LEye, 17:REar, 18:LEar, 19:LBigToe, 20:LSmallToe, 21:LHeel, 22:RBigToe, 23:RSmallToe, 24:RHeel. Figure~\ref{fig:2}B and ~\ref{fig:2}C show a sample from the input-output pairs used to train our network. The video frames of a subject performing 24-form Taiji are processed through the OpenPose network to extract 25 body joint locations, shown as a green overlay on the video frame in Figure~\ref{fig:2}B. Figure~\ref{fig:2}C shows the corresponding foot pressure maps for that pose/frame, measured using insole pressure sensors. For training PressNet and PressNet-Simple, we use the 25 2D joint locations estimated from each video frame as input and the foot pressure data as our target for regression.

\subsection{Foot Pressure}\label{fp}
Foot pressure is collected at 100 fps using a Tekscan F-Scan insole pressure measurement system. Each subject is provided a pair of canvas shoes outfitted with cut-to-fit capacitive pressure measurement insoles. Table~\ref{tab:data_stat} provides the mean and standard deviation of the foot pressure data for each recorded performance ("take") of each subject. The maximum values are clipped when greater than 862 kPa, providing an upper bound based on the technical limits of the pressure measurement sensors. The foot pressure heatmaps are 2-channel images of size $60 \times 21 $ as shown in Figure~\ref{fig:2}C, and are synchronized with the video post-collection.

\subsection{Data Statistics}
To justify apriori the adequacy of our data set for machine learning, we present some observations on the data statistics,  below. Ultimately, our leave-one-subject-out cross validation experimental results provide a quantified validation of our method and dataset used. 

\begin{figure*}[!t] \centering
    \includegraphics[width=1\linewidth]{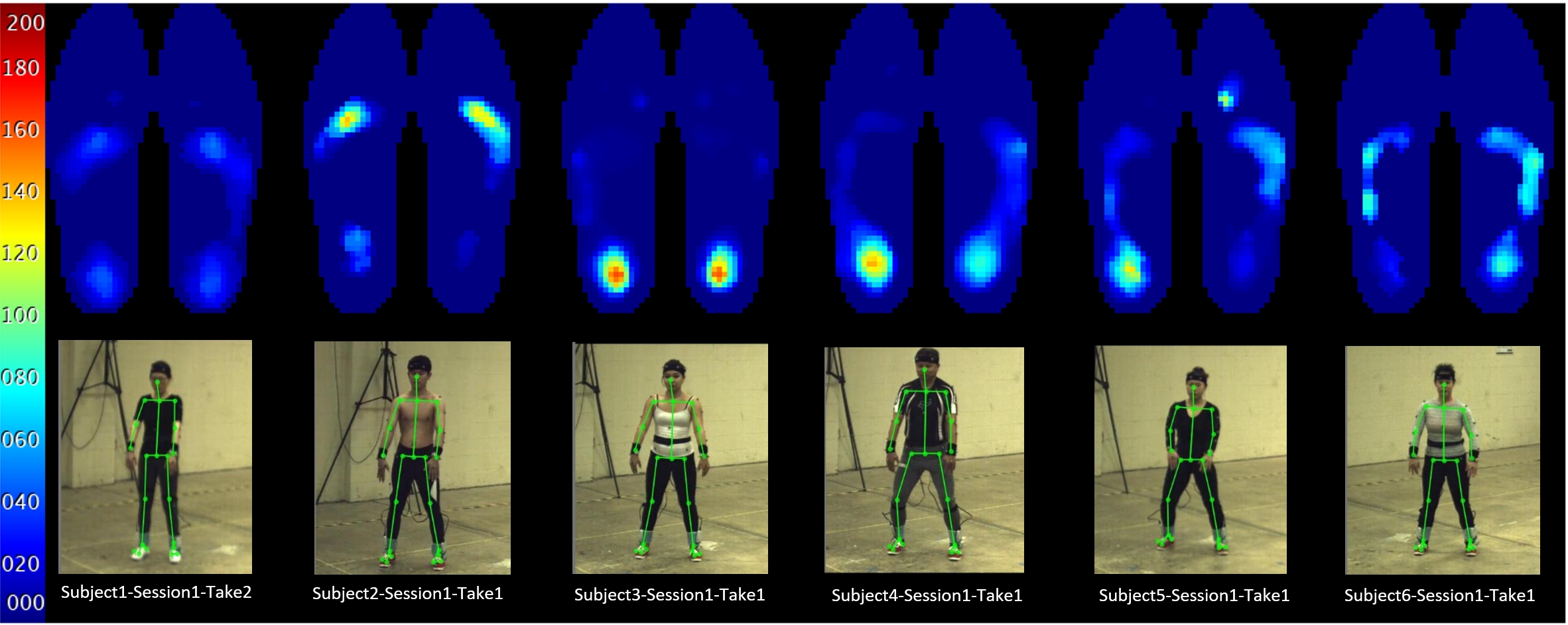} \\ 
    \includegraphics[width=1\linewidth]{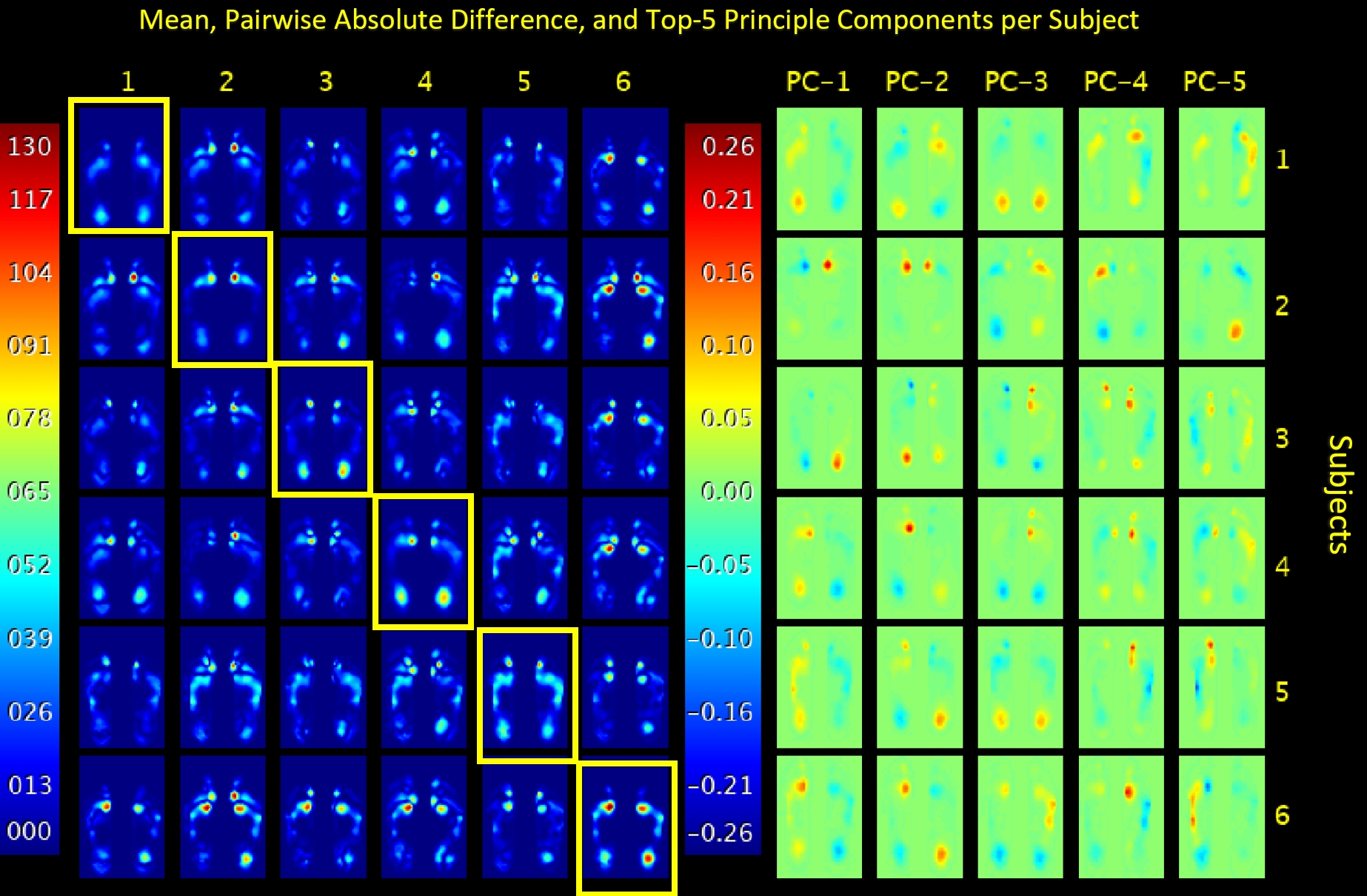}
    \vspace{-10pt}
    \caption{\textbf{Top:} For the same opening pose of 24-Form Taiji, the subjects' foot pressure maps vary greatly.  \textbf{Bottom:} PCA Foot pressure comparison showing inter-subject differences both in magnitude and pressure spatial distribution while maintaining the similar principal components. Left: Pairwise absolute difference between the mean foot pressure across all subjects. Mean pressure is provided on diagonal (yellow bounding box). Right: Top-5 Principal Components of Foot Pressure data per subject.}
    \label{fig:diffPCA}\vspace{-5pt}
\end{figure*}

Table~\ref{tab:data_stat} provides information about the foot pressure dataset. A ``take" refers to a >5 min long continuous motion sequence, while a ``session" refers to a collection of takes. Each subject performs 2 to 3 sessions of 24-form Taiji at an average of 3 takes per session, amounting to a total of 813,050 frames of video-foot pressure paired data. 
We have observed that:
\begin{enumerate} [noitemsep,topsep=0pt] \setlist{nosep}
    \item[(1)] {\bf Subject demographics}: there is diversity in the  subjects in terms of gender, age, mass, height and years of experience in Taiji practice  (Table~\ref{tab:subject_stats}). The range of experience in Taiji of our subjects includes  three amateurs and three professionals.
    \item[(2)] {\bf Body joint} location distributions: Figure~\ref{fig:hist} shows per joint kernel density plots of 2D body joints extracted from the video frames by the OpenPose network. These distributions support the hypothesis that the subjects are statistically similar to one another spatially.    
    \item[(3)] {\bf PCA analysis}:  Figure~\ref{fig:diffPCA} highlights the inter-subject and intra-subject variance of foot pressure data via PCA analysis. The left portion of Figure~\ref{fig:diffPCA} shows the mean foot pressure for each individual subject on the diagonal and the difference of means for pairs of subjects off-diagonal, for all the subjects.  The difference of mean pressure highlights that each subject has a unique pressure distribution relative to other subjects. The right portion of Figure~\ref{fig:diffPCA} highlights the top-5 principal components (PCs) of the foot pressure map data for each subject, with the rows represent individual subjects. From Figure~\ref{fig:diffPCA} we can see that each principal component encodes different types of information (variability in left/right foot pressure, in toe/heel pressure, and so on), and that the collection of top PCs encode similar modes of variation, although not in the exact same order. For example, the 1st principal component of Subject 1 encodes pressure shifts between the left and right foot, whereas the 3rd principal component of Subject 2 encodes that information). 
\end{enumerate}

\section{Our Approach and Motivation}
\begin{figure*}[!t] \centering
    \begin{subfigure}[b]{0.44\textwidth}
         \includegraphics[width=\linewidth]{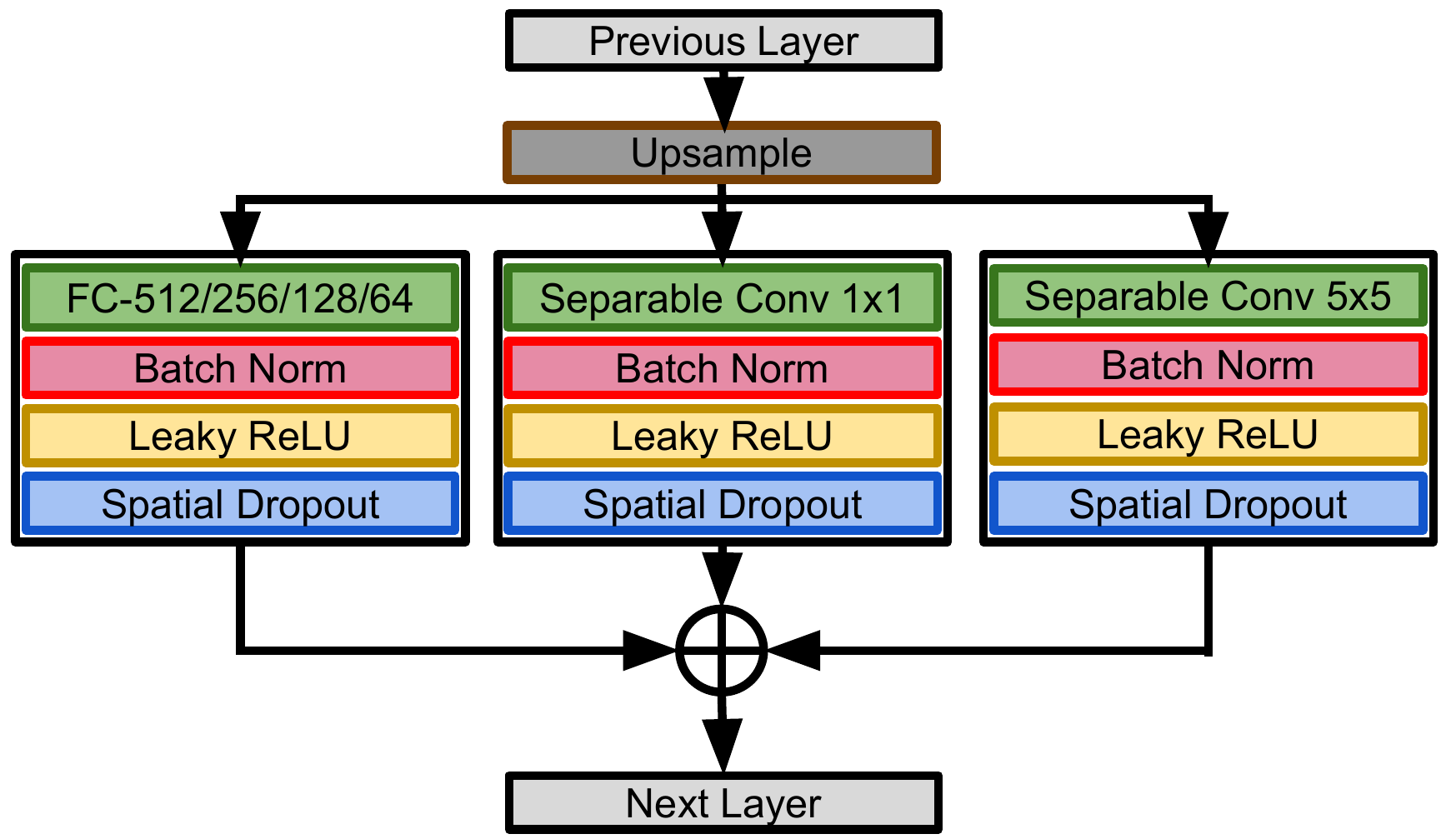}
        \caption{PressNet: Residual Block}
        \label{fig:res2:A}
    \end{subfigure}
    \begin{subfigure}[b]{0.26\textwidth}
         \includegraphics[width=\linewidth]{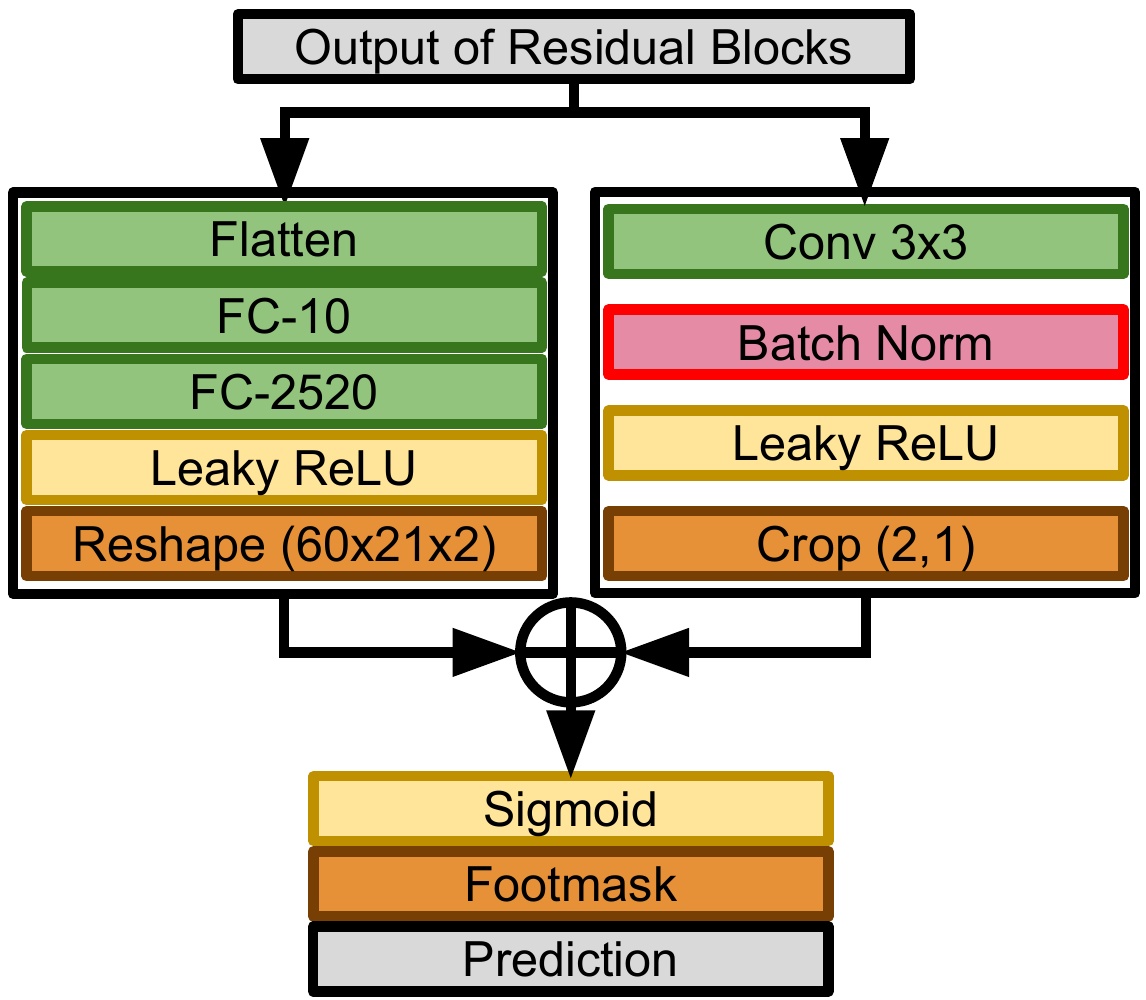} 
        \caption{PressNet: Final Layers}
        \label{fig:res2:B}
    \end{subfigure}
        \begin{subfigure}[b]{.28\textwidth}
          \includegraphics[width=\linewidth]{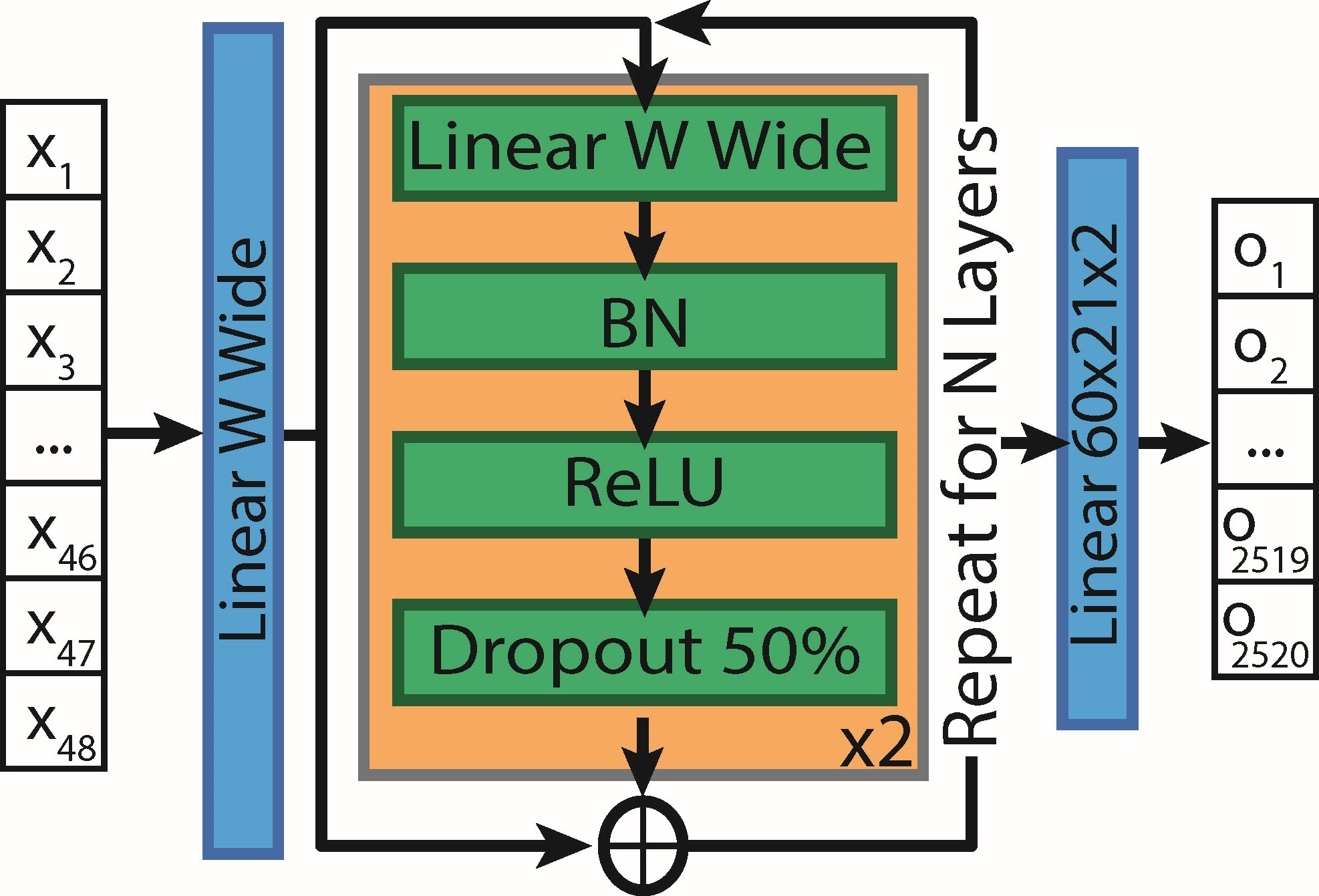} \vspace{.025in}
        \caption{PressNet-Simple}
        \label{fig:PressNet-Simple}
    \end{subfigure} \vspace{-10pt}
    \caption{Our foot pressure regression architectures  have a 48-coordinate input representing 24 2D joint locations (24x2=48) and  a 2520-prexel output representing  60x21 pressure maps for both feet (60x21x2=2520).  \textbf{A:} A residual block, one of the building blocks of PressNet network, upsamples the data and computes features. \textbf{B:} Final set of layers of PressNet include a fully connected layer and a concurrent branch to preserve spatial consistency.  \textbf{C:} The PressNet-Simple network architecture is defined by two hyperparameters: the depth (\# of layers, N) of the network and the width (\# of fully connected nodes, W) of those layers. } 
    \label{fig:res2} \vspace{-15pt}
\end{figure*}

Mapping from a single view of human pose to foot pressure (Figure \ref{fig:1}) is an inherently complex, ill-posed problem. As demonstrated in Figure \ref{fig:diffPCA} (top), for similar poses of different subjects their foot pressure maps may vary greatly. Additional factors such as mass, height, gender and foot shape also lead to variations in foot pressure under the same pose. Thus we formulate our problem as learning a distribution of output foot pressure conditioned on input human pose. For initial simplicity, and lacking  a better model, we assume the conditional distribution of pressure for a given pose is Gaussian, with a mean that can be learned through deep learning regression using MSE loss. Our networks are trained to learn a mapping between pose, encoded as 25 2D joint locations, to the mean of a corresponding foot pressure map intensity distribution. We train two deep residual architectures for regression (Figure~\ref{fig:res2}), PressNet, with 2D convolutional layers, and PressNet-Simple, with only fully connected layers, on data from multiple subjects using a leave-one-subject-out strategy.

\subsection{Data Pre-Processing}\label{dp}
Input body pose data from OpenPose is an array of size $(25\times 3)$. We subtract the hip joint as the center point to remove camera specific offsets during video recording. The hip joint is $(0,0)$ after centering and is removed from the training and testing data sets. Data is normalized per body joint by subtracting the feature's mean and dividing by its standard deviation, leading to a zero-mean, unit variance distribution. 
After pre-processing and normalization, the input array is of size $(24\times 2)$, which is flattened to a 1D vector of size $(48\times1)$ and used as input to our network.

Foot pressure data, which is originally recorded in kilopascals (kPa), has invalid prexels marked as Not a Number ({NaN}) representing regions outside the footmask, indicated as black pixels in Figure~\ref{fig:2}C. These prexels are set to zero since the network library cannot train with NaN values. Any prexel values greater than 862 kilopascals are clipped to 862 to reflect the measurement limitations of the pressure sensor technology. Furthermore, the data is normalized by dividing each prexel by its max intensity value in that location over the entire training set. The left and right normalized foot pressure maps are concatenated as two channels of a resulting ground truth foot pressure heatmap of size $(60 \times 21 \times 2)$, with prexel intensities in the range $[0,1]$. 

In order to study the impact of weight on network training, tests were run on both network architectures with foot pressure data optionally normalized by multiplying by a constant prexel area of $25.8064\text{mm}^2$ derived from the $(5.08\text{mm} \times 5.08\text{mm})$ prexel size and dividing by the mass of the subject reported in Table~\ref{tab:data_splits}. This process generates a foot pressure distribution that is mass and area normalized independent of the subject prior to the max normalization. When this unitless data format was tested it did not make significant improvements in network prediction accuracy. 

While  video  and  foot  pressure are synchronized temporally, the video data is collected at 50 Hz and pressure data at 100Hz. We subsample the video data by 5 and pressure data by 10 to get paired data samples at 10 Hz. This subsampling does not affect the temporal consistency of the data; for a slow-moving activity like Taiji the change of pose and foot pressure within 100 milliseconds is negligible. 

\subsection{Networks and Training}
Our networks are trained to learn the correlation between the pose, encoded as 25 joint locations, and the corresponding foot pressure map intensity distribution. We train a Convolutional Residual architecture, PressNet (Figure~\ref{fig:res2:A} and ~\ref{fig:res2:B}), and deep neural network, PressNet-Simple (Figure~\ref{fig:PressNet-Simple}), to regress foot pressure distributions from a given pose, over data from multiple subjects using a leave-one-subject-out strategy, formulated as a supervised learning (regression) problem.

\subsubsection{PressNet Network}
The design of the PressNet network (Figure~\ref{fig:res2:A} and ~\ref{fig:res2:B}) is initially motivated by the residual generator of the Improved Wasserstein GAN~\cite{gulrajani2017improved}. We do not use any discriminator since ground truth data is available.  We use a generator-inspired architecture because our input is 1D and the output is a 2D heatmap. This design aids in capturing information at different resolutions, acting like a decoder network for feature extraction. The primary aim of this network is to extract features without loss of spatial information across different resolutions. 

The PressNet network is a feed forward Convolutional Neural Network that takes as input a 1D vector of joints and outputs  2D foot pressure. The input layer is a flattened vector of joint coordinates of size $48\times 1$ (24 joints $\times$ 2 coordinates since the mid hip joint is removed), which contains the kinematic information about a pose. The input is processed through a fully connected layer with an output dimension of $6144\times 1$. This output is reshaped into an image of size $4\times3$ with 512 channels. The network contains four residual blocks that perform nearest neighbor upsampling. The first block upsamples the input by $(2,1)$ and the other three upsample by 2. 

The residual block of PressNet, shown in Figure~\ref{fig:res2:A}, has two parallel convolution layers with kernel sizes $5\times5$ and $1\times1$. There is an additional parallel fully connected layer, which takes the upsampled input and returns a flattened array of dimension equal to the output dimension of the residual block. This output is reshaped and added to the output of the other three parallel layers to constitute the output of the block.  The number of channels of each residual block is progressively halved as the resolution doubles, starting at 512 channels and decreasing to 64.

The output of the final residual block is split and sent to a convolutional branch and a fully connected branch. The convolutional branch contains a $3 \times 3 $ normal convolution layer to get a 2 channel output of shape $64\times 24$ and cropped to the size of the foot pressure map ($60\times21\times2$). On the fully connected branch, the activations are run through multiple fully connected layers and then reshaped to the size of the foot pressure map. The sizes of the fully connected layers for PressNet are 10 and 2520 (Figure~\ref{fig:res2:B}, Left).  The output of these branches are added together and then a foot pressure mask is applied to only learn the valid parts of the data. Finally, a sigmoid activation compresses the output to the range [0,1]. The convolutional branch serves to preserve spatial coherence while the fully connected branch has a field of view over the entire prediction. With the combined spatial coherence of the concurrent branch and fully connected layers in every residual convolutional block, PressNet has $\sim$3 million parameters.

All convolutional layers are separable convolution layers that split a kernel into two to perform depth-wise and point-wise convolutions.  Separable convolutions reduce the number of network parameters as well as increase the accuracy~\cite{chollet2017xception}. Batch normalization~\cite{ioffe2015batch} and spatial dropouts~\cite{tompson2015efficient} are applied after every convolution layer. Leaky ReLU~\cite{maas2013rectifier} is used as a common activation function throughout the network, except the output layer.

\subsubsection{PressNet-Simple Network}
The ``simple yet effective'' network of \cite{martinez2017simple}  was originally designed to jointly estimate the unobserved third dimension of a set of 2D body joint coordinates (pose) on a per frame basis.
It is used as a basis for the PressNet-Simple architecture, shown in Figure~\ref{fig:PressNet-Simple}, by adapting  the network architecture to use a modified pose input format and by completely reconfiguring the output format to produce pressure data matrices of each foot. The input pose coordinates are passed through a fully connected layer then through a sequence of N repeated layers. Each of the N layers has two iterations of the sequence: fully connected node (Linear), batch normalization (BN), rectified linear unit (ReLU), and 50\% dropout. The result of each of the N layer sequences is then added to the results from the previous layer sequence (N-1) and finally passed through a 2520 fully connected layer to produce the output foot pressure. The PressNet-Simple architecture  is configured via two hyper-parameters: the depth (\# of layers, N) of the network and the width (\# of fully connected nodes, W) of those layers. For this study, through empirical testing, it was determined that the optimal hyper-parameters are N=4 and W=2560. Because of the sequential nature of this network with fully connected layers, this network architecture does not maintain the spatial coherence that PressNet has established with upsampling and convolutional layers.

\subsubsection{Training Details}
We evaluate our networks on six splits of the dataset,  split by subject in a leave-one-subject-out (LOO) cross-validation. The validation data consists of the last 10 \% of the LOO training data. The goal of this cross-subject validation is to determine how well the network generalizes to an unseen individual. 

PressNet is trained for 35 epochs with a piecewise learning rate starting at $1e^{-4}$ with a drop factor of 0.5 every 10 epochs and a batch size of 32 on a NVIDIA Tesla P100 GPU with 12GB of memory. Data pre-processing is carried out before training as discussed in Section~\ref{dp}. PressNet takes 2.5 hours to train each Leave One Out data split. The problem is formulated as a regression with a sigmoid activation layer as the last activation layer since the output data is in the range $[0,1]$.  A binary footmask having ones marked for valid prexels and zeros marked for invalid prexels (produced by the foot pressure capturing system) is element-wise multiplied in the network. This enables the network to not have to learn the approximate shape of the foot in the course of training and solely learn foot pressure. 
Mean Squared Error (MSE) is used as the loss function along with Adam Optimizer for supervision, as we are learning the distribution of prexels~\cite{bishop2006pattern}. 

PressNet-Simple is trained with an inital learning rate of $1e^{-4}$ for 40 epochs at a batch size of 128 for all splits on a NVIDIA TitanX GPU with 12GB of memory. Data pre-processing is carried out before training as in Section~\ref{dp}. PressNet-Simple takes 1.5 to 2 hours to train on each of the 6 LOO combinations. The learning rate is reduced by 75\% every 7 epochs to ensure a decrease in validation loss with training. Mean Squared Error (MSE) is used as the loss function along with Adam Optimizer for supervision.


\begin{table}[!t] \centering
    \resizebox{1\linewidth}{!}{
    \begin{tabular}{|r||c|c|c|c|} \hline 
    \multicolumn{5}{|c|}{KNN Evaluation of MAE in kPa} \\ \hline
    &  MAE $\downarrow$ & SIM $\uparrow$ & KLD $\downarrow$ & IG $\uparrow$ \\ \hline
    K &  Mean / Std  &  Mean / Std &  Mean / Std &  Mean / Std \\ \hline
    1  & 10.37 / 3.07 & 0.31 / 0.17 & 16.50 / 7.71 & -3.21 / 1.45 \\
    2  & 10.10 / 2.86 & 0.33 / 0.16 & 13.93 / 7.59 & -2.76 / 1.42 \\
    3  & 9.97  / 2.77 & 0.35 / 0.15 & 12.37 / 7.30 & -2.48 / 1.38 \\
    4  & 9.89  / 2.70 & 0.35 / 0.15 & 11.27 / 6.99 & -2.27 / 1.33 \\
    5  & 9.82  / 2.65 & 0.36 / 0.15 & 10.45 / 6.74 & -2.12 / 1.29 \\
    6  & 9.78  / 2.62 & 0.36 / 0.15 & 9.81  / 6.55 & -1.99 / 1.25 \\
    7  & 9.74  / 2.59 & 0.36 / 0.14 & 9.27  / 6.36 & -1.89 / 1.22 \\
    8  & 9.71  / 2.57 & 0.37 / 0.14 & 8.82  / 6.19 & -1.80 / 1.19 \\
    9  & 9.69  / 2.55 & 0.37 / 0.14 & 8.44  / 6.04 & -1.72 / 1.17 \\
    10 & 9.66  / 2.54 & 0.37 / 0.14 & 8.11  / 5.91 & -1.66 / 1.15 \\ \hline
    \end{tabular}
    } 
    \vspace{-10pt} 
    \caption[Table of KNN K Values]{Evaluation of K Nearest Neighbor (KNN) foot pressure results for K=1 through K=10. When K>1 a weighted mean of pressures is computed with closer neighbors   contributing  more  than  further ones. All statistics are in kPa and arrows indicate the  direction of improvement for each metric.}
    \label{tab:knnrange} \vspace{-5pt} 
\end{table}

\section{Results}
\subsection{KNN Baseline\label{sec:KNN}}
K-Nearest Neighbor (KNN) regression~\cite{bishop2006pattern} has been employed as a baseline. Pre-processing is carried out similar to training PressNet and PressNet-Simple. The input pose data is normalized by mean and standard deviation of input, calculated using hip joint centered data by ignoring zero confidence values. The distance metric for the KNN algorithm is calculated as the sum of the Euclidean distances between corresponding joints. This distance $d$ can be represented for two OpenPose human pose detections locations ($a$ and $b$) and $J$ joints by: 
\begin{equation}\label{eq1}
    d(a,b) = \sum_{j \in J}  \left\lVert \left(a_j - b_j\right)\right\rVert_2  
\end{equation}
The KNN algorithm is applied to all six leave-one-subject-out splits. For each pose in the test split, which consists of data from one subject, the corresponding ``nearest'' poses are picked from the training split consisting of data from all other subjects. The foot pressure maps corresponding to these nearest neighbors are combine as a weighted mean for K>1 to generate the prediction for the input pose in the training split. The weighting is chosen so that closer poses contribute more than further poses to the final predicted foot pressure. Table~\ref{tab:knnrange} shows empirical results when testing K values ranging from 1 to 10. KNN with K=2  was selected as a baseline. In our leave-one-subject-out splits, KNN provides a measure of similarity between two poses of different subjects, thus establishing an upper-bound on foot pressure errors inherent in the dataset. Processing time for each of the 6 KNN evaluations is 5-6 hours for pose distance measurements and weighted averaging for K=2 pressure results. However, this is not scalable for large numbers of subjects as available memory puts practical limits on the total number of pose/pressure pairs in the training set. The MAE results achieved are a practical limit of KNN as additional samples will likely not make much improvement in exchanged for the increase in time to calculate a pressure estimate. A comparison of Table~\ref{tab:knnrange} and \ref{tab:DistributionMetrics} shows that the mean and standard deviation for the Similarity (SIM), KL-Divergence (KLD), and Information Gain (IG) evaluation metrics of K=10 have worse performance than either deep learning network.

\begin{table*}[!t] \centering
    \resizebox{0.8\linewidth}{!}{
    \begin{tabular}{|r||ccc|ccc|} \hline 
    \multicolumn{7}{|c|}{KNN, PressNet (PN), and PressNet-Simple (PN-S) Mean Absolute Errors (kPa)}  \\ \hline
     & \multicolumn{3}{c|}{Mean/Std} & \multicolumn{3}{c|}{Min/Median/Max} \\ \hline 
    Subject & KNN  & PN   & PN-S          & KNN  & PN   & PN-S \\ \hline
1 & 8.8 / 2.2 & 8.1 / \textbf{1.3} & \textbf{7.6} / 1.6 & \textbf{3.0} / 8.5 / 20.8 & 5.1 / 7.9 / 18.5 & 3.5 / \textbf{7.3} / \textbf{15.9}  \\ 
2 & 10.7 / 3.1 & \textbf{10.0} / 2.7 & 10.1 / \textbf{2.6} & \textbf{2.5} / 10.5 / 26.4 & 4.5 / \textbf{9.7} / \textbf{22.1} & 4.1 / 9.9 / 22.2  \\ 
3 & 9.4 / 2.8 & 8.7 / 2.3 & \textbf{8.7} / \textbf{2.2} & \textbf{2.3} / 9.1 / 21.5 & 3.7 / \textbf{8.4} / \textbf{18.9} & 3.9 / 8.5 / 19.4  \\ 
4 & 10.2 / 3.2 & 9.2 / \textbf{2.4} & \textbf{8.9} / 2.6 & \textbf{2.7} / 9.9 / 26.1 & 3.6 / 9.0 / 21.6 & 3.1 / \textbf{8.7} / \textbf{21.0}  \\ 
5 & 11.1 / 3.2 & 10.4 / \textbf{2.4} & \textbf{10.0} / 2.5 & \textbf{3.8} / 10.6 / 29.5 & 5.2 / 10.1 / 22.8 & 4.4 / \textbf{9.8} / \textbf{22.0}  \\ 
6 & 10.4 / 2.6 & 10.5 / \textbf{1.9} & \textbf{9.9} / 2.2 & \textbf{3.7} / 10.1 / 22.3 & 5.7 / 10.3 / 20.1 & 5.4 / \textbf{9.6} / \textbf{18.9}  \\ \hline \rowcolor{Gray}
Female Mean & 9.9 / 2.7 & 9.4 / \textbf{2.0} & \textbf{9.0} / 2.1 & \textbf{3.2} / 9.6 / 23.5 & 4.9 / 9.2 / 20.1 & 4.3 / \textbf{8.8} / \textbf{19.0}  \\  \rowcolor{Gray}
Female Std & \textbf{0.9} / 0.4 & 1.0 / 0.4 & 1.0 / \textbf{0.3} & \textbf{0.6} / \textbf{0.8} / 3.5 & 0.8 / 1.0 / \textbf{1.7} & 0.7 / 1.0 / 2.1  \\ \hline \rowcolor{Gray}
Male Mean & 10.5 / 3.2 & 9.6 / \textbf{2.5} & \textbf{9.5} / 2.6 & \textbf{2.6} / 10.2 / 26.3 & 4.0 / 9.3 / 21.9 & 3.6 / \textbf{9.3} / \textbf{21.6}  \\  \rowcolor{Gray}
Male Std & \textbf{0.2} / \textbf{0.0} & 0.4 / 0.1 & 0.6 / 0.0 & \textbf{0.1} / \textbf{0.3} / \textbf{0.1} & 0.4 / 0.4 / 0.2 & 0.5 / 0.6 / 0.6  \\ \hline \rowcolor{Gray}
All Mean & 10.1 / 2.9 & 9.5 / \textbf{2.2} & \textbf{9.2} / 2.3 & \textbf{3.0} / 9.8 / 24.4 & 4.6 / 9.2 / 20.7 & 4.1 / \textbf{9.0} / \textbf{19.9}  \\  \rowcolor{Gray}
All Std & \textbf{0.8} / 0.4 & 0.9 / 0.4 & 0.9 / \textbf{0.4} & \textbf{0.6} / \textbf{0.8} / 3.1 & 0.8 / 0.9 / \textbf{1.6} & 0.7 / 0.9 / 2.1  \\  \hline
    \end{tabular}
    } 
    \vspace{-10pt} 
    \caption[Table of Mean Absolute Errors]{The mean absolute error of foot pressure estimation results for KNN baseline, PressNet (PN), and PressNet-Simple (PN-S) for each subject. Only outputs generated from input poses with 25 valid joints are included in the evaluation. Lowest values are shown in bold.}
    \label{tab:errortable} \vspace{-5pt} 
\end{table*}

\begin{table*}[!t] \centering
    \resizebox{0.8\linewidth}{!}{
    \begin{tabular}{|r||ccc|ccc|} \hline 
    \multicolumn{7}{|c|}{KNN, PressNet (PN), and PressNet-Simple (PN-S) Mean Absolute Errors (kPa) Mass Scaled Training Data}  \\ \hline
     & \multicolumn{3}{c|}{Mean/Std} & \multicolumn{3}{c|}{Min/Median/Max} \\ \hline
    Subject & KNN  & PN   & PN-S          & KNN  & PN   & PN-S \\ \hline
1 & 8.8 / 2.2 & 7.5 / 1.6 & \textbf{6.6} / \textbf{1.3} & \textbf{3.0} / 8.5 / 20.8 & 3.8 / 7.3 / 17.2 & 3.8 / \textbf{6.4} / \textbf{13.7}  \\ 
2 & 10.7 / 3.1 & \textbf{10.2} / 2.6 & 10.3 / \textbf{2.6} & \textbf{2.5} / 10.5 / 26.4 & 4.7 / \textbf{10.0} / \textbf{22.4} & 3.7 / 10.1 / 23.7  \\ 
3 & 9.4 / 2.8 & \textbf{8.4} / \textbf{2.2} & 8.7 / 2.2 & \textbf{2.3} / 9.1 / 21.5 & 3.5 / \textbf{8.1} / \textbf{18.2} & 3.3 / 8.5 / 19.7  \\ 
4 & 10.2 / 3.2 & 9.3 / \textbf{2.4} & \textbf{9.1} / 2.6 & \textbf{2.7} / 9.9 / 26.1 & 3.4 / 9.0 / 22.7 & 3.4 / \textbf{9.0} / \textbf{22.2}  \\ 
5 & 11.1 / 3.2 & 10.1 / \textbf{2.4} & \textbf{9.9} / 2.5 & \textbf{3.8} / 10.6 / 29.5 & 4.3 / 9.8 / 21.3 & 4.0 / \textbf{9.6} / \textbf{21.1}  \\ 
6 & 10.4 / 2.6 & 10.6 / 2.1 & \textbf{9.6} / \textbf{2.0} & \textbf{3.7} / 10.1 / 22.3 & 5.9 / 10.4 / 20.2 & 5.3 / \textbf{9.3} / \textbf{19.0}  \\ \hline \rowcolor{Gray}
Female Mean & 9.9 / 2.7 & 9.2 / 2.1 & \textbf{8.7} / \textbf{2.0} & \textbf{3.2} / 9.6 / 23.5 & 4.4 / 8.9 / 19.2 & 4.1 / \textbf{8.5} / \textbf{18.4}  \\ \rowcolor{Gray}
Female Std & \textbf{0.9} / 0.4 & 1.3 / \textbf{0.3} & 1.3 / 0.4 & \textbf{0.6} / \textbf{0.8} / 3.5 & 0.9 / 1.2 / \textbf{1.6} & 0.8 / 1.3 / 2.8  \\ \hline \rowcolor{Gray}
Male Mean & 10.5 / 3.2 & 9.7 / \textbf{2.5} & \textbf{9.7} / 2.6 & \textbf{2.6} / 10.2 / 26.3 & 4.0 / \textbf{9.5} / \textbf{22.5} & 3.5 / 9.5 / 22.9  \\ \rowcolor{Gray}
Male Std & \textbf{0.2} / 0.0 & 0.5 / 0.1 & 0.6 / \textbf{0.0} & \textbf{0.1} / \textbf{0.3} / \textbf{0.1} & 0.6 / 0.5 / 0.2 & 0.1 / 0.5 / 0.8  \\ \hline \rowcolor{Gray}
All Mean & 10.1 / 2.9 & 9.4 / 2.2 & \textbf{9.0} / \textbf{2.2} & \textbf{3.0} / 9.8 / 24.4 & 4.3 / 9.1 / 20.3 & 3.9 / \textbf{8.8} / \textbf{19.9}  \\  \rowcolor{Gray}
All Std & \textbf{0.8} / 0.4 & 1.1 / \textbf{0.3} & 1.2 / 0.5 & \textbf{0.6} / \textbf{0.8} / 3.1 & 0.9 / 1.1 / \textbf{2.1} & 0.7 / 1.2 / 3.2  \\ \hline  
    \end{tabular}
    } 
    \vspace{-10pt} 
    \caption[Table of Mean Absolute Errors Scaled]{The mean absolute error of foot pressure estimation results for KNN baseline, PressNet (PN), and PressNet-Simple (PN-S) for each subject using mass scaled training data. The results are generated from training data scaled by subject mass and rescaled back after prediction. Only outputs generated from input poses with 25 valid joints are included in the evaluation. Lowest values are shown in bold.}
    \label{tab:errortableScaled} \vspace{-5pt} 
\end{table*}

\subsection{Quantitative Evaluation of Network Output}
Three metrics for quantitative evaluation of our networks have been used:
\begin{enumerate}[noitemsep,topsep=0pt]
\item  Mean Absolute Error of Estimated Foot Pressure maps (kPa) as compared to ground truth pressure;
\item  Similarity, KL Divergence, and Information Gain metrics that compare the spatial distributions of pressure maps independent of pressure intensity; and
\item  Euclidean ($\ell_2$) distance of Center of Pressure (mm) as compared to CoP calculated directly from ground truth foot pressure.
\end{enumerate}

\subsubsection{Mean Absolute Error of Predicted Foot Pressure} \label{MAE}
Mean absolute error $ E $ is used to quantify the difference between ground truth foot pressure $ Y $ and predicted foot pressure $\hat{Y} $ over $ N $ foot pressure prexels as:
\begin{equation}
    E = \frac{1}{|N|} \sum_{n \in N} \left|Y_n - \hat{Y_n}\right|
\end{equation}
Mean across all the cross-validation splits is taken for our final accuracy rates. 

\begin{table*}[!t] \centering
    \resizebox{0.8\linewidth}{!}{
    \begin{tabular}{|r||ccc|ccc|} \hline 
    \multicolumn{7}{|c|}{KNN, PressNet (PN), and PressNet-Simple (PN-S) Mean Absolute Errors (kPa) Weight Normalized}  \\ \hline
     & \multicolumn{3}{c|}{Mean/Std} & \multicolumn{3}{c|}{Min/Median/Max} \\ \hline
    Subject & KNN  & PN   & PN-S          & KNN  & PN   & PN-S \\ \hline
1 & 10.5 / 2.6 & 9.7 / \textbf{1.6} & \textbf{9.0} / 1.9 & \textbf{3.6} / 10.2 / 24.9 & 6.1 / 9.5 / 22.1 & 4.1 / \textbf{8.7} / \textbf{19.1}  \\ 
2 & 10.0 / 2.9 & \textbf{9.3} / 2.5 & 9.5 / \textbf{2.4} & \textbf{2.4} / 9.8 / 24.7 & 4.2 / \textbf{9.1} / \textbf{20.7} & 3.9 / 9.3 / 20.7  \\ 
3 & 9.2 / 2.8 & 8.6 / 2.2 & \textbf{8.5} / \textbf{2.1} & \textbf{2.3} / 9.0 / 21.2 & 3.6 / \textbf{8.2} / \textbf{18.6} & 3.8 / 8.3 / 19.0  \\ 
4 & 8.3 / 2.6 & 7.4 / \textbf{2.0} & \textbf{7.2} / 2.1 & \textbf{2.2} / 8.0 / 21.2 & 2.9 / 7.2 / 17.5 & 2.5 / \textbf{7.1} / \textbf{17.0}  \\ 
5 & 11.5 / 3.3 & 10.9 / \textbf{2.5} & \textbf{10.4} / 2.6 & \textbf{3.9} / 11.1 / 30.7 & 5.4 / 10.5 / 23.7 & 4.6 / \textbf{10.2} / \textbf{22.8}  \\ 
6 & 11.8 / 3.0 & 11.9 / \textbf{2.2} & \textbf{11.2} / 2.5 & \textbf{4.3} / 11.5 / 25.3 & 6.5 / 11.7 / 22.8 & 6.1 / \textbf{10.9} / \textbf{21.5}  \\ \hline \rowcolor{Gray}
Female Mean & 10.8 / 2.9 & 10.3 / \textbf{2.1} & \textbf{9.8} / 2.3 & \textbf{3.5} / 10.4 / 25.5 & 5.4 / 10.0 / 21.8 & 4.7 / \textbf{9.5} / \textbf{20.6}  \\  \rowcolor{Gray}
Female Std & \textbf{1.0} / \textbf{0.3} & 1.2 / 0.3 & 1.1 / 0.3 & \textbf{0.7} / \textbf{1.0} / 3.4 & 1.1 / 1.3 / 1.9 & 0.9 / 1.1 / \textbf{1.6}  \\ \hline \rowcolor{Gray}
Male Mean & 9.2 / 2.8 & 8.4 / \textbf{2.2} & \textbf{8.3} / 2.3 & \textbf{2.3} / 8.9 / 22.9 & 3.5 / 8.2 / 19.1 & 3.2 / \textbf{8.2} / \textbf{18.9}  \\  \rowcolor{Gray}
Male Std & \textbf{0.9} / \textbf{0.2} & 1.0 / 0.3 & 1.1 / 0.2 & \textbf{0.1} / \textbf{0.9} / 1.8 & 0.6 / 0.9 / \textbf{1.6} & 0.7 / 1.1 / 1.9  \\ \hline \rowcolor{Gray}
All Mean & 10.2 / 2.9 & 9.6 / \textbf{2.2} & \textbf{9.3} / 2.3 & \textbf{3.1} / 9.9 / 24.6 & 4.8 / 9.4 / 20.9 & 4.2 / \textbf{9.1} / \textbf{20.0}  \\  \rowcolor{Gray}
All Std & \textbf{1.2} / \textbf{0.3} & 1.5 / 0.3 & 1.3 / 0.3 & \textbf{0.8} / \textbf{1.2} / 3.2 & 1.3 / 1.5 / 2.2 & 1.1 / 1.2 / \textbf{1.9}  \\  \hline
    \end{tabular}
    } 
    \vspace{-10pt} 
    \caption[Table of Mean Absolute Errors Scaled]{The subject mass (average mass / subject mass) normalized mean absolute error of foot pressure estimation results for KNN baseline, PressNet (PN), and PressNet-Simple (PN-S) for each subject. Only outputs generated from input poses with 25 valid joints are included in the evaluation. Lowest values are shown in bold.}
    \label{tab:errortableMass} \vspace{-5pt} 
    
\end{table*}
\begin{table*}[!t] \centering
    \resizebox{0.8\linewidth}{!}{
    \begin{tabular}{|r||ccc|ccc|} \hline 
    \multicolumn{7}{|c|}{KNN, PressNet (PN), and PressNet-Simple (PN-S) Mean Absolute Errors (kPa) Height Normalized}  \\ \hline
     & \multicolumn{3}{c|}{Mean/Std} & \multicolumn{3}{c|}{Min/Median/Max} \\ \hline
    Subject & KNN  & PN   & PN-S          & KNN  & PN   & PN-S \\ \hline
1 & 8.9 / 2.2 & 8.2 / \textbf{1.4} & \textbf{7.7} / 1.6 & \textbf{3.1} / 8.6 / 21.1 & 5.2 / 8.0 / 18.7 & 3.5 / \textbf{7.4} / \textbf{16.1}  \\ 
2 & 10.1 / 2.9 & \textbf{9.4} / 2.5 & 9.5 / \textbf{2.5} & \textbf{2.4} / 9.9 / 24.9 & 4.2 / \textbf{9.2} / \textbf{20.8} & 3.9 / 9.3 / 20.9  \\ 
3 & 9.5 / 2.9 & 8.8 / 2.3 & \textbf{8.8} / \textbf{2.2} & \textbf{2.4} / 9.2 / 21.8 & 3.7 / \textbf{8.5} / \textbf{19.2} & 3.9 / 8.6 / 19.6  \\ 
4 & 9.8 / 3.0 & 8.8 / \textbf{2.3} & \textbf{8.5} / 2.4 & \textbf{2.6} / 9.5 / 24.9 & 3.4 / 8.5 / 20.6 & 2.9 / \textbf{8.3} / \textbf{20.0}  \\ 
5 & 11.5 / 3.3 & 10.8 / \textbf{2.5} & \textbf{10.4} / 2.6 & \textbf{3.9} / 11.1 / 30.6 & 5.4 / 10.5 / 23.7 & 4.6 / \textbf{10.1} / \textbf{22.8}  \\ 
6 & 10.9 / 2.8 & 11.0 / \textbf{2.0} & \textbf{10.4} / 2.3 & \textbf{3.9} / 10.6 / 23.4 & 6.0 / 10.8 / 21.1 & 5.7 / \textbf{10.1} / \textbf{19.9}  \\ \hline \rowcolor{Gray}
Female Mean & 10.2 / 2.8 & 9.7 / \textbf{2.0} & \textbf{9.3} / 2.2 & \textbf{3.3} / 9.9 / 24.2 & 5.1 / 9.5 / 20.7 & 4.4 / \textbf{9.0} / \textbf{19.6}  \\  \rowcolor{Gray}
Female Std & \textbf{1.0} / 0.4 & 1.2 / 0.4 & 1.2 / \textbf{0.4} & \textbf{0.7} / \textbf{1.0} / 3.8 & 0.8 / 1.2 / \textbf{2.0} & 0.8 / 1.2 / 2.4  \\ \hline \rowcolor{Gray}
Male Mean & 9.9 / 3.0 & 9.1 / \textbf{2.4} & \textbf{9.0} / 2.4 & \textbf{2.5} / 9.7 / 24.9 & 3.8 / 8.8 / 20.7 & 3.4 / \textbf{8.8} / \textbf{20.4}  \\  \rowcolor{Gray}
Male Std & \textbf{0.2} / 0.0 & 0.3 / 0.1 & 0.5 / \textbf{0.0} & \textbf{0.1} / \textbf{0.2} / \textbf{0.0} & 0.4 / 0.3 / 0.1 & 0.5 / 0.5 / 0.4  \\ \hline \rowcolor{Gray}
All Mean & 10.1 / 2.9 & 9.5 / \textbf{2.2} & \textbf{9.2} / 2.3 & \textbf{3.0} / 9.8 / 24.4 & 4.7 / 9.3 / 20.7 & 4.1 / \textbf{9.0} / \textbf{19.9}  \\  \rowcolor{Gray}
All Std & \textbf{0.9} / 0.3 & 1.1 / 0.4 & 1.0 / \textbf{0.3} & \textbf{0.7} / \textbf{0.8} / 3.1 & 0.9 / 1.1 / \textbf{1.6} & 0.9 / 1.0 / 2.0  \\  \hline
    \end{tabular}
    } 
    \vspace{-10pt} 
    \caption[Table of Mean Absolute Errors Scaled]{The subject height (average height / subject height) normalized mean absolute error of foot pressure estimation results for KNN baseline, PressNet (PN), and PressNet-Simple (PN-S) for each subject. Only outputs generated from input poses with 25 valid joints are included in the evaluation. Lowest values are shown in bold.}
    \label{tab:errortableHeight} \vspace{-5pt} 
    
\end{table*}
\begin{table*}[!t] \centering
    \resizebox{0.8\linewidth}{!}{
    \begin{tabular}{|r||ccc|ccc|} \hline 
    \multicolumn{7}{|c|}{KNN, PressNet (PN), and PressNet-Simple (PN-S) Mean Absolute Errors (kPa) Weight and Height Normalized}  \\ \hline
     & \multicolumn{3}{c|}{Mean/Std} & \multicolumn{3}{c|}{Min/Median/Max} \\ \hline
    Subject & KNN  & PN   & PN-S          & KNN  & PN   & PN-S \\ \hline
1 & 10.6 / 2.6 & 9.8 / \textbf{1.6} & \textbf{9.2} / 1.9 & \textbf{3.7} / 10.3 / 25.2 & 6.2 / 9.6 / 22.4 & 4.2 / \textbf{8.8} / \textbf{19.3}  \\ 
2 & 9.4 / 2.8 & \textbf{8.8} / 2.3 & 8.9 / \textbf{2.3} & \textbf{2.2} / 9.3 / 23.3 & 3.9 / \textbf{8.6} / \textbf{19.5} & 3.7 / 8.7 / 19.5  \\ 
3 & 9.3 / 2.8 & 8.7 / 2.3 & \textbf{8.7} / \textbf{2.2} & \textbf{2.3} / 9.1 / 21.4 & 3.7 / \textbf{8.3} / \textbf{18.8} & 3.9 / 8.4 / 19.3  \\ 
4 & 7.9 / 2.5 & 7.1 / \textbf{1.9} & \textbf{6.9} / 2.0 & \textbf{2.1} / 7.7 / 20.2 & 2.8 / 6.9 / 16.7 & 2.4 / \textbf{6.7} / \textbf{16.2}  \\ 
5 & 11.9 / 3.5 & 11.3 / \textbf{2.6} & \textbf{10.9} / 2.7 & \textbf{4.1} / 11.5 / 31.8 & 5.7 / 10.9 / 24.6 & 4.8 / \textbf{10.5} / \textbf{23.7}  \\ 
6 & 12.4 / 3.2 & 12.5 / \textbf{2.3} & \textbf{11.8} / 2.6 & \textbf{4.5} / 12.0 / 26.6 & 6.8 / 12.3 / 24.0 & 6.4 / \textbf{11.5} / \textbf{22.6}  \\ \hline \rowcolor{Gray}
Female Mean & 11.1 / 3.0 & 10.6 / \textbf{2.2} & \textbf{10.1} / 2.3 & \textbf{3.6} / 10.7 / 26.3 & 5.6 / 10.3 / 22.5 & 4.8 / \textbf{9.8} / \textbf{21.2}  \\  \rowcolor{Gray}
Female Std & \textbf{1.2} / \textbf{0.3} & 1.4 / 0.3 & 1.3 / 0.3 & \textbf{0.8} / \textbf{1.1} / 3.7 & 1.2 / 1.5 / 2.2 & 1.0 / 1.2 / \textbf{2.0}  \\ \hline \rowcolor{Gray}
Male Mean & 8.7 / 2.6 & 7.9 / \textbf{2.1} & \textbf{7.9} / 2.1 & \textbf{2.2} / 8.5 / 21.7 & 3.4 / 7.7 / 18.1 & 3.0 / \textbf{7.7} / \textbf{17.9}  \\  \rowcolor{Gray}
Male Std & \textbf{0.8} / \textbf{0.2} & 0.9 / 0.2 & 1.0 / 0.2 & \textbf{0.1} / \textbf{0.8} / 1.6 & 0.6 / 0.8 / \textbf{1.4} & 0.6 / 1.0 / 1.7  \\ \hline \rowcolor{Gray}
All Mean & 10.3 / 2.9 & 9.7 / \textbf{2.2} & \textbf{9.4} / 2.3 & \textbf{3.1} / 10.0 / 24.7 & 4.8 / 9.4 / 21.0 & 4.2 / \textbf{9.1} / \textbf{20.1}  \\  \rowcolor{Gray}
All Std & \textbf{1.6} / 0.3 & 1.8 / 0.3 & 1.6 / \textbf{0.3} & \textbf{1.0} / \textbf{1.5} / 3.8 & 1.5 / 1.8 / 2.9 & 1.2 / 1.5 / \textbf{2.4}  \\  \hline
    \end{tabular}
    } 
    \vspace{-10pt} 
    \caption[Table of Mean Absolute Errors Scaled]{The subject mass (average mass / subject mass) and subject height (average height / subject height) normalized mean absolute error of foot pressure estimation results for KNN baseline, PressNet (PN), and PressNet-Simple (PN-S) for each subject. Only outputs generated from input poses with 25 valid joints are included in the evaluation. Lowest values are shown in bold.}
    \label{tab:errortableMassHeight} \vspace{-5pt} 
\end{table*}

\begin{table*}[!t] \centering 
\resizebox{.9\linewidth}{!}{
\begin{tabular}{|r||ccc|ccc|ccc|} \hline
    \multicolumn{10}{|c|}{Spatial Distribution Error Metrics of KNN, PressNet (PN), and PressNet-Simple (PN-S)} \\  \hline
Sub & \multicolumn{3}{c|}{KNN (Mean / Std)} & \multicolumn{3}{c|}{PN (Mean / Std)} & \multicolumn{3}{c|}{PN-S (Mean / Std)} \\ \hline
\# & SIM & KLD & IG & SIM & KLD & IG & SIM & KLD & IG \\ \hline
1 & 0.30 / 0.15 & 17.02 / 6.67 & -4.35 / 1.89 & \textbf{0.44} / 0.10 & 6.99 / 4.18 & -1.93 / 1.13 & 0.44 / \textbf{0.10} & \textbf{2.79} / \textbf{1.99} & \textbf{-0.74} / \textbf{0.50} \\ 
2 & 0.22 / 0.15 & 14.34 / 8.97 & -1.23 / 0.79 & 0.22 / \textbf{0.15} & 11.84 / 8.25 & -0.91 / 0.64 & \textbf{0.23} / 0.15 & \textbf{5.21} / \textbf{5.38} & \textbf{-0.41} / \textbf{0.41} \\ 
3 & 0.31 / 0.16 & 14.80 / 7.64 & -2.76 / 1.45 & \textbf{0.41} / \textbf{0.11} & 3.87 / 3.40 & -0.84 / 0.80 & 0.39 / 0.13 & \textbf{2.95} / \textbf{2.70} & \textbf{-0.54} / \textbf{0.47} \\ 
4 & 0.32 / 0.16 & 14.42 / 7.79 & -3.70 / 1.89 & \textbf{0.48} / 0.11 & 3.71 / 2.71 & -1.21 / 0.95 & 0.45 / \textbf{0.11} & \textbf{1.88} / \textbf{1.39} & \textbf{-0.52} / \textbf{0.45} \\ 
5 & 0.40 / 0.17 & 12.19 / 6.64 & -2.65 / 1.22 & \textbf{0.49} / \textbf{0.13} & 3.23 / 2.66 & -0.76 / 0.58 & 0.47 / 0.13 & \textbf{1.94} / \textbf{1.52} & \textbf{-0.44} / \textbf{0.37} \\ 
6 & 0.38 / 0.16 & 10.48 / 7.67 & -1.56 / 0.97 & 0.44 / \textbf{0.11} & 2.15 / 2.02 & -0.35 / 0.35 & \textbf{0.45} / 0.12 & \textbf{1.54} / \textbf{1.35} & \textbf{-0.22} / \textbf{0.23} \\ \hline \rowcolor{Gray}
Mean & 0.31 / 0.16 & 14.35 / 7.35 & -2.71 / 1.35 & 0.39 / 0.13 & 7.40 / 4.67 & -1.35 / 0.89 & \textbf{0.39} / \textbf{0.12} & \textbf{3.01} / \textbf{2.54} & \textbf{-0.53} / \textbf{0.42} \\  \rowcolor{Gray}
Std & \textbf{0.06} / \textbf{0.01} & 2.07 / \textbf{0.78} & 1.10 / 0.42 & 0.09 / 0.01 & 3.28 / 2.07 & 0.49 / 0.25 & 0.08 / 0.02 & \textbf{1.22} / 1.41 & \textbf{0.15} / \textbf{0.09} \\ \hline
\end{tabular}
}
\vspace{-10pt} 
\caption{Spatial Distribution Error analysis (Mean/Std) of foot pressure results for KNN baseline, PressNet (PN), and PressNet-Simple (PN-S) for each subject. The metrics used are Similarity (SIM), KL Divergence (KLD), Information Gain (IG). Only outputs generated from input poses with 25 valid joints are included in the evaluation. Best values are shown in bold. Best is defined as closer to 1 for SIM, closer to 0 for KLD and closer to 0 for IG.}
\label{tab:DistributionMetrics} \vspace{-5pt} 
\end{table*}

\par Table~\ref{tab:errortable} shows the mean absolute errors (MAE) of predicted foot pressure for each data split of KNN baseline, PressNet, and PressNet-Simple. For each pressure prediction method, only frames that have 25 detect joints are included in the analysis to ensure that the MAE reflects each prediction method's effectiveness and is not negatively impacted by the quality of the pose detection method. For the nearest neighbor classifier, Subject 1 is the best performing subject with a MAE of 8.8 kPa while Subject 5 has the largest error of 11.1 kPa and the MAE over all subjects is 10.1 kPa. For PressNet, Subject 1 has the best results with a MAE of 8.1 kPa, the worst individual is Subject 6 with a MAE of 10.5 kPa, and the MAE over all subjects is 9.5 kPa. For PressNet-Simple, Subject 1 is the best with a MAE of 7.6 kPa, the worst individual is Subject 2 with a MAE of 10.1 kPa, and the MAE over all subjects is 9.2 kPa. For comparison, there is a 3 kPa measurement threshold of the foot pressure recording devices, as mentioned in Section~\ref{fp}.  An interesting observation from Table~\ref{tab:errortable} is that KNN consistently has the smallest min MAE for all subjects while median and max error of MAE is smallest primarily with PressNet-Simple. This is an indication that KNN has a larger standard deviation but the error distribution encompasses more accurate results for some frames.

Tables~\ref{tab:errortableMass}, ~\ref{tab:errortableHeight}, and ~\ref{tab:errortableMassHeight} use the same mean absolute error (MAE) data in Table~\ref{tab:errortable} but are scaled by various normalization factors to provide insight into the inter-subject performance similarities and differences. Subject mass, subject height and both height and mass were used in normalizations to determine if any inter-subject inconsistencies exist when normalizing for weight and/or height. Table~\ref{tab:errortableMass} scales the MAE relative to the subject mass by calculating the ratio of average mass divided by subject mass resulting in kPa values normalized by subject mass, to provide a  mass-adjusted comparison of subjects. With this scaling, it can be seen that for all tested methods, Subject 4 performs the best while Subject 6 performs the worst. Table~\ref{tab:errortableHeight} applies the same normalization approach as Table~\ref{tab:errortableMass} but using subject height rather than mass resulting in a height-normalized result. Table~\ref{tab:errortableMassHeight} builds upon Table~\ref{tab:errortableMass} and Table~\ref{tab:errortableHeight} by applying both normalization methods to provide results normalized for subject weight and height. Through review of Tables~\ref{tab:errortable}, ~\ref{tab:errortableMass}, ~\ref{tab:errortableHeight}, and ~\ref{tab:errortableMassHeight} it does not appear that subject weight, subject height, or both combined raise any concerns about inter-subject relationships. 

\begin{table*}[!t] \centering 
\resizebox{1\linewidth}{!}{
\begin{tabular}{|r||ccc|c||ccc|c||ccc|c|} \hline
    \multicolumn{13}{|c|}{CoP X, Y, \& L2 Error of KNN, PressNet (PN), and PressNet-Simple (PN-S) in mm} \\  \hline
Sub & \multicolumn{4}{c||}{KNN (Mean / Std)} & \multicolumn{4}{c||}{PN (Mean / Std)} & \multicolumn{4}{c|}{PN-S (Mean / Std)} \\ \hline
\# & $\Delta X$ & $\Delta Y$ & D & $D_{mean}$ & $\Delta X$ & $\Delta Y$ & D & $D_{mean}$ & $\Delta X$ & $\Delta Y$ & D & $D_{mean}$ \\  \hline
1 & -7.0 / 78.9 & -1.0 / 72.0 & 7.1 & 64.3 / 77.2 & 2.9 / 54.4 & 0.3 / 50.0 & 3.0 & \textbf{46.3} / \textbf{51.7} & -3.6 / 58.1 & 1.0 / 53.0 & 3.7 & 48.3 / 55.9 \\
2 & -14.4 / 85.2 & 9.4 / 110.9 & 17.2 & 98.6 / 94.1 & -9.9 / 82.3 & 14.4 / 111.7 & 17.5 & 97.5 / 93.8 & -13.6 / 84.2 & 10.9 / 102.4 & 17.4 & \textbf{95.0} / \textbf{87.7} \\
3 & 17.2 / 72.5 & 7.4 / 77.2 & 18.7 & 71.9 / 76.2 & 13.0 / 50.8 & 3.0 / 53.9 & 13.3 & 54.1 / 49.5 & 14.8 / 49.4 & 2.1 / 51.7 & 15.0 & \textbf{51.8} / \textbf{48.7} \\
4 & 5.9 / 106.7 & 2.7 / 83.1 & 6.5 & 92.4 / 97.5 & 13.0 / 64.5 & 1.5 / 57.4 & 13.1 & 66.5 / 55.5 & 11.8 / 61.5 & 1.7 / 50.1 & 11.9 & \textbf{61.1} / \textbf{51.0} \\
5 & -3.2 / 68.0 & 9.6 / 80.5 & 10.1 & 75.1 / 74.3 & 0.0 / 54.7 & 4.2 / 66.9 & 4.2 & 65.0 / 56.8 & -0.1 / 50.1 & 13.7 / 54.7 & 13.7 & \textbf{55.8} / \textbf{50.6} \\
6 & -1.0 / 66.1 & -8.3 / 70.4 & 8.4 & 70.4 / 66.7 & -2.5 / 50.2 & -13.9 / 54.6 & 14.1 & 58.0 / 48.3 & 1.5 / 47.5 & -14.8 / 46.1 & 14.9 & \textbf{52.6} / \textbf{42.8} \\ \hline \rowcolor{Gray}
Mean & -0.4 / 79.6 & 3.3 / 82.3 & 11.3 & 78.8 / 81.0 & 2.8 / 59.5 & 1.6 / 65.8 & 10.9 & 64.6 / 59.3 & 1.8 / 58.5 & 2.4 / 59.7 & 12.8 & \textbf{60.8} / \textbf{56.1} \\ \rowcolor{Gray}
Std & 10.0 / 13.7 & 6.4 / 13.5 & 4.8 & \textbf{12.4} / \textbf{11.0} & 8.2 / 11.2 & 8.3 / 21.2 & 5.4 & 16.2 / 15.7 & 9.5 / 12.5 & 9.1 / 19.3 & 4.4 & 15.8 / 14.6 \\ \hline
\end{tabular}
}
\vspace{-10pt} 
\caption{Mean and standard deviation (Mean/Std) of offset errors ($\Delta X, \Delta Y$) in X and Y dimensions and L2 distance $D=\sqrt{(\Delta x)^2 + (\Delta y)^2}$. $D_{mean}$ is the mean distance of ground truth CoP to  the CoP of  foot pressure maps predicted by KNN, PressNet, and PressNet-Simple, respectively. See Figure~\ref{fig:coperrorbar} for a graphical representation and Figure~\ref{fig:6a} for a display in 2D scatter points.}
\label{tab:CoPL2Error} \vspace{-5pt} 
\end{table*}

\begin{table*}[!t] \centering 
\resizebox{1\linewidth}{!}{
\begin{tabular}{|r||ccc|c||ccc|c||ccc|c|} \hline
    \multicolumn{13}{|c|}{CoP X, Y, \& L2 Error of KNN K=2, K=5, and K=10 in mm} \\  \hline
Sub & \multicolumn{4}{c||}{K=2 (Mean / Std)} & \multicolumn{4}{c||}{K=5 (Mean / Std)} & \multicolumn{4}{c|}{K=10 (Mean / Std)} \\ \hline
\# & $\Delta X$ & $\Delta Y$ & D & $D_{mean}$ & $\Delta X$ & $\Delta Y$ & D & $D_{mean}$ & $\Delta X$ & $\Delta Y$ & D & $D_{mean}$ \\  \hline
1 & -7.0 / 78.9 & -1.0 / 72.0 & 7.1 & 64.3 / 77.2 & -7.6 / 70.7 & -2.3 / 65.2 & 8.0 & 57.7 / 69.7 & -7.6 / 65.2 & -3.2 / 62.3 & 8.2 & \textbf{53.6} / \textbf{65.8} \\
2 & -14.4 / 85.2 & 9.4 / 110.9 & 17.2 & 98.6 / 94.1 & -15.1 / 80.6 & 9.4 / 106.3 & 17.8 & 94.8 / 89.2 & -15.6 / 79.3 & 10.0 / 104.7 & 18.6 & \textbf{94.3} / \textbf{87.0} \\
3 & 17.2 / 72.5 & 7.4 / 77.2 & 18.7 & 71.9 / 76.2 & 15.5 / 63.1 & 5.6 / 67.3 & 16.5 & 64.4 / 64.8 & 14.4 / 59.0 & 4.9 / 63.0 & 15.2 & \textbf{61.4} / \textbf{59.3} \\
4 & 5.9 / 106.7 & 2.7 / 83.1 & 6.5 & 92.4 / 97.5 & 7.5 / 90.4 & 1.1 / 68.8 & 7.5 & 80.9 / 78.8 & 9.3 / 80.4 & 0.7 / 63.4 & 9.4 & \textbf{74.8} / \textbf{69.4} \\
5 & -3.2 / 68.0 & 9.6 / 80.5 & 10.1 & 75.1 / 74.3 & -3.1 / 64.3 & 9.9 / 71.5 & 10.4 & 69.6 / 66.9 & -2.2 / 62.5 & 10.5 / 68.1 & 10.8 & \textbf{67.3} / \textbf{63.9} \\
6 & -1.0 / 66.1 & -8.3 / 70.4 & 8.4 & 70.4 / 66.7 & -0.3 / 59.6 & -7.1 / 62.1 & 7.1 & 63.6 / 58.4 & -0.2 / 54.7 & -7.3 / 56.7 & 7.3 & \textbf{59.8} / \textbf{51.8} \\ \hline \rowcolor{Gray}
Mean & -0.4 / 79.6 & 3.3 / 82.3 & 11.3 & 78.8 / 81.0 & -0.5 / 71.4 & 2.8 / 73.5 & 11.2 & 71.8 / 71.3 & -0.3 / 66.8 & 2.6 / 69.7 & 11.6 & \textbf{68.5} / \textbf{66.2} \\ \rowcolor{Gray}
Std & 10.0 / 13.7 & 6.4 / 13.5 & 4.8 & \textbf{12.4} / 11.0 & 9.9 / 10.8 & 6.2 / 14.9 & 4.3 & 12.5 / \textbf{10.1} & 10.0 / 9.7 & 6.6 / 16.0 & 4.0 & 13.2 / 10.8 \\ \hline
\end{tabular}
}
\vspace{-10pt} 
\caption{Mean and standard deviation (Mean/Std) of offset errors ($\Delta X, \Delta Y$) in X and Y dimensions and L2 distance $D=\sqrt{(\Delta x)^2 + (\Delta y)^2}$. $D_{mean}$ is the mean distance of ground truth CoP to  the CoP of  foot pressure maps predicted by KNN where K=2, K=5, and K=10, respectively.}
\label{tab:CoPL2ErrorKNNs} \vspace{-5pt} 
\end{table*}

\begin{table*}[!t] \centering 
\resizebox{1\linewidth}{!}{
\begin{tabular}{|r||ccc|c||ccc|c||ccc|c|} \hline
    \multicolumn{13}{|c|}{CoP Robust Analysis X, Y, \& Spread Error of KNN, PressNet (PN), and PressNet-Simple (PN-S) in mm} \\  \hline
Sub & \multicolumn{4}{c||}{KNN (Median / rStd)} & \multicolumn{4}{c||}{PN (Median / rStd)} & \multicolumn{4}{c|}{PN-S (Median / rStd)} \\ \hline
\# & $\Delta X$ & $\Delta Y$ & D & D$_{median}$ & $\Delta X$ & $\Delta Y$ & D & D$_{median}$ &  $\Delta X$ & $\Delta Y$ & D & D$_{median}$ \\  \hline
1 & -5.2 / 45.6 & 0.7 / 40.7 & 5.3 & 49.3 / 40.2 & 3.3 / 33.2 & 1.6 / 32.9 & 3.6 & 37.3 / 28.2 & -5.0 / 32.4 & -0.2 / 31.3 & 5.1 & \textbf{35.1} / 30.0\\ 
2 & -12.4 / 56.5 & 3.7 / 63.0 & 12.9 & 69.7 / 57.4 & -13.6 / 51.4 & 8.2 / 66.0 & 15.9 & \textbf{66.6} / \textbf{52.9} & -17.4 / 54.1 & 3.6 / 62.4 & 17.7 & 67.8 / 55.2\\ 
3 & 11.5 / 44.8 & 1.6 / 43.9 & 11.7 & 49.2 / 40.3 & 12.1 / 35.8 & -1.6 / 35.6 & 12.2 & 41.6 / 32.2 & 8.9 / 35.2 & -2.2 / 33.3 & 9.1 & \textbf{39.9} / 30.1\\ 
4 & 5.3 / 54.7 & 1.7 / 49.5 & 5.6 & 61.6 / 50.6 & 14.7 / 46.9 & 2.6 / 38.4 & 14.9 & 48.2 / 38.6 & 10.7 / 44.3 & 2.7 / 35.6 & 11.0 & \textbf{46.7} / 35.5\\ 
5 & -3.2 / 37.8 & 8.9 / 43.4 & 9.4 & 48.6 / 42.1 & 1.5 / 37.1 & 8.6 / 43.1 & 8.7 & 46.5 / 38.8 & -1.6 / 29.7 & 12.2 / 34.3 & 12.3 & \textbf{38.0} / 30.0\\ 
6 & 1.8 / 39.8 & -7.0 / 39.5 & 7.3 & 48.5 / 42.2 & 3.6 / 36.9 & -6.8 / 35.6 & 7.7 & 42.7 / 33.0 & 4.9 / 31.6 & -9.2 / 32.5 & 10.4 & \textbf{38.6} / 29.8\\ 
\hline \rowcolor{Gray}
Mean & -0.3 / 46.5 & 1.6 / 46.7 & 8.7 & 54.5 / 45.5 & 3.6 / 40.2 & 2.1 / 41.9 & 10.5 & 47.1 / 37.3 & 0.1 / 37.9 & 1.2 / 38.3 & 10.9 & \textbf{44.3} / 35.1\\ 
\rowcolor{Gray}
Std & 7.7 / 6.9 & 4.7 / 8.0 & 2.9 & \textbf{8.2 / 6.4} & 9.1 / 6.6 & 5.4 / 11.2 & 4.3 & 9.4 / 7.9 & 9.5 / 8.6 & 6.5 / 10.9 & 3.8 & 11.1 / 9.2\\ 
\hline 
\end{tabular}
}
\vspace{-10pt} 
\caption{Geometric median and robust standard deviation (rStd = 1.4826 $\times$ median absolute deviation) of offset errors ($\Delta X, \Delta Y$) in X and Y dimensions and L2 distance $D=\sqrt{(\Delta x)^2 + (\Delta y)^2}$.
$D_{median}$ is the median and rStd (med/rStd) of  distances between ground truth CoP and the CoP of  foot pressure maps predicted by KNN, PressNet, and PressNet-Simple, respectively. See Figure~\ref{fig:coperrorbar} for a graphical representation.}
\label{tab:CoPMedianError} \vspace{-5pt} 
\end{table*}

\begin{table*}[!t] \centering 
\resizebox{1\linewidth}{!}{
\begin{tabular}{|r||ccc|c||ccc|c||ccc|c|} \hline
    \multicolumn{13}{|c|}{CoP Robust Analysis X, Y, \& Spread Error of KNN K=2, K=5, and K=10 in mm} \\  \hline
Sub & \multicolumn{4}{c||}{K = 2 (Median / rStd)} & \multicolumn{4}{c||}{K = 5 (Median / rStd)} & \multicolumn{4}{c|}{K = 10 (Median / rStd)} \\ \hline 
\# & $\Delta X$ & $\Delta Y$ & D & D$_{median}$ & $\Delta X$ & $\Delta Y$ & D & D$_{median}$  &  $\Delta X$ & $\Delta Y$ & D & D$_{median}$ \\  \hline
1 & -5.2 / 45.6 & 0.7 / 40.7 & 5.3 & 49.3 / 40.2 & -4.43 / 40.48 & -0.51 / 35.83 & 4.46 & 43.42 / 35.32  &  -3.75 / 35.45 & -0.51 / 34.24  &  3.78  & \textbf{40.58 / 33.44} \\
2 & -12.4 / 56.5 & 3.7 / 63.0 & 12.9 & 69.7 / 57.4 & -13.07 / 54.78  & 3.20 / 61.58 & 13.46 & \textbf{68.80} / \textbf{55.14} & -14.43 / 56.39  & 2.79 / 62.3  & 14.70  & 69.05 / 55.28 \\
3 & 11.5 / 44.8 & 1.6 / 43.9 & 11.7 & 49.2 / 40.3 & 11.39 / 41.75 &  1.46 / 39.65 &   11.48 & 45.26 / 35.74 &  11.52 / 41.09 & 1.41 / 38.19 & 11.60 &   \textbf{44.35 / 35.88} \\
4 & 5.3 / 54.7 & 1.7 / 49.5 & 5.6 & 61.6 / 50.6 & 7.10 / 53.13 & 2.10 / 45.65 & 7.40 &  57.53 / 45.55 & 8.21 / 49.86  & 1.51 / 42.69  & 8.34  &  \textbf{52.72 / 42.11} \\
5 & -3.2 / 37.8 & 8.9 / 43.4 & 9.4 & 48.6 / 42.1 & -2.85 / 35.86  &  8.10 / 40.99  &  8.59  & 45.18 / 39.52  & -2.55 / 35.03  &  8.44 / 40.22 & 8.82 & \textbf{44.03 / 37.48} \\
6 & 1.8 / 39.8 & -7.0 / 39.5 & 7.3 & 48.5 / 42.2 & 2.07 / 37.04  & -6.35 / 37.12  &  6.68  & 44.30 / 37.90 & 2.70 / 35.68 & -6.34 / 35.46 & 6.89 & \textbf{42.75 / 36.06} \\ \hline \rowcolor{Gray}
Mean & -0.3 / 46.5 & 1.6 / 46.7 & 8.7 & 54.5 / 45.5 & 0.03 / 43.84  &  1.33 / 43.47 & 8.68 & 50.75 / 41.53  & 0.28 / 42.25 & 1.22 /  42.18 & 9.02 & \textbf{48.91 / 40.04} \\
\rowcolor{Gray}
Std & 7.7 / 6.9 & 4.7 / 8.0 & 2.9 & \textbf{8.2 / 6.4} & 7.99 / 7.43  &  4.32 / 8.68  &  3.00  &  9.38 / 6.96  & 8.52 / 8.17 & 4.37 / 9.43 & 3.45 & 9.76 / 7.30\\ 
\hline 
\end{tabular}
}
\vspace{-10pt} 
\caption{Geometric median and robust standard deviation (rStd = 1.4826 $\times$ median absolute deviation) of offset errors ($\Delta X, \Delta Y$) in X and Y dimensions and L2 distance $D=\sqrt{(\Delta x)^2 + (\Delta y)^2}$.
$D_{median}$ is the median and rStd (med/rStd) of  distances between ground truth CoP and the CoP of  foot pressure maps predicted by KNN K=2, K = 5 and K = 10 respectively.}
\label{tab:CoPMedianErrorKNN5and50} \vspace{-5pt} 
\end{table*}

\begin{figure*}[!t]
\begin{center}
\resizebox*{.81\textwidth}{!}{
    \begin{tabular}{cc}
        \includegraphics[width=0.95\linewidth]{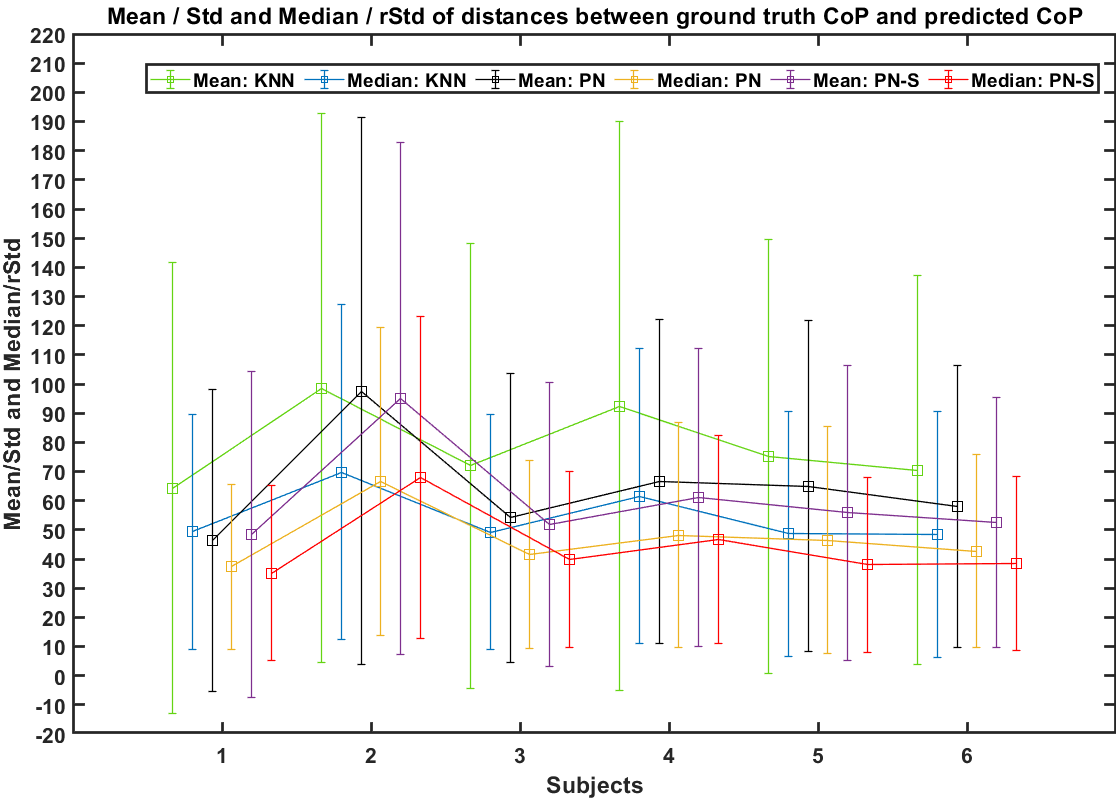} &
        \includegraphics[width=1\linewidth]{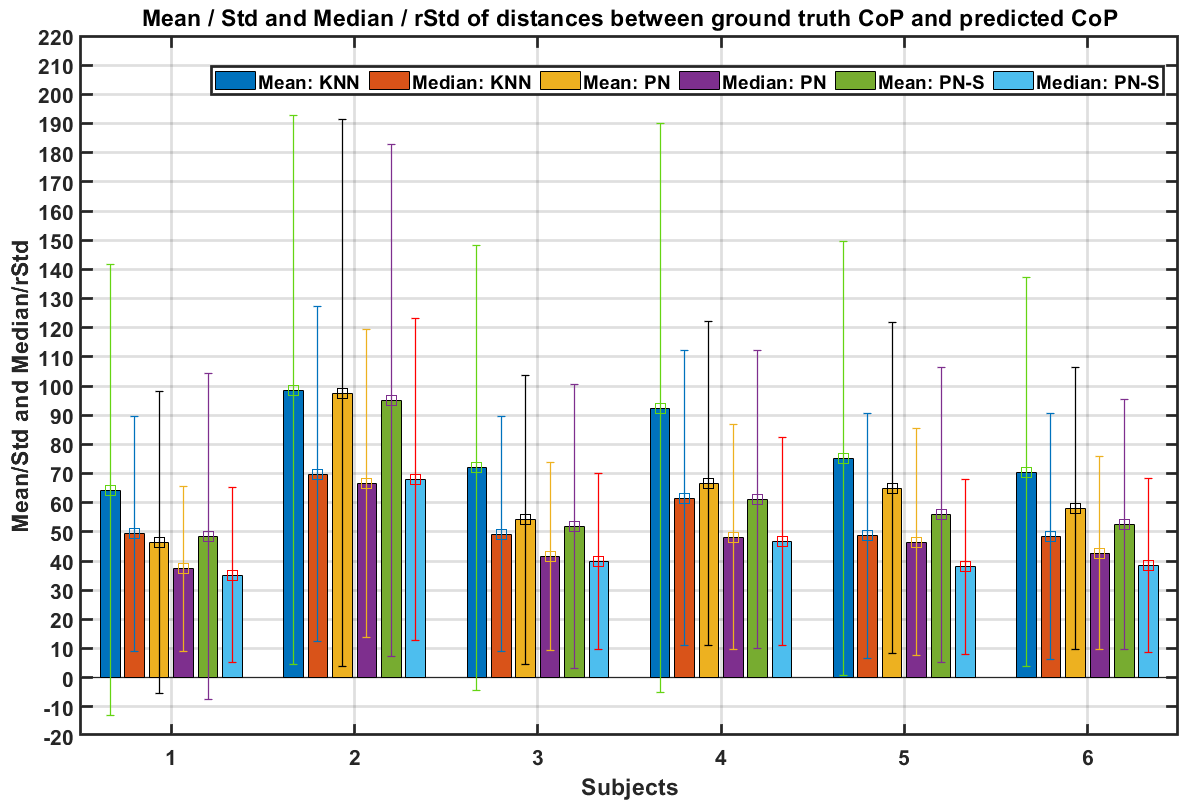}
    \end{tabular}
    }
    \end{center}\vspace{-10pt}
   \caption[CoP Mean/Std and Median/rStd]{Plots representing Subjectwise Mean/Std and Median/rStd from Tables \ref{tab:CoPL2Error} and \ref{tab:CoPMedianError}}
\label{fig:coperrorbar} \label{fig:coperrorcurve} \vspace{-5pt}
\end{figure*}

\begin{figure*}[!t]
\begin{center}
\resizebox*{.79\textwidth}{!}{
    \begin{tabular}{cc}
    \includegraphics[width=1\linewidth]{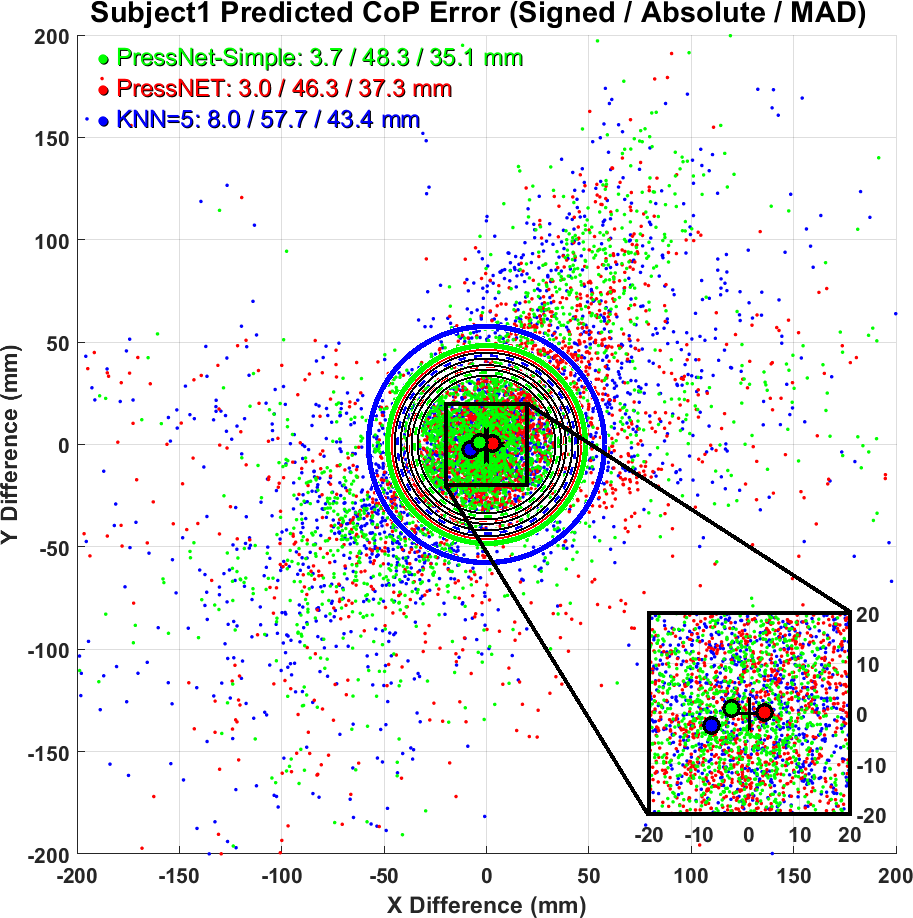} &
    \includegraphics[width=1\linewidth]{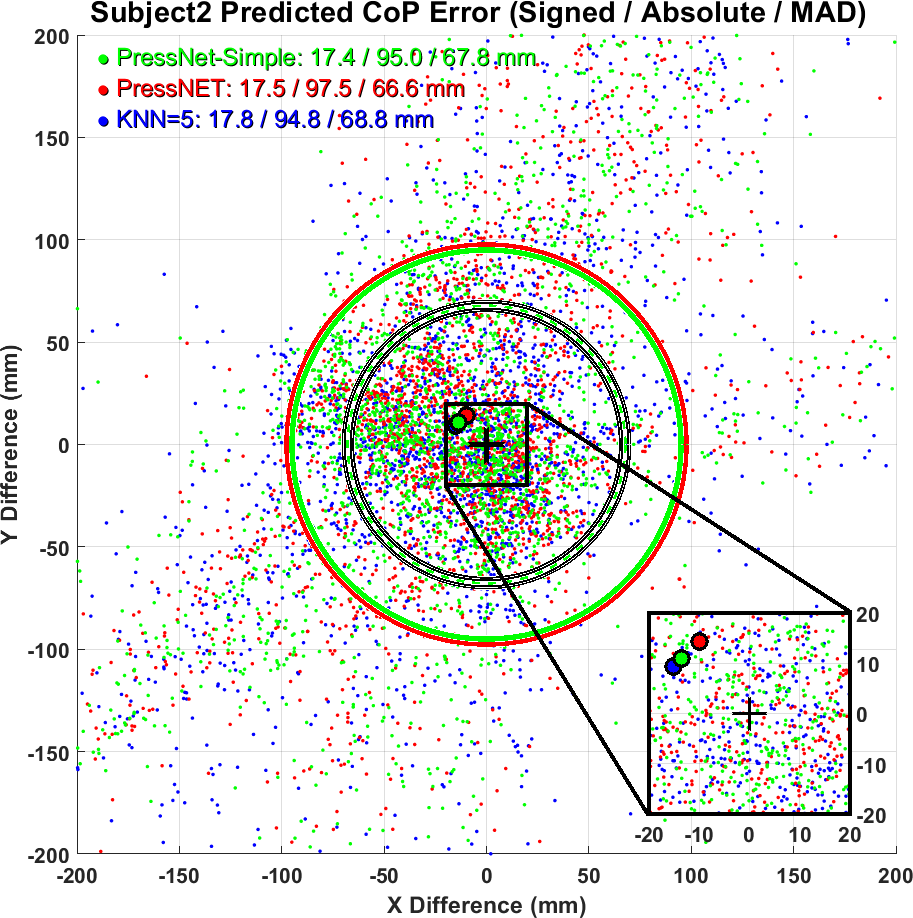} \\ & \\
    \includegraphics[width=1\linewidth]{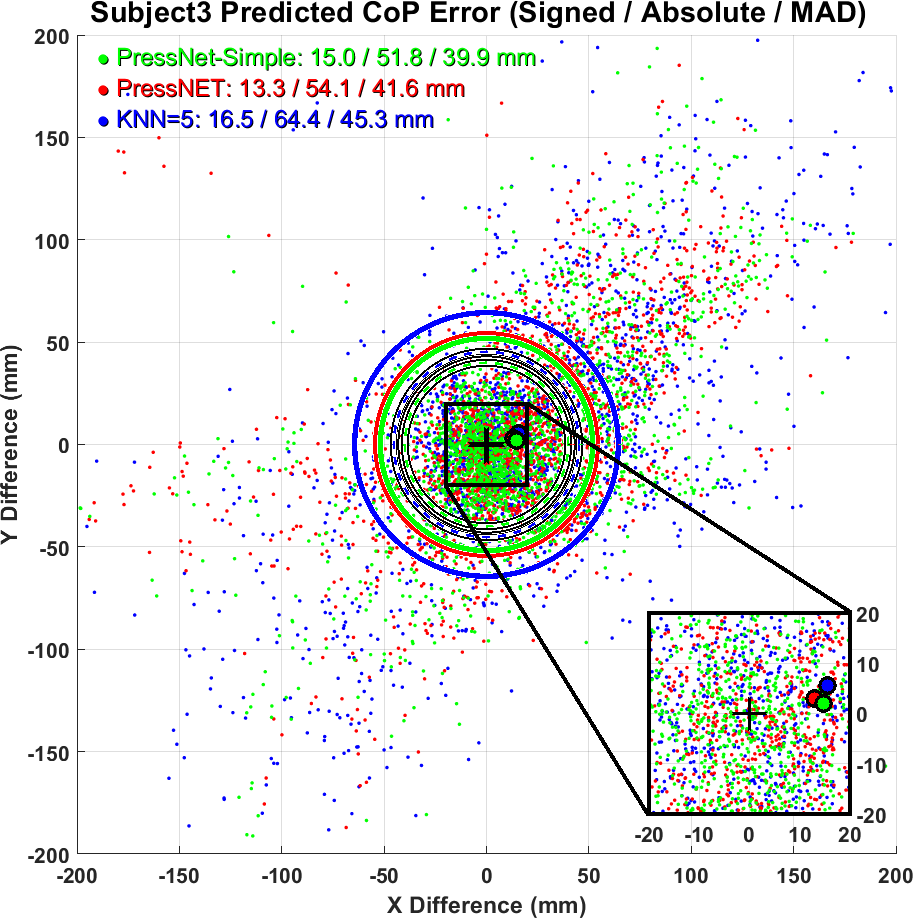} &
    \includegraphics[width=1\linewidth]{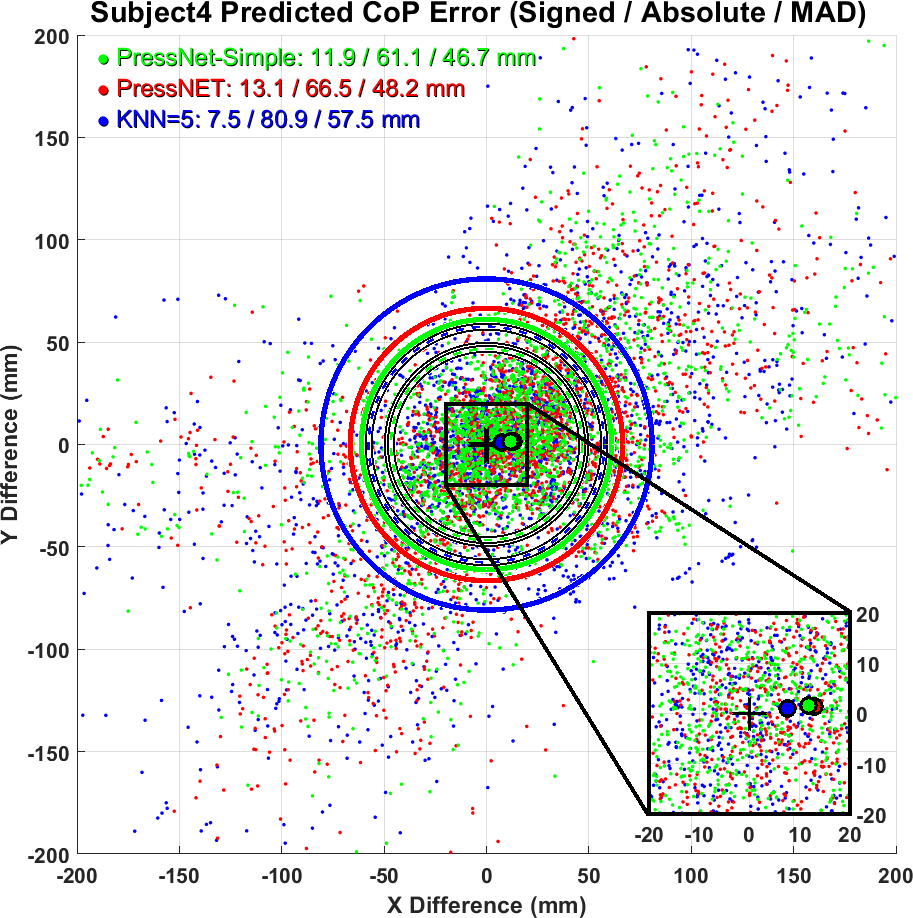} \\ & \\
    \includegraphics[width=1\linewidth]{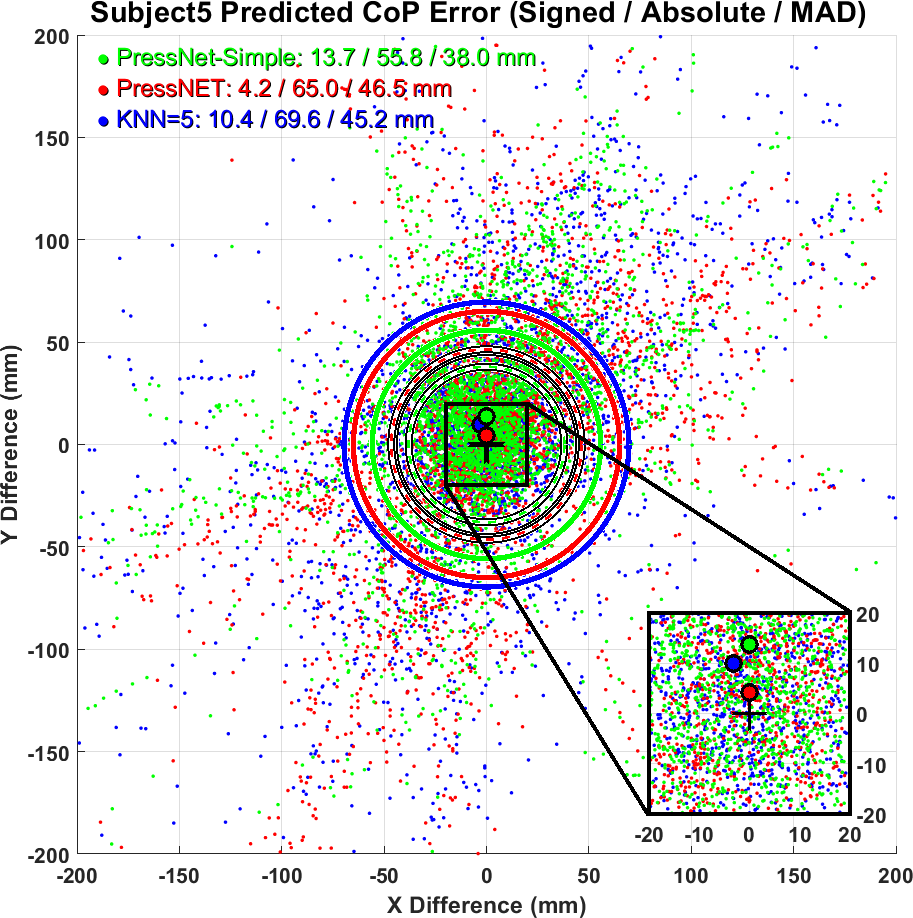} &
    \includegraphics[width=1\linewidth]{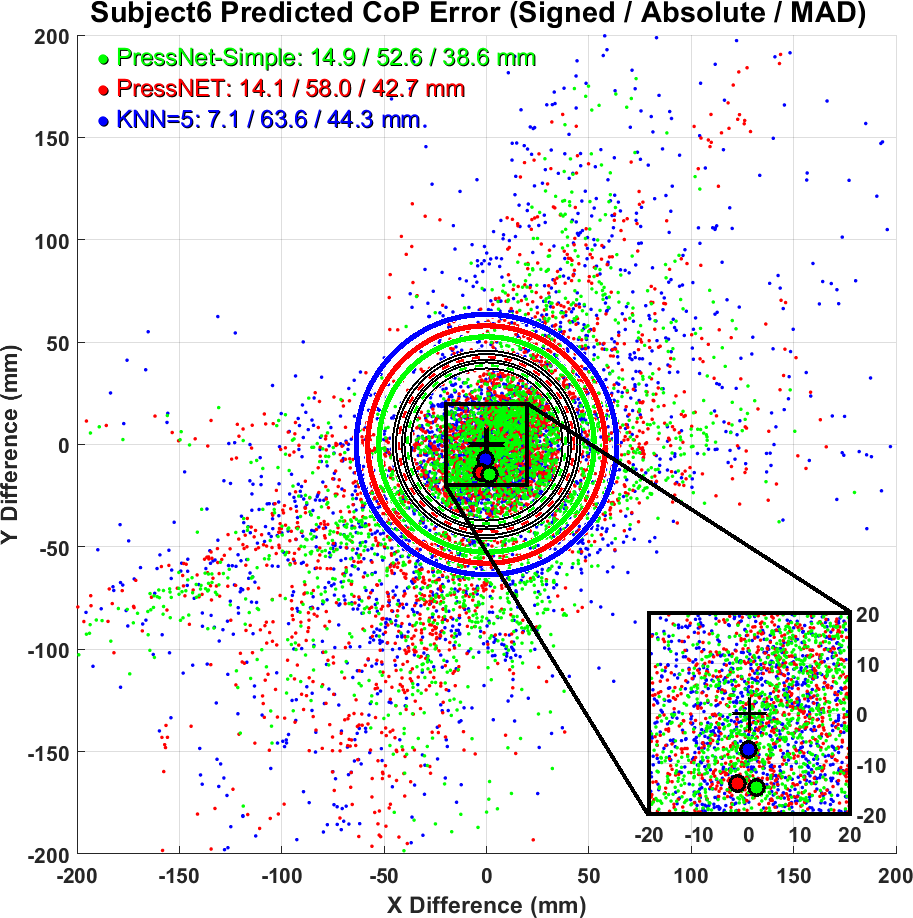} \\
    \end{tabular}
    }
    \end{center}\vspace{-10pt}
   \caption[CoP Euclidean error scatter Plots 1-6]{2D offsets between ground truth CoP (black) and CoP calculated from pressure maps predicted by KNN=5 (blue), PressNet (red) and PressNet-Simple (green), respectively. Large dots are the signed mean of each scatter, indicating relatively small directional bias (< 13 mm Table~\ref{tab:CoPL2Error}). The concentric circles, with mean (solid) and median (dashed) CoP distance error as its radius, are shown for each method, suggesting PN and PN-S outperform KNN.}
\label{fig:6a} \label{fig:6b} \vspace{-5pt}
\end{figure*}

Table~\ref{tab:errortableScaled} reflects results uses a completely different normalization that is applied to the training data prior to network training. The training data is normalized by subject mass and the constant prexel area.  The predicted results are rescaled using the inverse of this process prior to analysis. The comparison of Table~\ref{tab:errortable} and Table~\ref{tab:errortableScaled} show that weight scaling prior to training of the network makes small improvements of approximately 0.15 kPa in PressNet-Simple and similar for PressNet. The standard deviation of the subject means do appear to be better using pressure directly rather than mass scaling for training data.

To test whether MAE results are significant, a paired t-test was performed between the mean absolute errors of PressNet compared with KNN, PressNet-Simple compared with KNN, and PressNet-Simple compared with PressNet for each LOO data split. This test showed that the results from PressNet and PressNet-Simple are statistically significant improvements over the KNN and that PressNet-Simple is statistically significant relative to PressNet.

\begin{figure*}[!t] \centering
    \resizebox*{1.0\textwidth}{!}{
    \begin{tabular}{cc}
    \includegraphics[width=\linewidth]{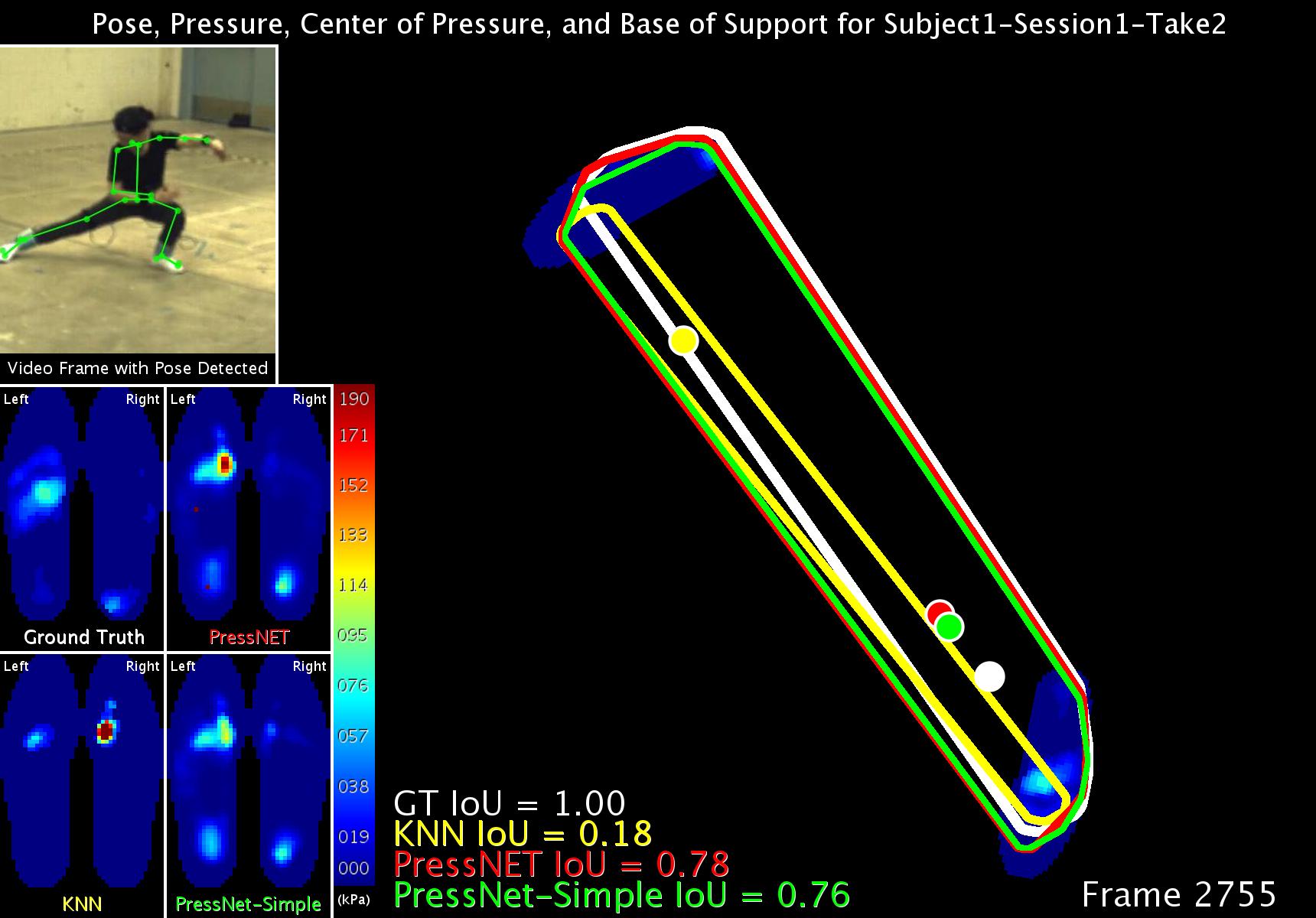}  & \includegraphics[width=\linewidth]{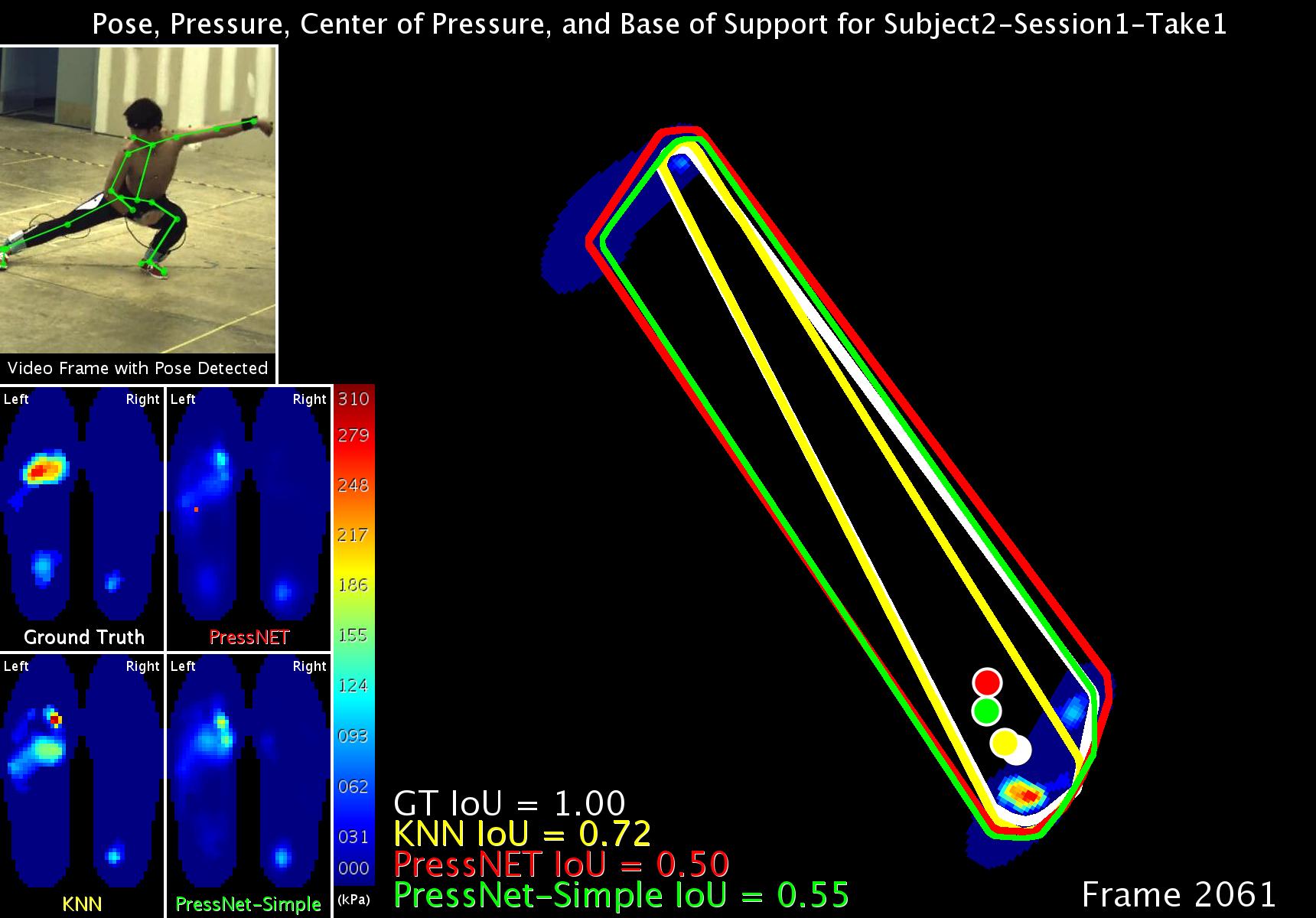}  \\ & \\
    \includegraphics[width=\linewidth]{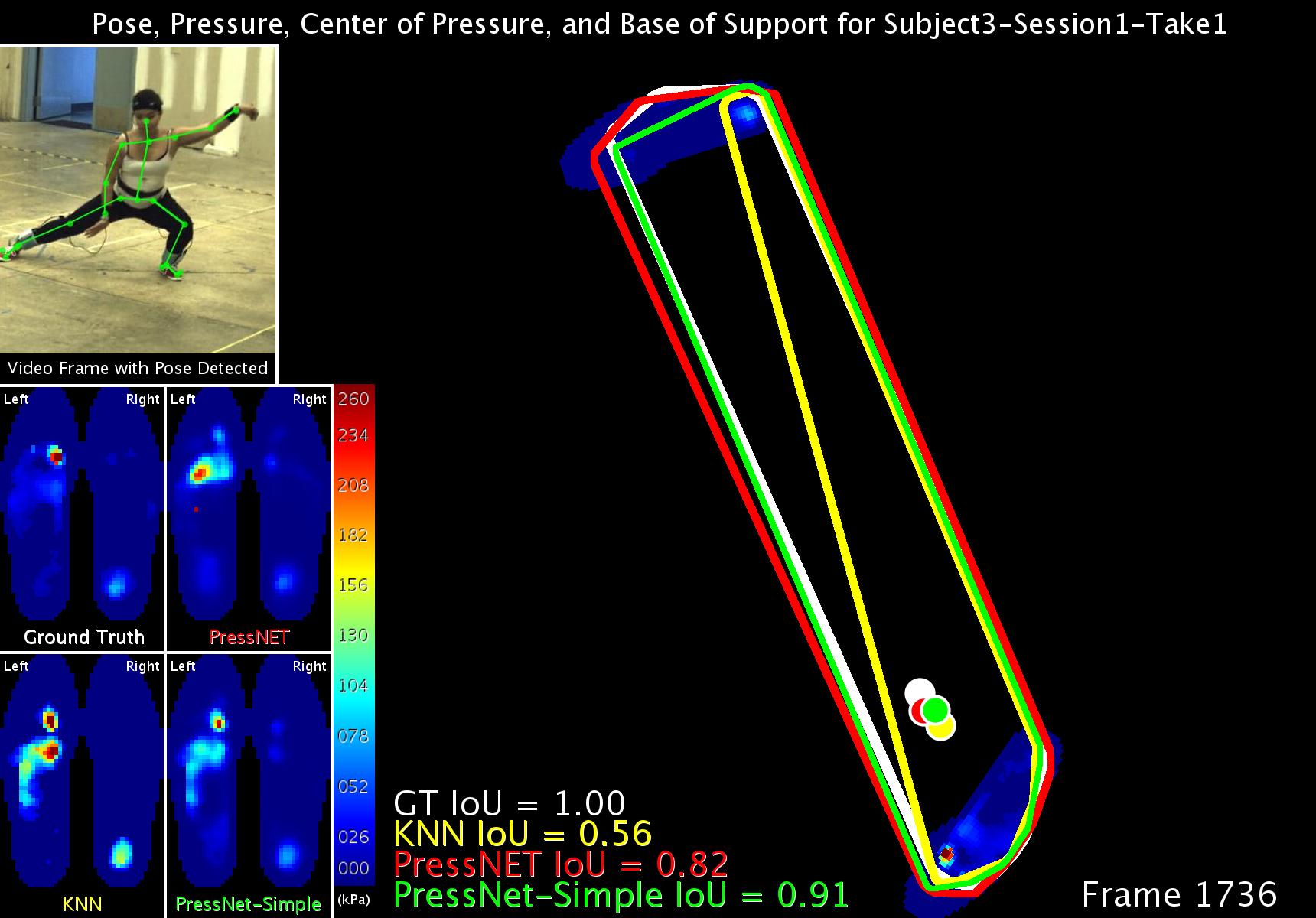}  & \includegraphics[width=\linewidth]{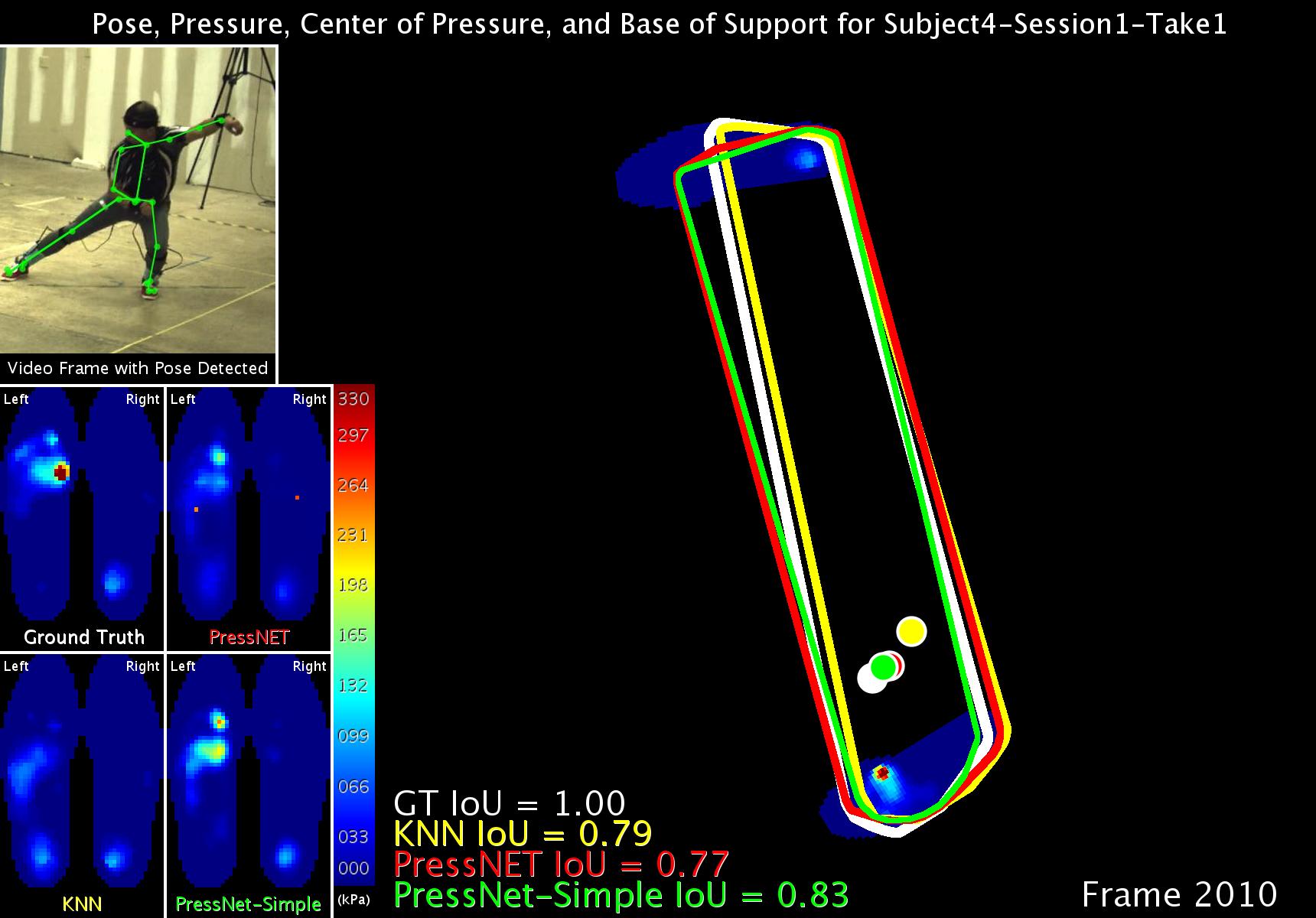} \\ & \\
    \includegraphics[width=\linewidth]{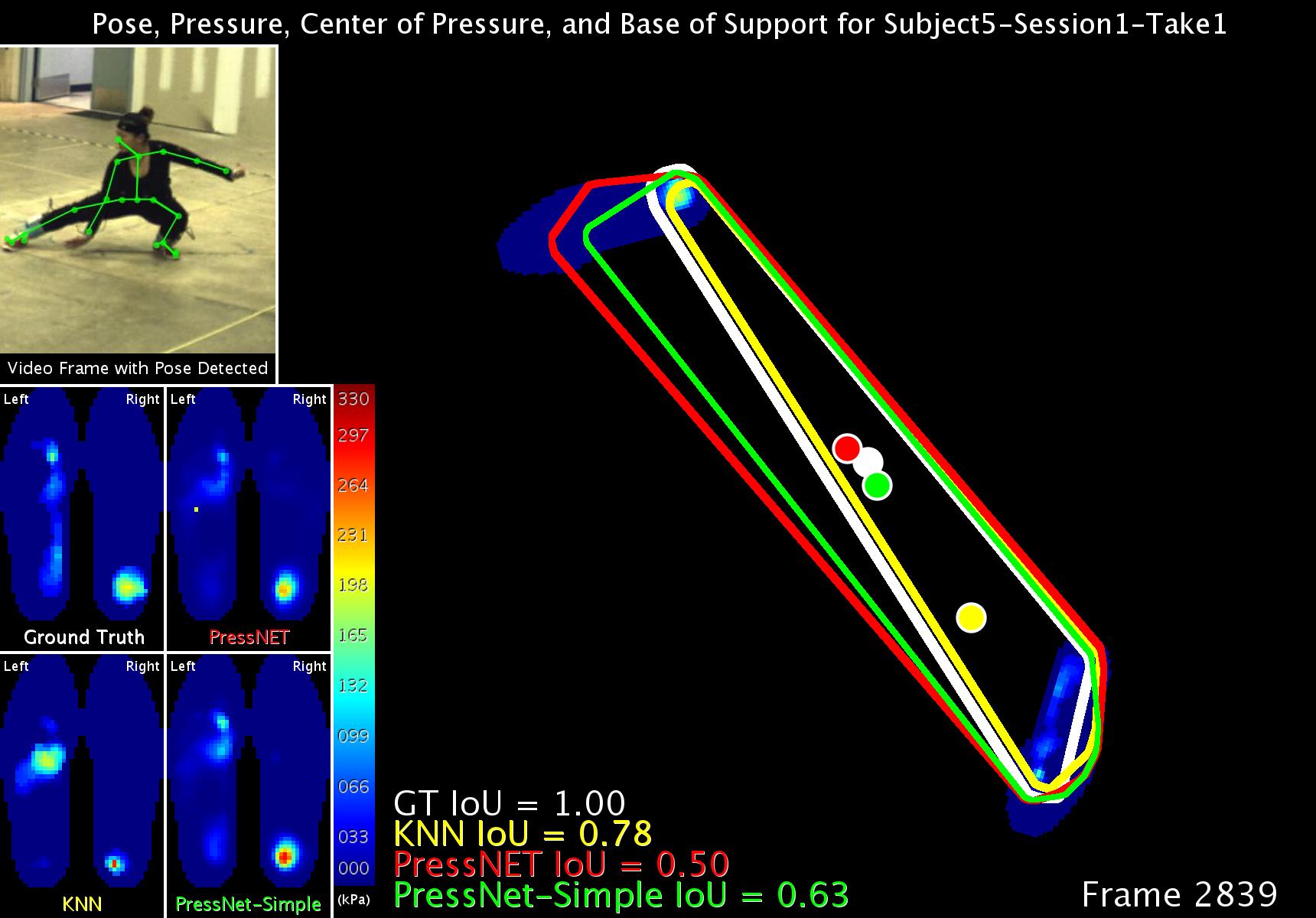} & \includegraphics[width=\linewidth]{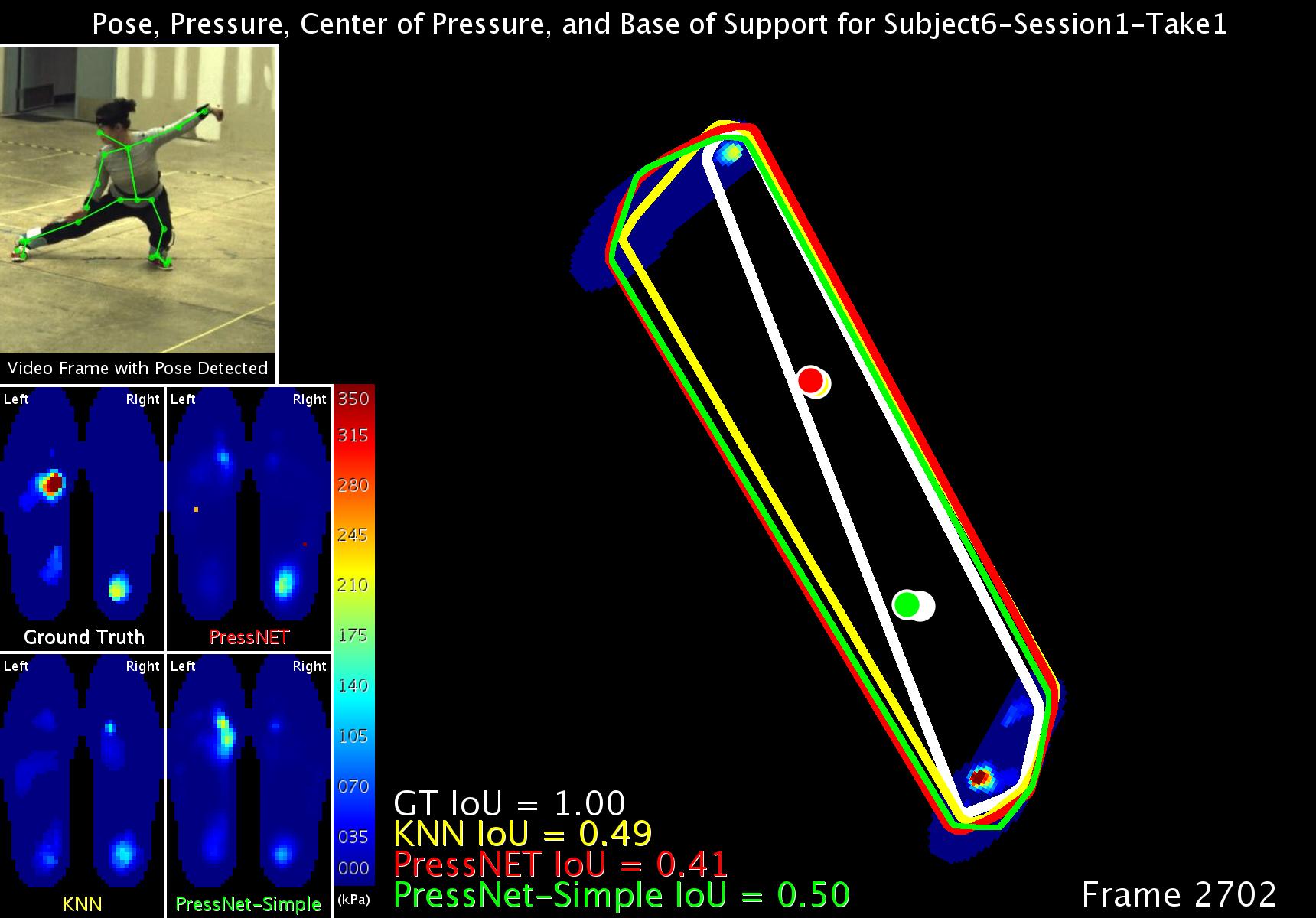} \\
    \end{tabular}
    }
    \vspace{-10pt} 
    \caption{Base of Support (BoS) and CoP estimations from sample frames of the subjects with similar poses. Ground truth, KNN, PressNet, PressNet-Simple foot pressure and input video frame with OpenPose BODY25 skeleton superimposed. Foot pressure is scaled for each subject based on their range of pressure. BoS and CoP of Ground Truth (white), KNN (cyan), PressNet (red) and PressNet-Simple (green) plotted as an overlay with the ground truth foot pressure projected onto the floor plane. Intersection over Union (IoU) measures of ground truth BoS region with BoS regions predicted by each method are also included for each frame.}
    \vspace{-5pt} 
    \label{fig:screenshotsCommon}
\end{figure*}

\begin{figure*}[!t] \centering
    \resizebox*{1.0\textwidth}{!}{
    \begin{tabular}{cc}
    \includegraphics[width=\linewidth]{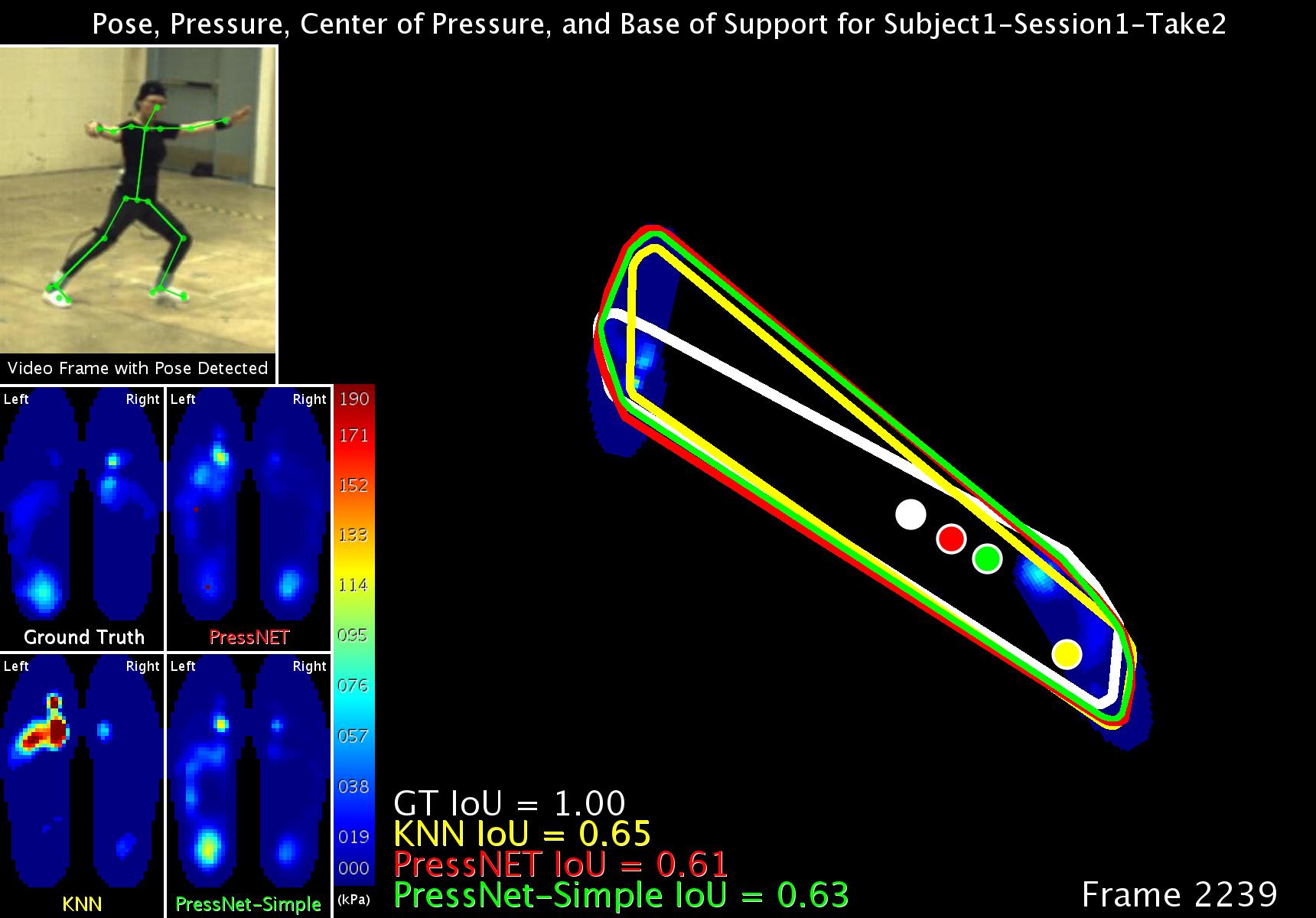}  & \includegraphics[width=\linewidth]{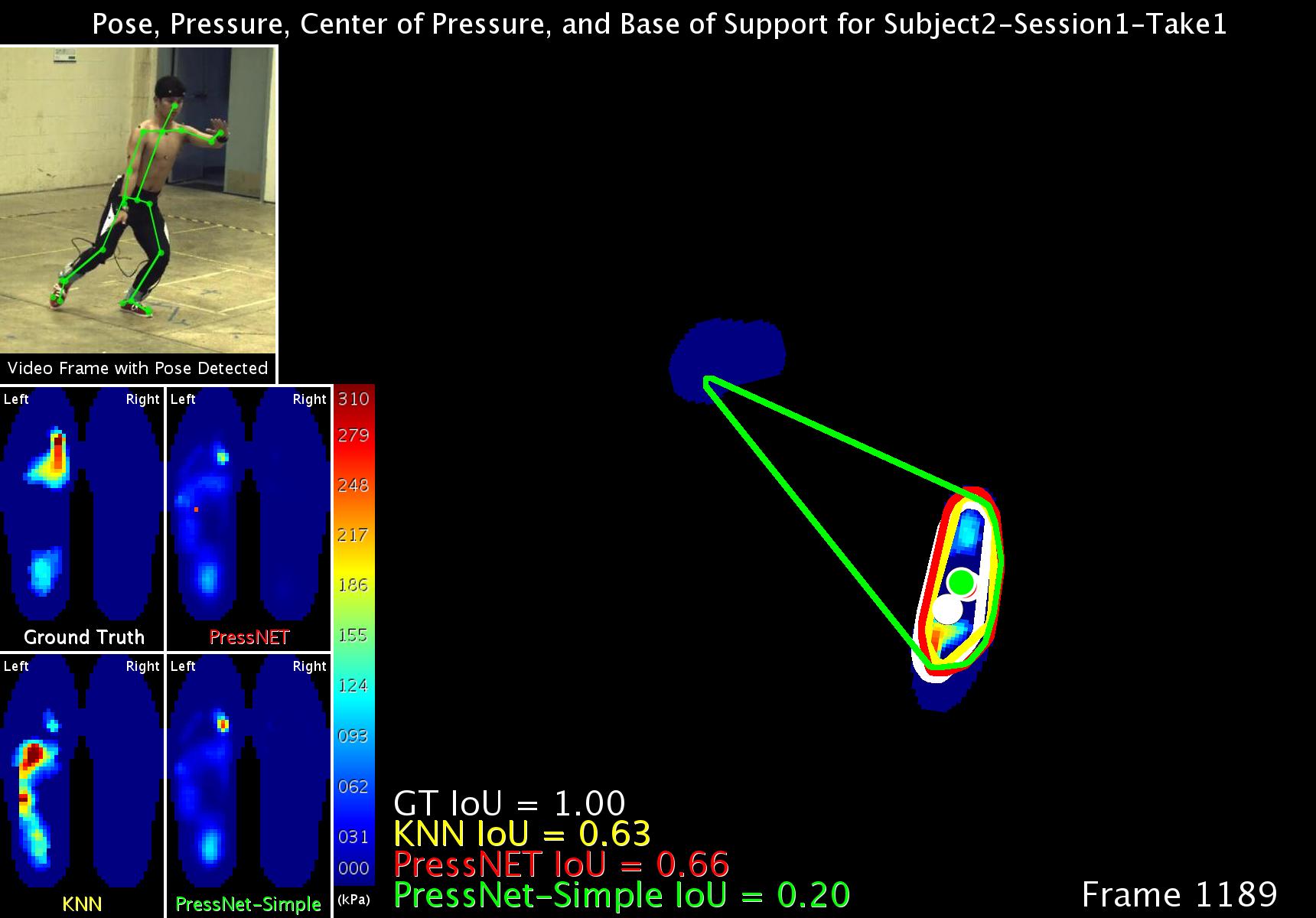}  \\ & \\
    \includegraphics[width=\linewidth]{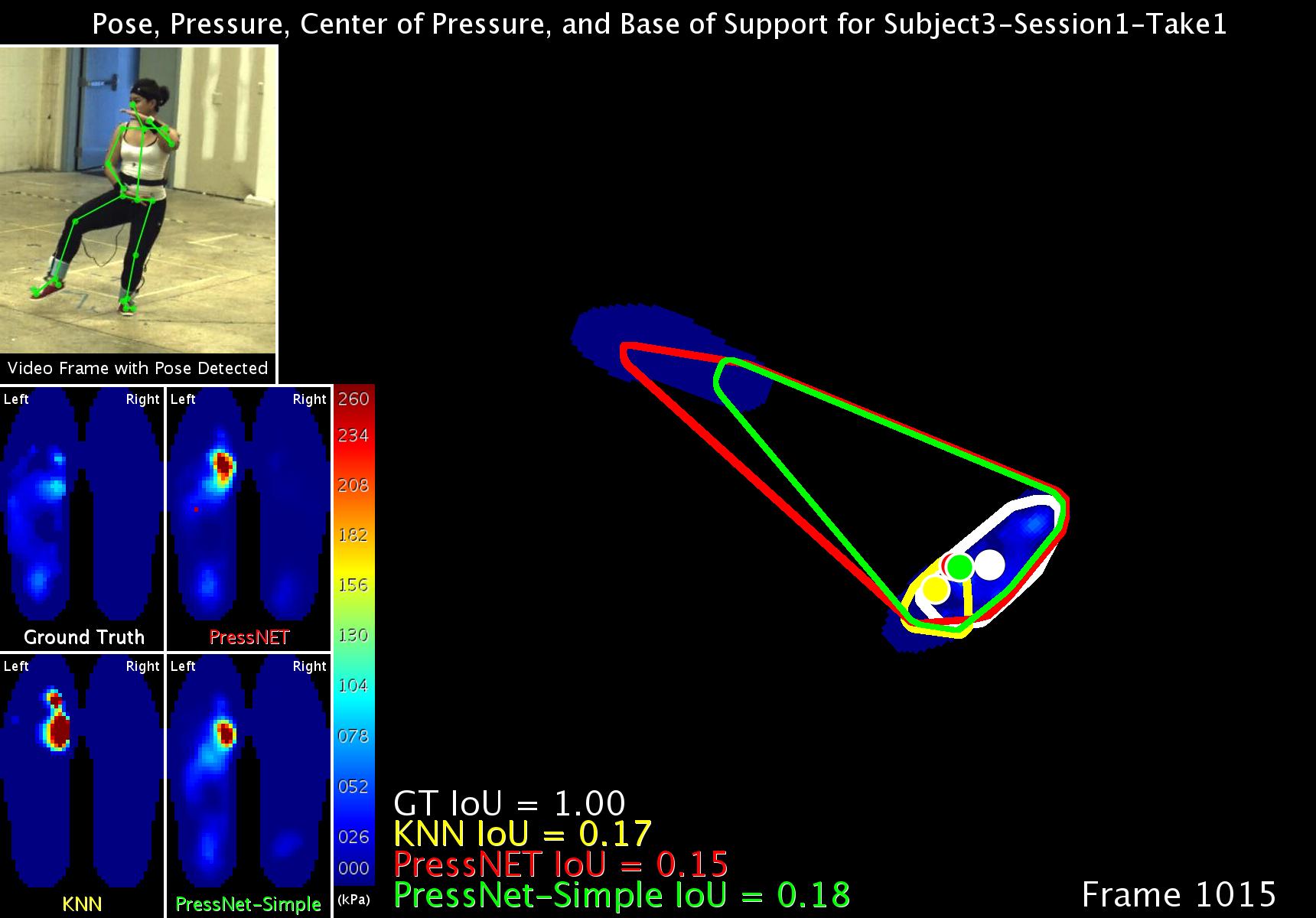}  & \includegraphics[width=\linewidth]{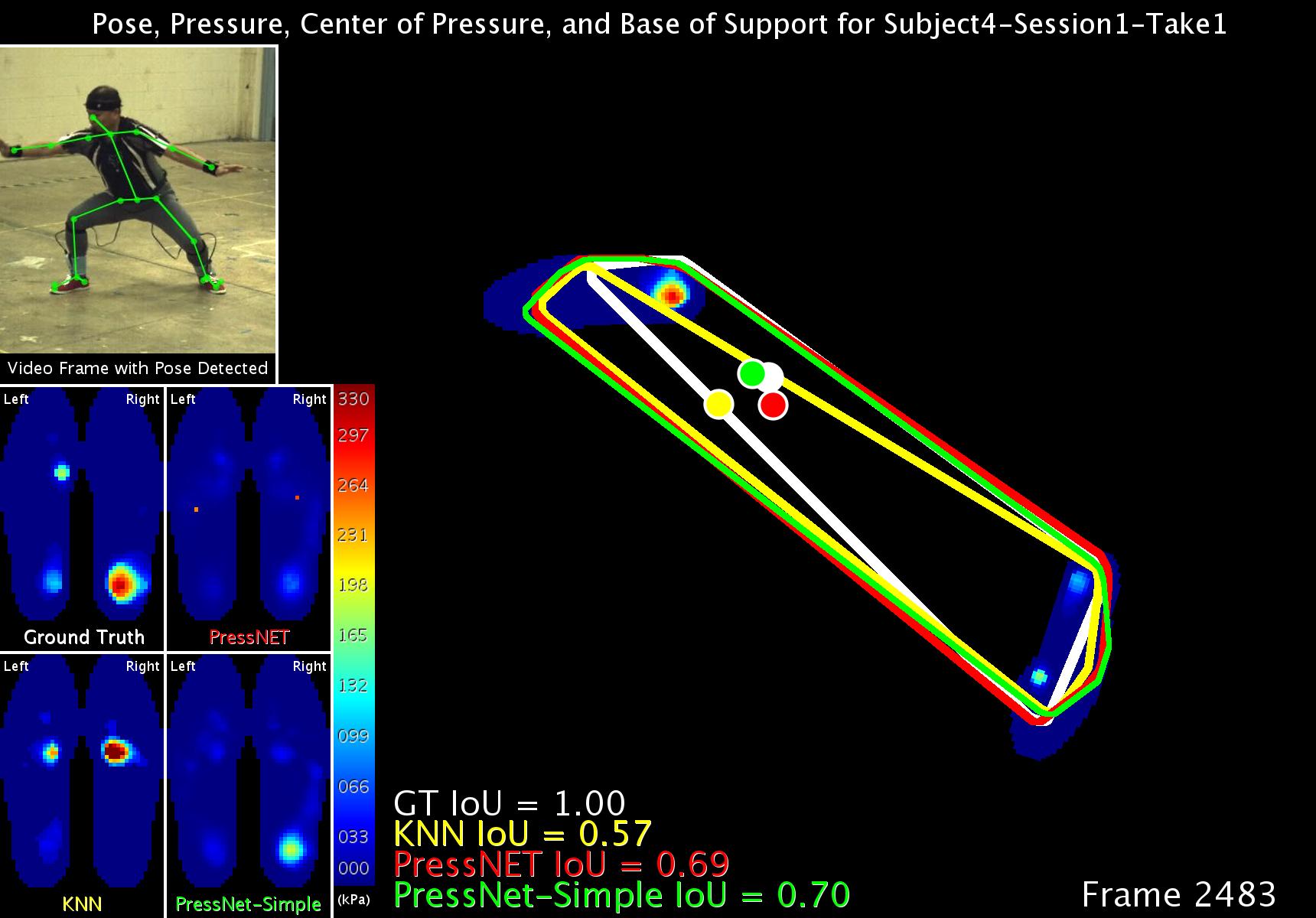} \\ & \\
    \includegraphics[width=\linewidth]{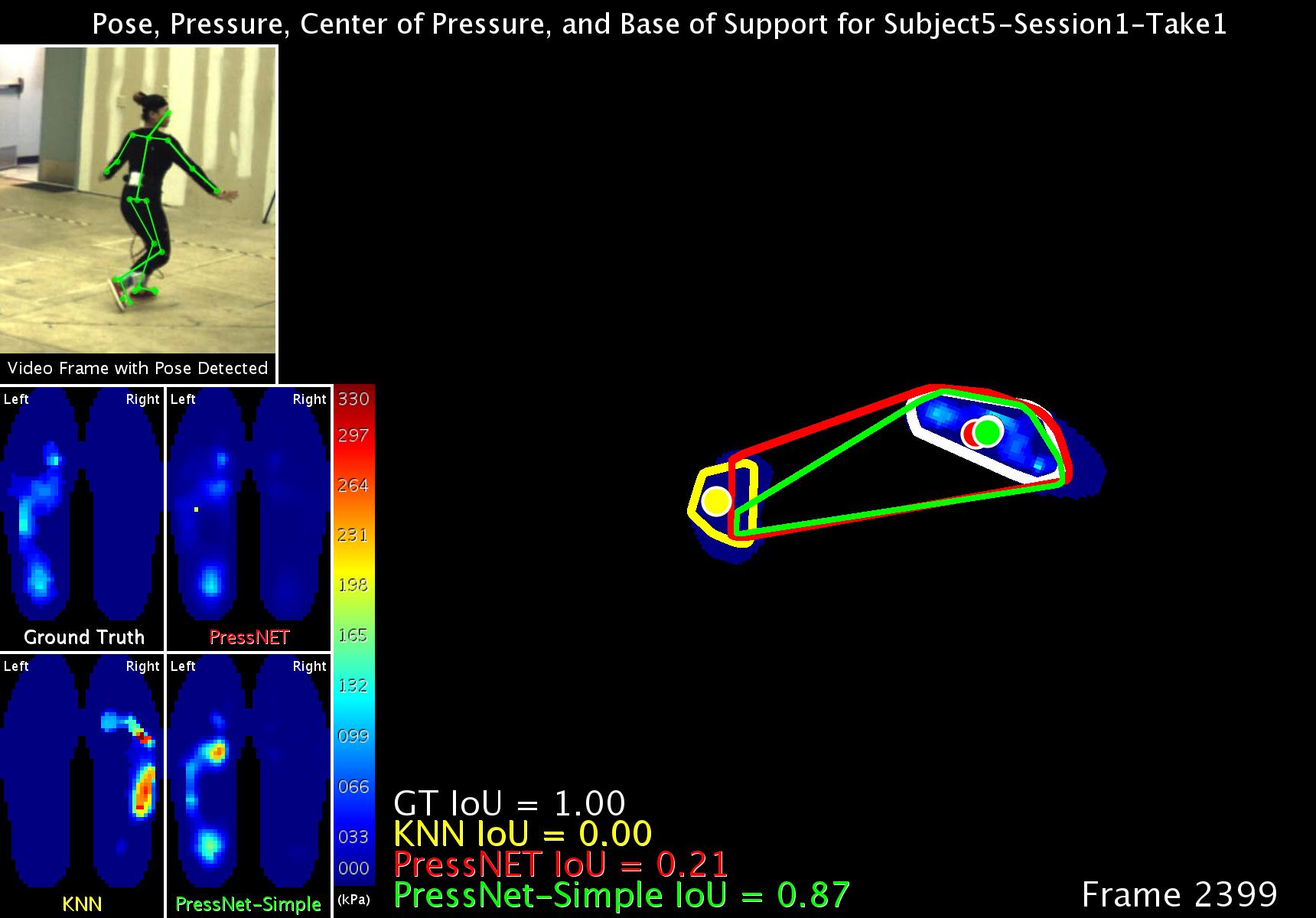} & \includegraphics[width=\linewidth]{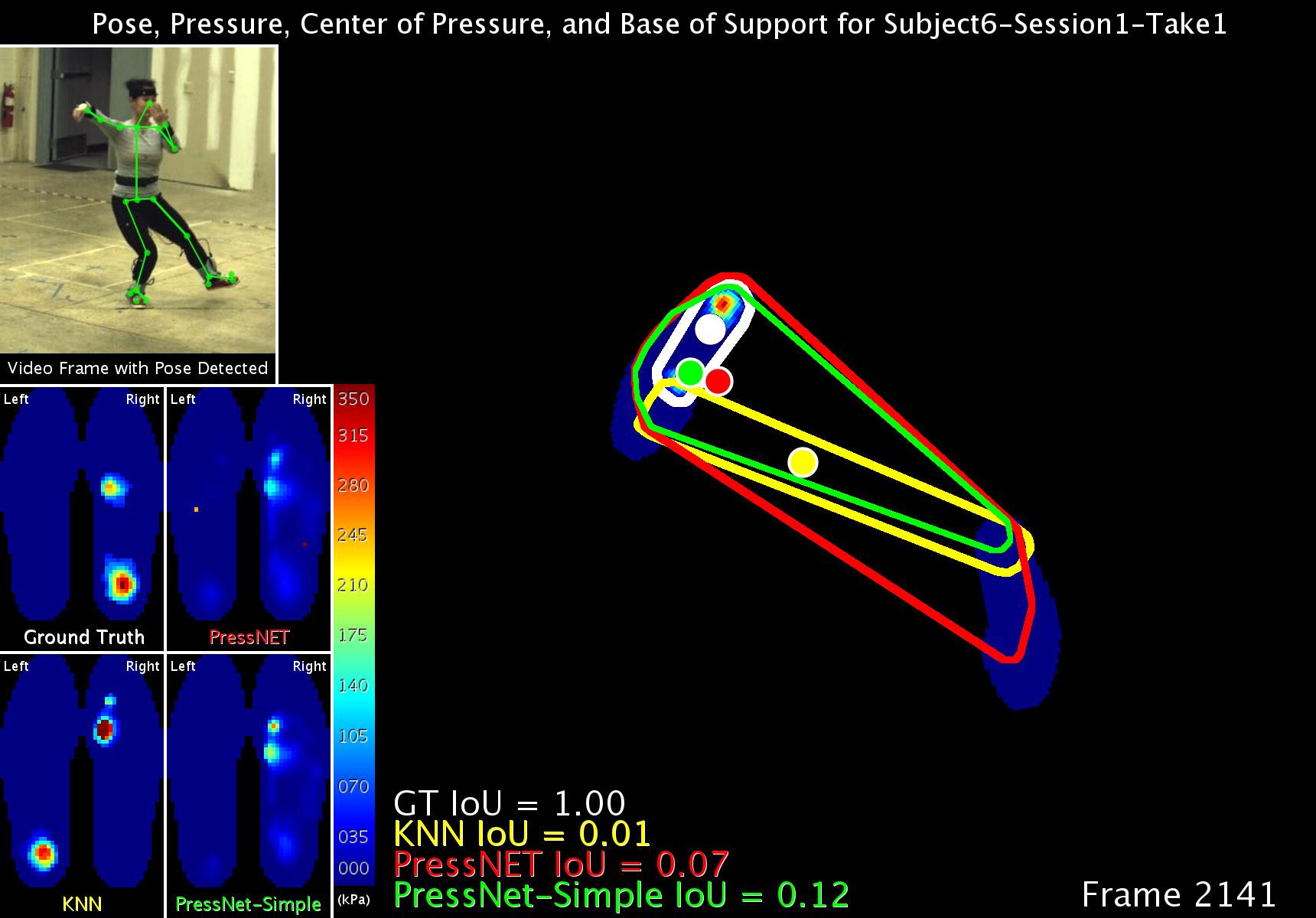} \\
    \end{tabular}
    }
    \vspace{-10pt} 
    \caption{Sample output frames from each of the 6 subjects emphasizing superior MAE performance of PressNet-Simple over KNN and PressNet.  \textbf{Left Side}: Ground truth, KNN, PressNet, PressNet-Simple foot pressure and input video frame with OpenPose BODY25 skeleton superimposed. Foot pressure is scaled for each subject based on their range of pressure. \textbf{Right Side}: BoS and CoP of Ground Truth (white), KNN (yellow), PressNet (red) and PressNet-Simple (green) plotted as an overlay with the ground truth foot pressure projected onto the floor plane. Intersection over Union (IoU) measures of ground truth BoS region with BoS regions predicted by each method are also included for each frame.} \vspace{-5pt} 
    \label{fig:screenshotsBest}
\end{figure*}

\vspace{-1em}
\subsubsection{Spatial Distribution Metrics}

Table~\ref{tab:DistributionMetrics} provides three different spatial distribution comparison metrics intended to compare the spatial distribution of a predicted foot pressure and ground truth independent of the magnitude of the pressure.  Similarity, KL Divergence, and Information Gain have been used to provide these magnitude normalized comparisons.  While the Similarity metric does not discriminate between the two deep learning methods, it does show that both are better than KNN at matching the ground truth distributions.  KL Divergence and Information gain both show that PressNet-Simple is a large improvement over both KNN and PressNet.

\subsubsection{$\ell_2$ Distance for CoP Error}
As a step towards analyzing gait stability from video, Center of Pressure (CoP) from regressed foot pressure maps of KNN, PressNet, and PressNet-Simple have been computed and quantitatively compared to ground truth (GT) CoP locations computed from the insole foot pressure maps. CoP is calculated as the weighted mean of the pressure elements (prexels) in the XY ground plane. A minimum threshold of 3 kPa is applied to both the ground truth and the predicted pressure, similar to the procedure used in~\cite{KEIJSERS200987} based on the limitations of the insole pressure measurement hardware. The $\ell_2$ distance is used to quantify the 2D error between ground truth and predicted CoP locations. Table~\ref{tab:CoPL2Error} shows the $\ell_2$ errors calculated for mean X and Y errors of KNN, PressNet, and PressNet-Simple. This table shows that the mean Euclidean distance of CoP error for the PressNet network pressure map predictions is smaller by 14.2 mm than that of KNN, with a standard deviation that is approximately 25\% smaller. The PressNet average error for all leave one subject out experiments yields a CoP Euclidean error of 64.6 mm with a standard deviation of 59.3 mm. The PressNet-Simple average error for all leave one subject out experiments yields a CoP Euclidean error of 60.8 mm with a standard deviation of 56.1 mm. To investigate the impact of K, Table~\ref{tab:CoPL2ErrorKNNs} presents CoP quantification for K=2, K=5, and K=10.  While there are small reductions in $D_{mean}$, the D values are equivalent which means that higher K values make small improvements in CoP mean error while maintaining accuracy.

Table~\ref{tab:CoPMedianError} presents a median-based analysis of CoP offset errors as a robust alternative to the mean/std statistics in Table~\ref{tab:CoPL2Error}. Central location X,Y is estimated by 2D geometric median, computed by Weiszfeld's algorithm \cite{VardiZhang2000}.  Spread of data is estimated by a robust standard deviation measure, derived as median absolute deviation (MAD) from the median, multiplied by a constant 1.4826 that scales MAD to be a {\it consistent estimator} of population standard deviation \cite{madestimator}.  Overall distance error is characterized by median and robust std of Euclidean distances between ground truth CoP and  predicted CoP locations. To investigate the impact of K, Table~\ref{tab:CoPMedianErrorKNN5and50} presents CoP quantification for K=2, K=5, and K=10.  While there are small reductions in $D_{median}$, the D values are equivalent which means that higher K values make small improvements in CoP median error while maintaining accuracy. These error values are smaller than those is Tables~\ref{tab:CoPL2Error} and \ref{tab:CoPL2ErrorKNNs} because median and MAD suppress the effects of outliers, however the conclusion about relative merits of each method remain the same, with PressNet-Simple outperforming the other methods. 

As shown in Figure~\ref{fig:6a}, the distribution of KNN, PressNet and PressNet-Simple offset errors are concentrated around zero millimeters although with different sized distributions. The mean CoP error for KNN, PressNet, and PressNet-Simple methods are also plotted to provide a reference for how accurate CoP is relative to the ground truth. The mean error in both dimensions is included in Table~\ref{tab:CoPL2Error} for each method. As a point of comparison, a Center of Mass (CoM) localization accuracy of 1\% of  subject height (16.2mm for the dataset, Table~\ref{tab:subject_stats}) is as accurate as the variation between multiple existing CoM calculation methods~\cite{virmavirta2014determining}.

\subsection{Qualitative Evaluation based on MAE}
Figures~\ref{fig:screenshotsCommon} and \ref{fig:screenshotsBest} visualize ground truth, foot pressure predictions and their BoS and CoP for some example frames of the six different subjects.  The foot pressure predictions and ground truth are rendered with independent pressure scales (weight related) to display the full range of pressure for each subject. The color bar in each example frame represents foot pressure intensity in kilopascals (kPa). It ranges from a dark shade of blue, representing 0 kPa, to a dark shade of red, corresponding to the upper limit of foot pressure observed during the performance. The Center of Pressure locations and Base of Support boundaries computed from the ground truth and predicted pressures have also been included, with the ground truth in white, KNN in blue, PressNet in red, and PressNet-Simple in green. In addition to the qualitative comparison by visualization, the respective mean absolute errors with respect to ground truth frames have been calculated and included in (Table~\ref{tab:errortable}) to provide a quantitative comparison of performance. The frames have been chosen to show the ability of PressNet-Simple to generalize to different subjects.

From visual inspection, it is evident that the pressure maps generated by PressNet and PressNet-Simple are more similar to ground truth heatmaps than KNN pressure maps. This is supported by the mean and standard deviation for Similarity, KL Divergence and Information gain metrics of each subject. KNN results are visually poor because KNN is merely picking the K frames with the smallest Euclidean distances between corresponding joints in a cross-subject evaluation. As the style of performance and body dynamics differs for each subject, KNN is unable to generalize to a change in subjects, leading to worse performance on metrics evaluating the spatial distribution of pressure.

Video examples similar to the frames shown in Figures~\ref{fig:screenshotsCommon} and \ref{fig:screenshotsBest} are available on the research project page at: \href{http://vision.cse.psu.edu/research/dynamicsFromKinematics/index.shtml}{Research Project Page}. Observing foot pressure predictions temporally over a sequence of frames, it is observed that KNN predictions are highly inconsistent and fluctuating, whereas the PressNet and PressNet-Simple predictions are temporally smooth and consistent. Since the system operates on a per-frame basis, KNN picks the frames with the nearest pose in the dataset to the current frame, which makes the predictions fluctuate over time. Even though our network is trained using the same per-frame approach, it has learned to predict temporally stable transformation of pose to pressure, making the predictions smooth and more similar to ground truth. All predictions appear to have larger errors when the detected pose is inaccurate, which is most common when multiple joints are occluded in the video. 

\begin{figure}[!t] \centering
    \resizebox*{1.00\linewidth}{!}{
    \begin{tabular}{c}
    \includegraphics[width=\linewidth]{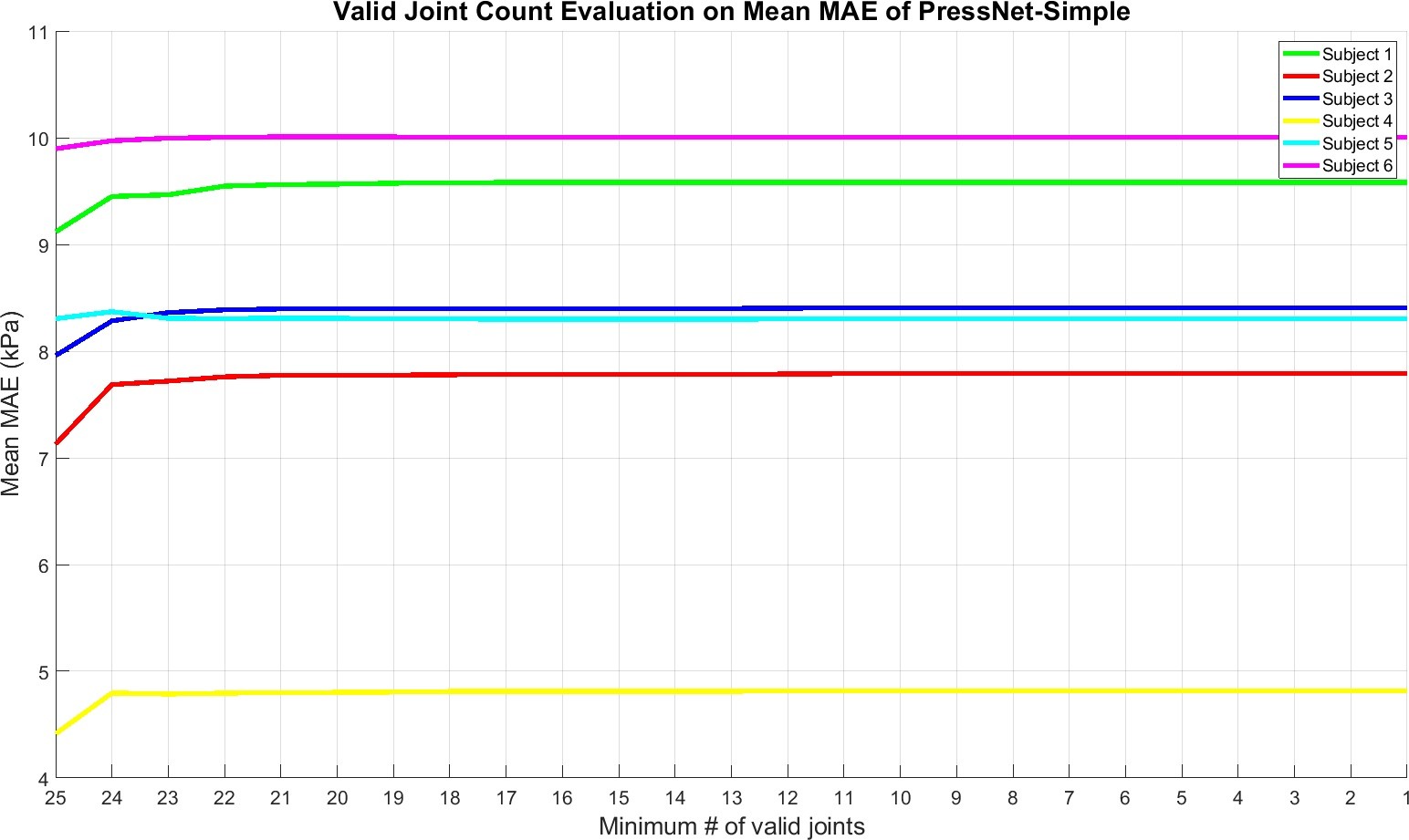} \\
    \includegraphics[width=\linewidth]{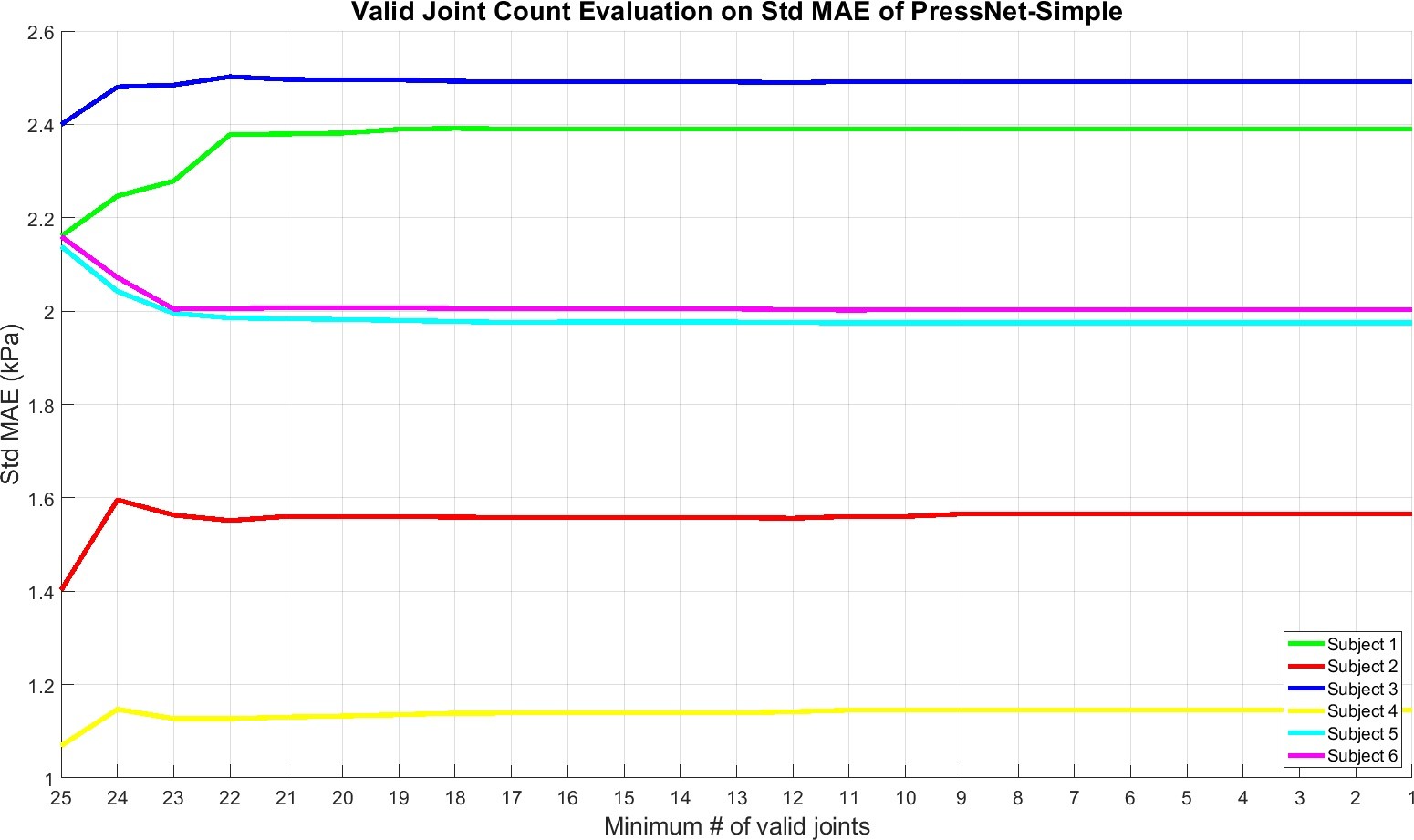}  \\
    \includegraphics[width=\linewidth]{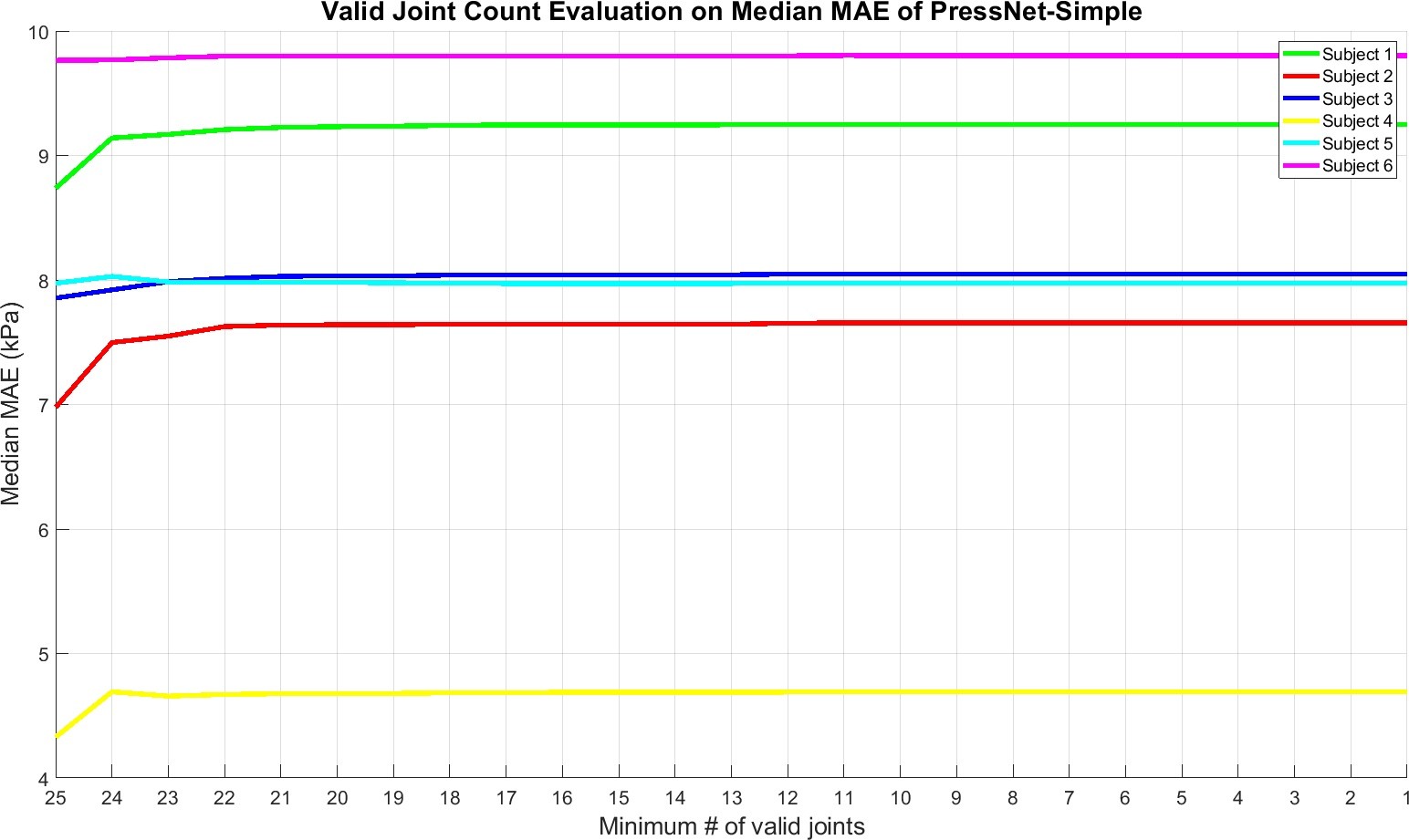} \\
    \includegraphics[width=\linewidth]{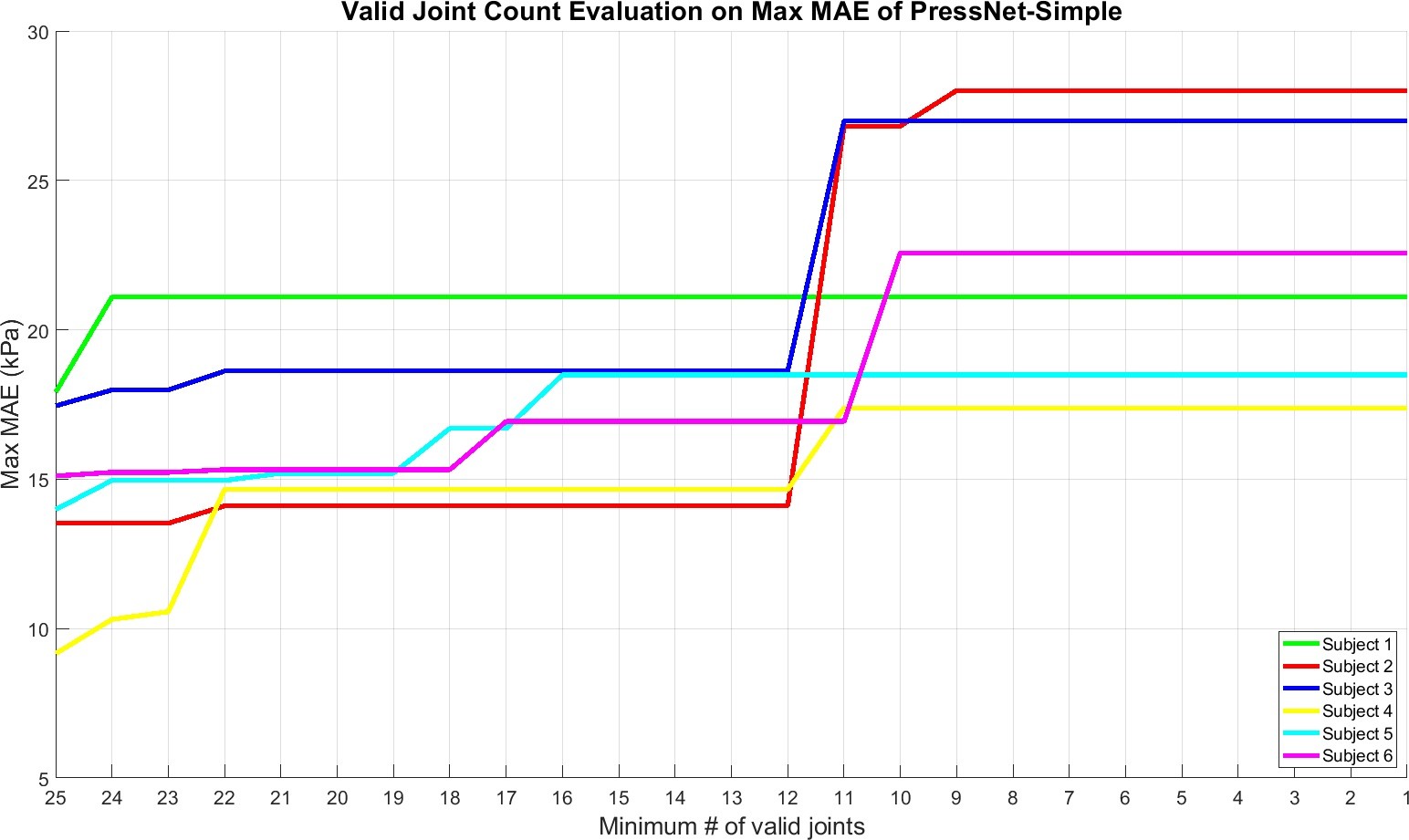} \\
    \end{tabular}
    }
    \caption{Analysis of the impact of joint count on MAE of PressNet-Simple. Mean, Standard Deviation, Median, and Maximum (Top to Bottom) MAE are plotted based on the minimum number of valid joints, as defined by OpenPose, to visualize the impact of the accuracy of input joint accuracy on the PressNet-Simple network.}
    \label{fig:jointCountImpact}
\end{figure}

As a natural condition of video based joint detection, OpenPose does not always provide valid joint locations for all 25 joints in the Body25 model. As a result, all 3 pressure prediction methods must accommodate poses with any number of valid joints. In this research we focus on frames when all 25 joints are valid, as defined by OpenPose giving non-zero confidence to all joints.  Figure~\ref{fig:jointCountImpact} presents the impact on PressNet-Simple error statistics for decreasing numbers of valid joint counts. Shown are the mean, standard deviation, median and maximum MAE for each subject. While mean, standard deviation, and median errors increase noticeably when only a single joint is missing, they do not increase further as the number of missing joints increases. On the other hand, maximum errors increase steadily as more joints get occluded, reaching a plateau when around half of the body joints are missing.

\begin{figure}[!t] \centering
    \resizebox*{1.00\linewidth}{!}{
    \begin{tabular}{c}
    \includegraphics[width=\linewidth]{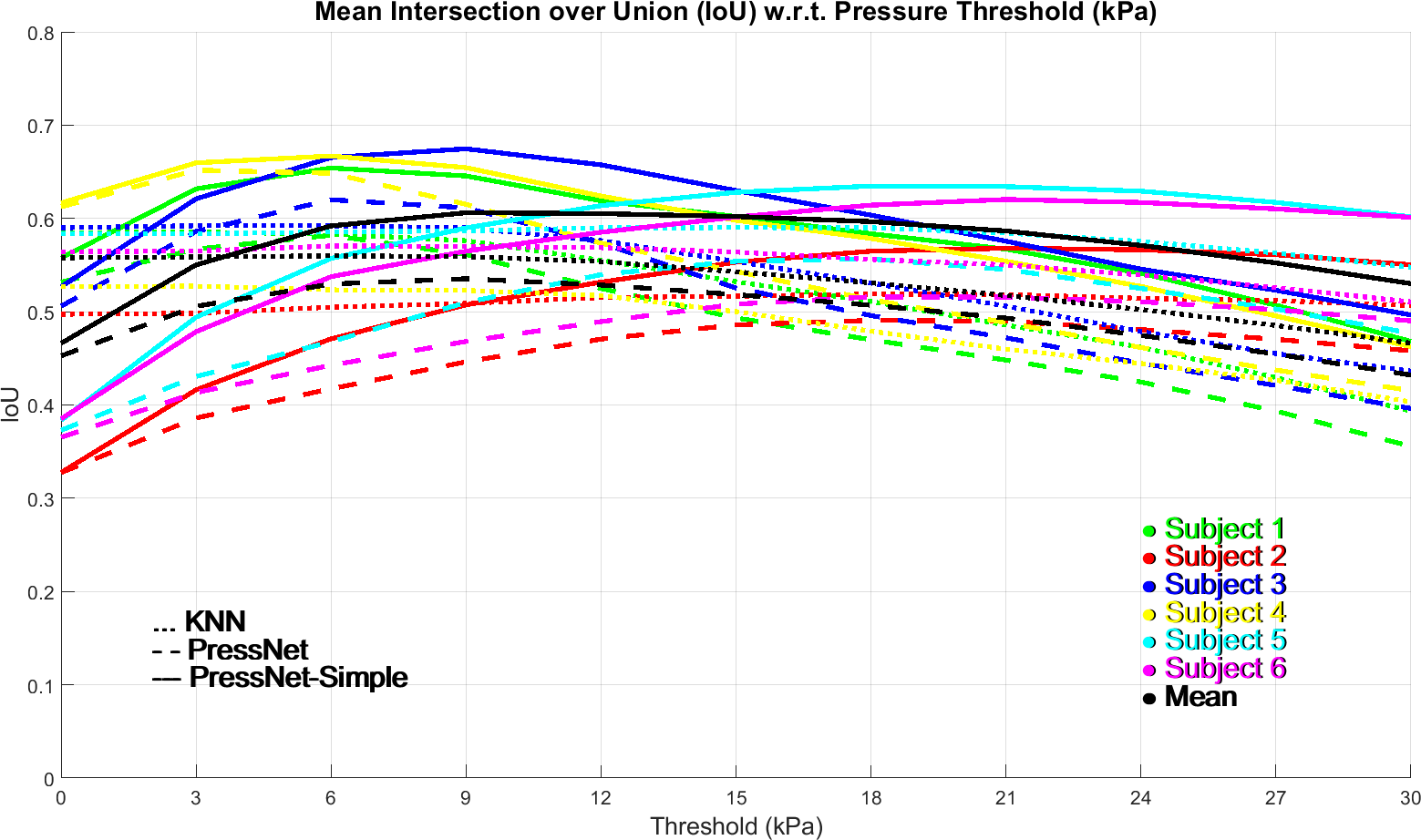} \\
    \end{tabular}
    }
    \caption{Analysis of the impact of thresholds (1-31, 3 kPa steps) on BoS using IoU as a comparison metric between each method and ground truth. This provides only a relative evaluation of the impact of thresholds on performance.}
    \label{fig:IoU_thresh}
\end{figure}

\begin{figure}[!t] \centering
    \resizebox*{1.00\linewidth}{!}{
    \begin{tabular}{c}
    \includegraphics[width=\linewidth]{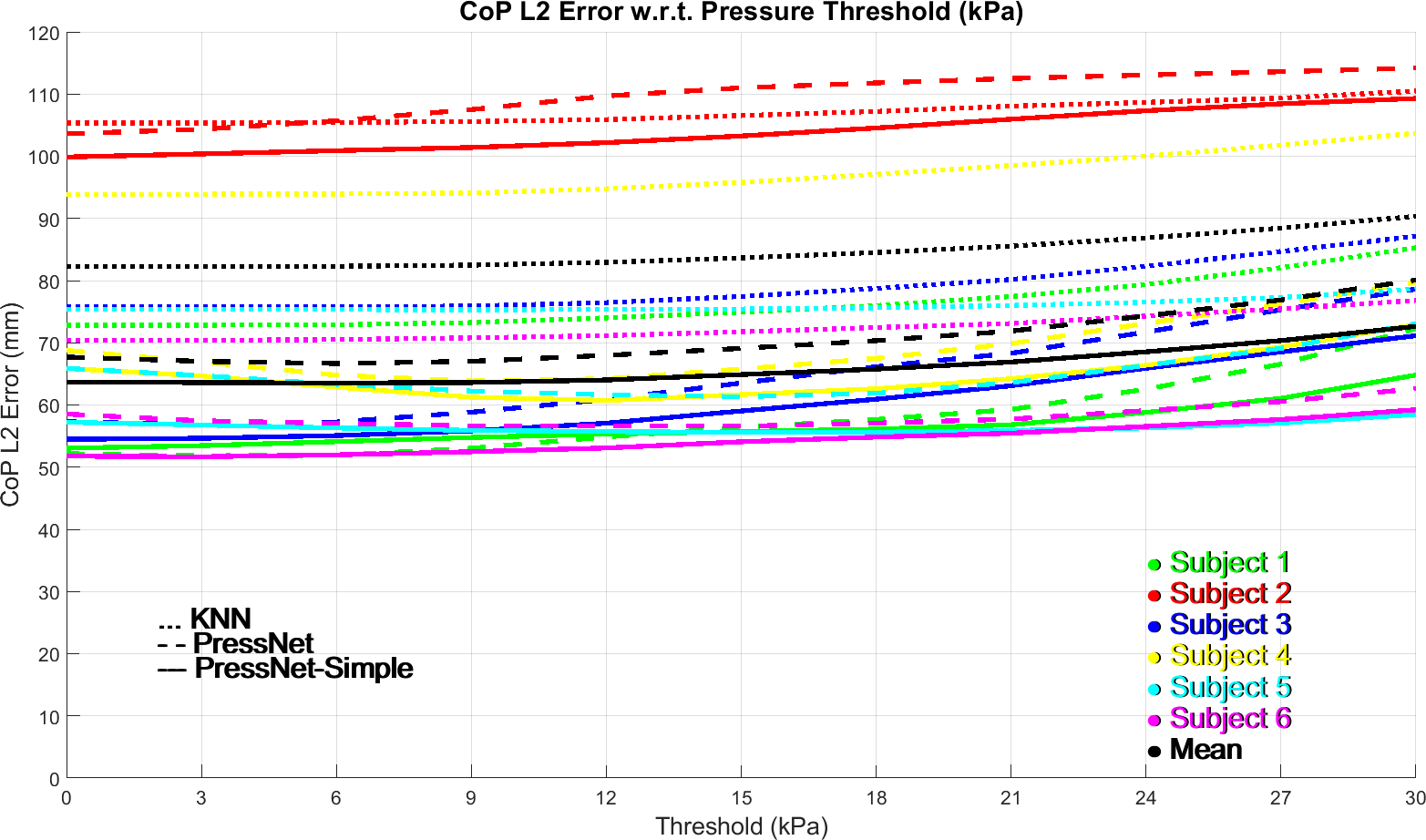} \\
    \end{tabular}
    }
    \caption{Analysis of the impact of thresholds (1-31, 3 kPa steps) on CoP using Euclidean Error as a comparison metric between each method and ground truth. This provides only a relative evaluation of the impact of thresholds on performance. Table~\ref{tab:CoPMedianError} MAD robust errors using  25-joint poses reports smaller error results independent of input data quality and outliers.}
    \label{fig:CoP_thresh}
\end{figure}

\subsection{Threshold Imapct on BoS and CoP Accuracy}
Figures~\ref{fig:IoU_thresh} and~\ref{fig:CoP_thresh} present the impact of the pressure threshold used when evaluating CoP and BoS.  Throughout this research, a 3 kPa threshold was used as it is the limit of the insole sensor technology and this impacts the calculation of both CoP and BoS for each method as well as the ground truth data. Figure~\ref{fig:IoU_thresh} provides Intersection over Union (IoU) for each method tested on one take from each subject. While showing where the IoU of PressNet-Simple is best, it also shows that the range for optimal IoU threshold is between 8 and 15 kPa. Figure~\ref{fig:CoP_thresh} shows that CoP is less sensitive to the selected threshold, with thresholds below 15 kPa yielding roughly equal accuracy and errors increasing slowly beyond 15 kPa. It is intended to present a relative evaluation of the impact of threshold on our performance metrics, CoP and IoU of BoS. This analysis is completed using mean error, rather than MAD robust analysis, and using all frames, independent of the number of detected joints in the input data.  Using the optimal thresholds for CoP (Figure~\ref{fig:CoP_thresh}) one should expect even lower MAD errors than what is shown here based on our robust error analysis results in Table~\ref{tab:CoPMedianError} and Figure~\ref{fig:6b}.

\section{Summary and Conclusion}
The feasibility of regressing foot pressure from 2D joints detected in video has been explored. This is the first work in the computer vision community to establish a direct mapping from 2D human body kinematics to foot pressure dynamics. The effectiveness of our PressNet-Simple network has been shown both quantitatively and qualitatively on a challenging, long, multi-modality Taiji performance dataset. Statistically significant improved results over K-Nearest Neighbor method in foot pressure map estimation from video have been demonstrated.

Furthermore, we demonstrate the use of regressed foot pressure results for estimation of {\em Center of Pressure}, a key component of postural and gait stability. The errors (Table~\ref{tab:CoPL2Error}) are approaching the accepted range for kinesiology studies of Center of Mass (CoM) \cite{virmavirta2014determining}, a corresponding dynamics concept to CoP in stability analysis. Common lab-based CoM estimation methods have been found to have a range of measurement error of around 1\% of subject height, or 16.2 mm for our dataset (Table~\ref{tab:data_stat}),  while the Euclidean mean signed CoP error for PressNet is 10.9 mm and PressNet-Simple is 12.8 mm respectively ("D" columns, Table~\ref{tab:CoPL2Error}).

We hope to extend this work to include more aspects of human body dynamics such as regressing directly to muscle activations, mass distributions, balance, and force. Our goal is to build {\em precision computer vision} tools that estimate various human body dynamics using passive and inexpensive visual sensors, with outcomes validated using bio-mechanically derived data (rather than approximations by human labelers). We foresee  introducing a new and exciting sub-field in computer vision going beyond visually satisfactory human joint/pose detection to the more challenging problem of capturing accurate, quantifiable human body dynamics for scientific applications.  

\section{Acknowledgments}
We would like to thank the six volunteers who contributed 24-form Taiji performances to this study. We would like to acknowledge Andy Luo for his help in rendering the images and videos. We thank the College of Engineering Dean's office of Penn State University for supporting our motion capture lab for research and education. This human subject research is approved through Penn State University IRB Study8085. This work is supported in part by NSF grant IIS-1218729.

\clearpage

{\small
\bibliographystyle{ieee}
\bibliography{ms}
}

\end{document}